\documentclass{article}
\usepackage[11pt]{extsizes}
\usepackage[utf8]{inputenc}
\usepackage{amsmath}
\usepackage{amssymb}
\usepackage[square,sort,comma,numbers]{natbib}
\usepackage[pdftex]{graphicx}
\usepackage[dvipsnames]{xcolor}
\usepackage[hidelinks]{hyperref}
\hypersetup{
    colorlinks = true,
    linkcolor={blue!50!black},
    citecolor={blue!50!black},
    urlcolor={blue!50!black}
}
\usepackage{authblk}
\usepackage[nottoc,numbib]{tocbibind}

\usepackage{geometry}
\usepackage{parskip} 
\usepackage[inline]{enumitem}

\usepackage{booktabs}  
\usepackage{amsfonts}
\usepackage{mathtools}
\usepackage{url}
\usepackage{enumitem}
\usepackage{authblk}
\usepackage{comment}
\usepackage{mdframed}
\usepackage{wrapfig}
\usepackage{floatrow}
\usepackage{xspace}
\usepackage[font=footnotesize,labelfont=bf]{caption}

\usepackage{adjustbox}
\usepackage{bbm}
\newcommand{\hub}{COVID-19 Forecast Hub\xspace}
\usepackage{hhline}
\usepackage{multirow}
\usepackage[all]{nowidow} % Tries to remove widows
\usepackage{colortbl}
\usepackage{booktabs}
\usepackage{bbding} % checkmark symbol
\usepackage{lscape}
\usepackage{microtype} % Improves typography, load after fontpackage is selected
\usepackage{ntheorem}

\theoremseparator{:} 

\newmdtheoremenv[%
  backgroundcolor=white,
  linecolor=blue!60!black,
  linewidth=2pt,
  topline=true,
  rightline=false,
  skipabove=10pt,
  skipbelow=10pt,
  leftline=false]{ourexample}{Application}
  
\newmdtheoremenv[%
  backgroundcolor=gray!20,
  linecolor=red!60!black,
  linewidth=2pt,
  topline=false,
  rightline=false,
  skipabove=10pt,
  skipbelow=10pt,
  leftline=false]{ourbox}{Formulation}

\newmdtheoremenv[%
  backgroundcolor=gray!20,
  linecolor=red!60!black,
  linewidth=2pt,
  topline=false,
  rightline=false,
  skipabove=10pt,
  skipbelow=10pt,
  leftline=false]{regbox}{Box}
  
\usepackage[p,osf]{cochineal}
\usepackage[T1]{fontenc}

\usepackage[scale=.95,type1]{cabin}
\usepackage[cochineal,bigdelims,cmintegrals,vvarbb]{newtxmath}
\usepackage[zerostyle=c,scaled=.94]{newtxtt}
\usepackage[cal=boondoxo]{mathalfa}
\usepackage{microtype}
\usepackage[graphicx]{realboxes}
\usepackage{rotating}

\usepackage{etoolbox}
\apptocmd{\thebibliography}{\raggedright}{}{}

\title{Data-Centric Epidemic Forecasting: A Survey} 
\author[]{Alexander Rodr\'iguez\footnote{These authors contributed equally to this work.} ,  Harshavardhan Kamarthi\textsuperscript{*}, Pulak Agarwal, \\ Javen Ho, Mira Patel, Suchet Sapre, and B. Aditya Prakash\footnote{To whom correspondence should be addressed: \texttt{\{arodriguezc, hkamarthi3, badityap\}@gatech.edu}}}
\affil[]{\small College of Computing, Georgia Institute of Technology, USA}
\date{}

\begin{document}

\maketitle

\begin{abstract}

The COVID-19 pandemic has brought forth the importance of epidemic forecasting for decision makers in multiple domains, ranging from public health to the economy as a whole. While forecasting epidemic progression is frequently conceptualized as being analogous to weather forecasting, however it has some key differences and remains a non-trivial task.  The spread of diseases is subject to multiple confounding factors spanning human behavior, pathogen dynamics, weather and environmental conditions. 
Research interest has been fueled by the increased availability of rich data sources capturing previously unobservable facets and also due to initiatives from government public health and funding agencies like forecasting challenges and large scale team science initiatives. 
This has resulted, in particular, in a spate of work on 'data-centered' solutions which have shown potential in enhancing our forecasting capabilities by leveraging non-traditional data sources as well as recent innovations in AI and machine learning.
This survey delves into various data-driven methodological and practical advancements and introduces a conceptual framework to navigate through them.
First, we enumerate the large number of epidemiological datasets and novel data streams that are relevant to epidemic forecasting, capturing various factors like symptomatic online surveys, retail and commerce, mobility, genomics data and more. 
Next, we discuss methods and modeling paradigms focusing on the recent data-driven statistical and deep-learning based methods as well as on the novel class of \emph{hybrid} models that combine domain knowledge of mechanistic models with the effectiveness and flexibility of statistical approaches.
We also discuss experiences and challenges that arise in real-world deployment of these forecasting systems including decision-making informed by forecasts. Finally, we highlight some challenges and open problems found across the forecasting pipeline.

\vspace{1in}

\par \noindent 
\textbf{Keywords:} Artificial Intelligence, Machine Learning, Data Mining, Epidemiology, Computational Modeling, Forecasting, Public Health, Literature Survey. %Data-driven Modeling.

\end{abstract}

\newpage
\tableofcontents

\newpage
\section{Introduction}
\label{sec:intro}

The devastating impact of the COVID-19 pandemic on human lives, economic development and society as a whole has exemplified our vulnerability to major infectious diseases and epidemics.
While the science of epidemic forecasting, in many respects, is still in its initial stages, the current pandemic and the ones before it (like H1N1 and Ebola) have shown its crucial importance. 
Preventing and responding to such pandemics requires actionable epidemic forecasts e.g. to design  effective healthcare policies and optimal supply chain decisions. 
Generating such forecasts however has multiple inter-disciplinary challenges~\cite{brauer2017mathematical}.
These range from understanding biological processes governing pathogen evolution, response to immunization, and drug resistance to population-level modeling of heterogeneous groups and their interactions within and across communities. %, key to understand the spatial spread of the disease. 

There has been an increasing interest in data-centered solutions for epidemic forecasting~\cite{marathe2013computational}, building on several initiatives in the past few years from both government public health and funding agencies. For instance, in 2013, the U.S. Centers for Disease Control and Prevention (CDC) introduced the FluSight challenge~\cite{biggerstaff2014estimates}, which has not only helped improve flu forecasting capabilities and public health decision making but also helped grow a community of researchers in this topic.
Similar initiatives have followed, for Ebola~\cite{viboud2018rapidd}, Dengue~\cite{johansson2019open}, and also COVID-19~\cite{cramer2021united}, led by institutions around the globe such as the European CDC~\cite{ecdc2021hub,bracher2021pre}, IARPA~\cite{iarpa_osi}, and PAHO~\cite{mensua2009pandemic} in Latin America. These forecasting initiatives have given an unprecedented opportunity to researchers to observe  both the successes and gaps in the current science of forecasting. Similarly, agencies like the National Science Foundation (NSF), the National Institutes of Health (NIH), and US Army Research have held a spate of recent symposia~\cite{nsf_pipp} and funding calls related to pandemic forecasting, which has given a much needed impetus to this topic. This interest has also culminated in the establishment of the first Center for Forecasting and Outbreak Analytics by the US CDC in 2021~\cite{cdc_center}.

Our survey delves into such data-driven computational methods, which have shown great potential in leveraging advances in data science and artificial intelligence
and the incorporation of novel sources of information, from biological to behavioral. 
Indeed, there has been an increased availability of data from reliable sources (several of them publicly accessible), a trend only accelerated by the COVID-19 pandemic. This includes richer epidemiological datasets and novel digital data streams like mobility \cite{aktay2020google,wesolowski2012quantifying}, online surveys \cite{rebeiro2021impact,delphi2021survey}, and wastewater samples \cite{peccia2020measurement}. 
Fueled by these factors, in the last two years, we have also seen a number of technical innovations using machine learning and deep learning techniques, which have opened new horizons in the science of epidemic forecasting.

\begin{figure}[htb]
\includegraphics [width = 0.98\textwidth]{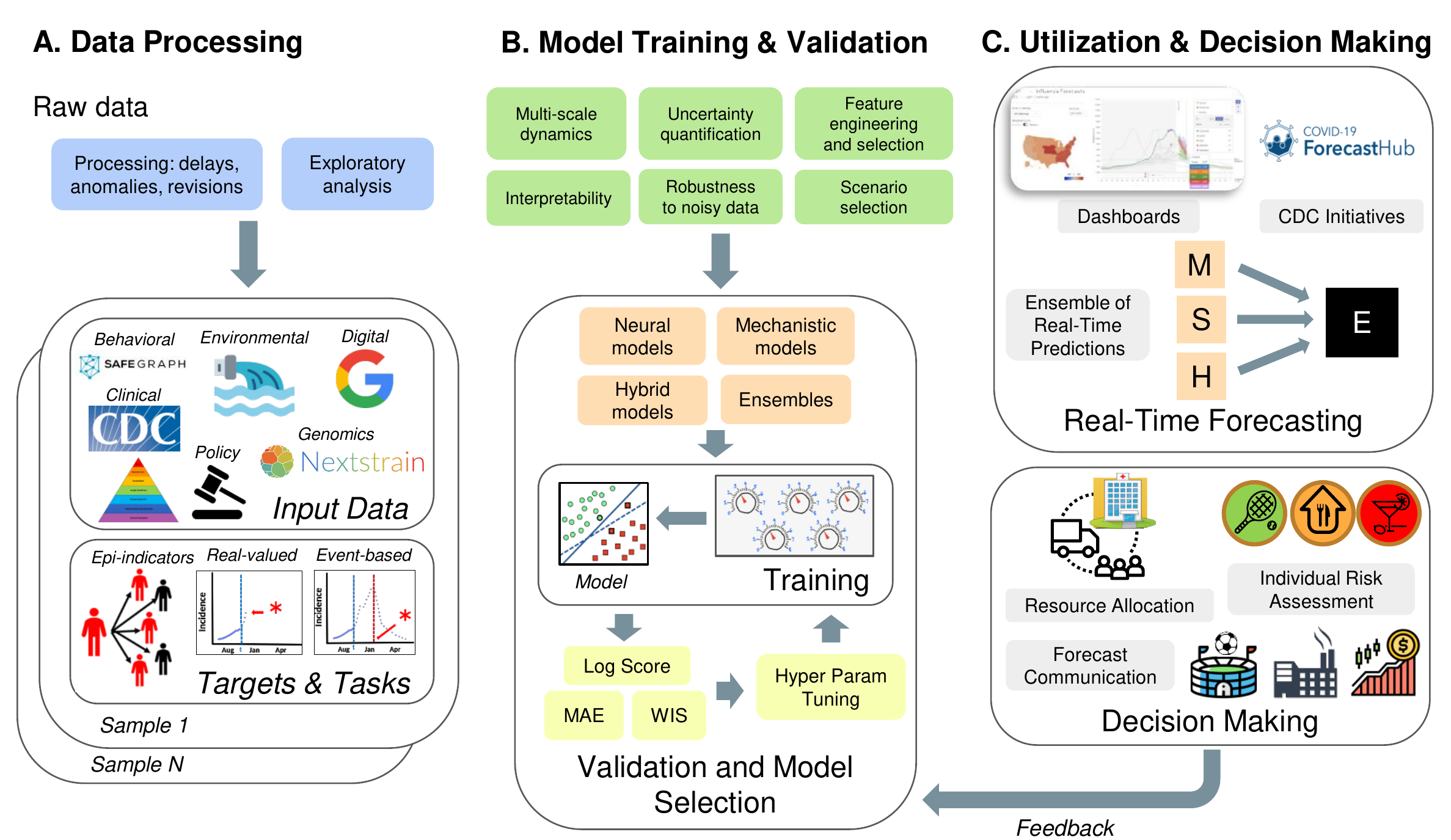}
\caption{\textbf{Overview of the data-centric epidemic forecasting pipeline.} (A) Before modeling, we have to prepare the data, which includes collection and exploratory analysis to handle data quality issues. At this stage, we also define epidemic targets and tasks. (B) Model development takes in consideration multiple facets from the epidemic spread (e.g., multiscale dynamics) and forecast utilization (e.g., uncertainty quantification). Validation and model selection require using quantitative metrics to assess how actionable and reliable the forecasts. (C) Real-time forecasts have multiple uses including dashboards, ensemble composition, and other public health initiatives. These serve as a platform for decision making for resource allocation, individual risk assessment, and general communication to the public.}
\label{fig:pipeline}
\end{figure}

This survey is an effort to encompass recent methodological and practical advances at an opportune time to help and enable the broader computational and data/ML/AI communities to engage in this area. 
We can conceptualize the data-centric computational epidemic forecasting pipeline as in Figure~\ref{fig:pipeline}. We give an overview of these components and broadly classify them into three: data processing, model training and validation, and utilization and decision making.
Broadly speaking, epidemic forecasting aims to provide information about the trajectory of the epidemic spread measured by multiple indicators (e.g., weekly number of patients visiting hospitals).
Our pipeline starts with data from heterogeneous sources and modalities which captures multiple facets of the epidemic spread.
After preparing the data and defining specific targets and resolution in spatial and temporal scales, modeling training and deployment takes place considering 
characteristics from the disease spread (e.g., multi-scale dynamics), data considerations (e.g., noisy data), and requirements from public health officials and the general public (e.g., uncertainty quantification). 
Our survey collects all these elements and examines recent developments and current important trends in each of them.
In Table~\ref{tab:datasets}, we summarize and classify the methodological work we surveyed. We emphasize the type of data used by each modeling paradigm, the tasks for which they have been utilized, and distinctive modeling features.

\subsection{Context}

Typically, earlier related surveys focus on traditional epidemiological methods with little emphasis on data-centric perspectives and usually circumscribe their scope to one disease~\cite{nsoesie2014systematic,chretien2014influenza}. In contrast, while no survey can be completely exhaustive, we provide a broader and more comprehensive perspective spanning multiple modeling approaches.
There are a few recent surveys that discuss the use/application of AI/ML techniques for various healthcare challenges arising from epidemics~\cite{shinde2020forecasting,chen2021survey}, which largely focus on clinical problems e.g. medical imaging. 
In contrast, we focus exclusively on epidemiological forecasting and provide a broader framework to understand the spectrum of modeling paradigms ranging from the traditional mechanistic approaches to statistical machine learning based models.
Indeed, here we go further than the previous work focusing on paradigms that aim to bridge the gap between epidemiology and ML communities by exploiting the advantages of both approaches (the so-called \emph{hybrid methods}). 
We also provide a comprehensive discussion on the challenges and solutions associated with real-time deployment of these forecasting systems. We study them across the multiple stages of the pipeline including data collection, modeling, evaluation, as well as decision making.
Finally, we provide a holistic discussion of open problems and research directions in the field of epidemic forecasting.

\begin{sidewaystable}
\caption{\textbf{Overview of methods discussed in this survey.}}
\label{tab:datasets}
\centering
\arrayrulecolor{black}
% \Rotatebox{90}{%
\resizebox{\linewidth}{!}{
\begin{tabular}{!{\color{black}\vrule}c!{\color{black}\vrule}l!{\color{black}\vrule}l!{\color{black}\vrule}l!{\color{black}\vrule}l!{\color{black}\vrule}l!{\color{black}\vrule}l!{\color{black}\vrule}l!{\color{black}\vrule}l!{\color{black}\vrule}l!{\color{black}\vrule}l!{\color{black}\vrule}l!{\color{black}\vrule}l!{\color{black}\vrule}l!{\color{black}\vrule}l!{\color{black}\vrule}l!{\color{black}\vrule}l!{\color{black}\vrule}l!{\color{black}\vrule}l!{\color{black}\vrule}l!{\color{black}\vrule}l!{\color{black}\vrule}l!{\color{black}\vrule}l!{\color{black}\vrule}} 
\toprule
\multicolumn{1}{!{\color{black}\vrule}l!{\color{black}\vrule}}{~} &  \multicolumn{1}{c!{\color{black}\vrule}}{\textbf{Modeling Paradigms}}      & \multicolumn{1}{c!{\color{black}\vrule}}{\textbf{Papers} } & \multicolumn{7}{c!{\color{black}\vrule}}{\textbf{Data} } & \multicolumn{3}{c!{\color{black}\vrule}}{\textbf{Tasks}}                                                                                                                                                 & \multicolumn{10}{c!{\color{black}\vrule}}{\textbf{Model Features}}                                                                                                                                                                                                                                                                                                                                                                                                                                                                                                                                                                                                                   \\ 
\arrayrulecolor{black}\cline{1-23}
\multicolumn{1}{!{\color{black}\vrule}l!{\color{black}\vrule}}{~} & \multicolumn{1}{l!{\color{black}\vrule}}{~}              &                                                                                                                                                                                                              & \multicolumn{1}{c!{\color{black}\vrule}}{\rotatebox{90}{Clinical surveillance data}}                                                                   & \multicolumn{1}{c!{\color{black}\vrule}}{\rotatebox{90}{Electronic surveillance data}}                                                                      & \multicolumn{1}{c!{\color{black}\vrule}}{\rotatebox{90}{Behavioral data}}               & \multicolumn{1}{r!{\color{black}\vrule}}{\rotatebox{90}{Genomics data}} & \multicolumn{1}{r!{\color{black}\vrule}}{\rotatebox{90}{Environmental data}}    & \multicolumn{1}{r!{\color{black}\vrule}}{\rotatebox{90}{Crowd-sourced predictions}}                                                                 & \multicolumn{1}{r!{\color{black}\vrule}}{\rotatebox{90}{Policy data}}                                                      & \multicolumn{1}{r!{\color{black}\vrule}}{\rotatebox{90}{Real-valued prediction}} & \multicolumn{1}{r!{\color{black}\vrule}}{\rotatebox{90}{Event-based prediction}} & \multicolumn{1}{r!{\color{black}\vrule}}{\rotatebox{90}{Epidemiological indicators}} & \multicolumn{1}{r!{\color{black}\vrule}}{\rotatebox{90}{Deep learning}} & \multicolumn{1}{r!{\color{black}\vrule}}{\rotatebox{90}{\begin{tabular}{@{}c@{}}Geographical granularity\\ (Cou=County, C= Country, \\ S=State, R=Region)\end{tabular}}} & \multicolumn{1}{r!{\color{black}\vrule}}{\rotatebox{90}{\begin{tabular}{@{}c@{}}Temporal granularity\\ (D=Days, W=Weeks) \end{tabular}}} & \multicolumn{1}{r!{\color{black}\vrule}}{\rotatebox{90}{Gradient-based learning}} & \multicolumn{1}{r!{\color{black}\vrule}}{\rotatebox{90}{Uncertainty estimation}} & \multicolumn{1}{r!{\color{black}\vrule}}{\rotatebox{90}{Handle data quality issues}} & \multicolumn{1}{r!{\color{black}\vrule}}{\rotatebox{90}{Spatio-temporal modeling}} & \multicolumn{1}{r!{\color{black}\vrule}}{\rotatebox{90}{Interpretability}} & \multicolumn{1}{r!{\color{black}\vrule}}{\rotatebox{90}{Transfer learning}} & \multicolumn{1}{r!{\color{black}\vrule}}{\rotatebox{90}{Expert in the loop}}  \\ 
\arrayrulecolor{black}\hline
\multirow{3}{*}{\rotatebox{90}{\textbf{Mech}}}                                      & \rotatebox{0}{\textbf{Mass-Action Models}}                                                                                       & \cite{wu2020nowcasting,ferguson2020impact,morozova2021one,gibson2020real,gu2020covid}                                                                                                              & \Checkmark                                                                    & ~                                                                      & ~                                                         & ~                                                      & ~                                                            & ~                                                                 & ~                                                     & \Checkmark                                                                 & \Checkmark                                                     & \Checkmark                                                                    & ~                                                       & C/S                                                              & D/W                                                              & ~                                                                 & ~                                                                & ~                                                                    & ~                                                                  & ~                                                          & ~                                                           & ~                                                             \\ 
\cline{2-23}
                                                                  & \rotatebox{0}{\textbf{Metapopulation Models}}                                                                                   & \cite{mossong2008social,balcan2009multiscale,chinazzi2020effect,pei2018forecasting,venkatramanan2021forecasting,srivastava2020learning,gopalakrishnan2020globally,geng2021kernel,wang2021bridging} & \Checkmark                                                                    & ~                                                                      & \Checkmark                                                         & ~                                                      & ~                                                            & ~                                                                 & ~                                                     & \Checkmark                                                                 & \Checkmark                                                     & ~                                                                    & ~                                                       & C/S/Cou/R                                        & D/W                                                      & ~                                                                 & ~                                                                & ~                                                                    & ~                                                                  & ~                                                          & ~                                                           & ~                                                             \\ 
\cline{2-23} & \rotatebox{0}{\textbf{Agent-Based Models}}                                                                                    & \cite{eubank2004modelling,bisset2009epifast,nsoesie2013simulation,tabataba2017epidemic,mousavi2012enhanced,grefenstette2013fred,venkatramanan2018using}                                            & \Checkmark                                                                    & ~                                                                      & \Checkmark                                                         & ~                                                      & ~                                                            & ~                                                                 & \Checkmark                                                     & \Checkmark                                                                 & \Checkmark                                                     & ~                                                                    & ~                                                       & C/Cou                                                     & D/W                                                       & ~                                                                 & ~                                                                & ~                                                                    & ~                                                                  & ~                                                          & ~                                                           & ~                                                             \\ 
                                                                  
\hline
\multirow{20}{*}{\rotatebox{90}{\textbf{Statistical}}}                                            & 
\multicolumn{21}{l}{\textbf{Regression Models}} &~\\
%{\rotatebox{0}{\textbf{Regression Models}}}  &\multicolumn{20}{c}{~}&~\\
\cline{2-23} & Sparse Linear Models                             & \cite{polgreen2008using,ginsberg2009detecting,liu2020machine,mciver2014wikipedia,santillana2014using}                                                                                              & \Checkmark                                                                    & \Checkmark                                                                      & ~                                                         & ~                                                      & ~                                                            & ~                                                                 & ~                                                     & \Checkmark                                                                 & \Checkmark                                                     & ~                                                                    & ~                                                       & C/S                                                      & W                                                           & \Checkmark                                                                 & ~                                                                & ~                                                                    & ~                                                                  & \Checkmark                                                          & ~                                                           & ~                                                             \\ 
\cline{2-23}
                                                                                                                            & Auto-regressive Models                           & \cite{soebiyanto2010modeling,ray2018prediction,broniatowski2015using,kandula2019near,yang2015accurate,aiken2020real,yang2017advances}                                                              & \Checkmark                                                                    & \Checkmark                                                                      & ~                                                         & ~                                                      & \Checkmark                                                            & ~                                                                 & ~                                                     & \Checkmark                                                                 & \Checkmark                                                     & ~                                                                    & ~                                                       & S/C/Cty/R                                          & W                                                           & \Checkmark                                                                 & ~                                                                & ~                                                                    & ~                                                                  & \Checkmark                                                          & ~                                                           & ~                                                             \\ 
\cline{2-23}
                                                                                                                            & Complex Regression Models                        & \cite{santillana2015combining,signorini2011use,chakraborty2014forecasting,wang2015dynamic}                                                                                                         & \Checkmark                                                                    & \Checkmark                                                                      & \Checkmark                                                         & ~                                                      & \Checkmark                                                            & ~                                                                 & ~                                                     & \Checkmark                                                                 & ~                                                     & ~                                                                    & ~                                                       & R/C/S                                               & W                                                           & \Checkmark                                                                 & ~                                                                & \Checkmark                                                                    & ~                                                                  & ~                                                          & ~                                                           & ~                                                             \\ 
\cline{2-23}
                                                                                                                            & Hierarchical Models                              & \cite{ning2019accurate,yang2021use,matsubara2014funnel,zou2018multi}                                                                                                                               & \Checkmark                                                                    & \Checkmark                                                                      & ~                                                         & ~                                                      & ~                                                            & \Checkmark                                                                 & ~                                                     & \Checkmark                                                                 & ~                                                     & ~                                                                    & ~                                                       & S/C                                                     & D/W                                                     & \Checkmark                                                                 & \Checkmark                                                                & ~                                                                    & \Checkmark                                                                  & \Checkmark                                                          & \Checkmark                                                           & ~                                                             \\ 
\cline{2-23}
                                                                  & \textbf{Vision and Language Models} &~&~&~&~&~&~&~&~&~&~&~&~&~&~&~&~&~&~&~&~&~\\         
\cline{2-23} & Vision Models                                    & \cite{butler2014satellite}                                                                                                                                                                         & \Checkmark                                                                    & \Checkmark                                                                      & ~                                                         & ~                                                      & ~                                                            & ~                                                                 & ~                                                     & \Checkmark                                                                 & ~                                                     & ~                                                                    & ~                                                       & C                                                            & D                                                            & \Checkmark                                                                 & ~                                                                & ~                                                                    & ~                                                                  & ~                                                          & ~                                                           & ~                                                             \\ 
\cline{2-23}
                                                                                                                            & Language-based Models                            & \cite{lampos2010flu,culotta2010towards,lamb2013separating,paul2014twitter,zou2019transfer}                                                                                                         & \Checkmark                                                                    & \Checkmark                                                                      & ~                                                         & ~                                                      & ~                                                            & ~                                                                 & ~                                                     & \Checkmark                                                                 & ~                                                     & ~                                                                    & ~                                                       & C                                                 & D/W                                                       & \Checkmark                                                                 & ~                                                                & ~                                                                    & ~                                                                  & ~                                                          & \Checkmark                                                           & ~                                                             \\ 
\cline{2-23}
                                                                                                                            & Probabilistic topic models                       & \cite{paul2012model,paul2011you,chen2014flu,hua2018social,rekatsinas2015sourceseer}                                                                                                                & \Checkmark                                                                    & \Checkmark                                                                      & ~                                                         & ~                                                      & ~                                                            & ~                                                                 & ~                                                     & \Checkmark                                                                 & ~                                                     & ~                                                                    & ~                                                       & C                                                           & W                                                           & ~                                                                 & \Checkmark                                                                & ~                                                                    & ~                                                                  & \Checkmark                                                          & ~                                                           & ~                                                             \\ 
\cline{2-23}
                                                                  & \multicolumn{21}{l}{\textbf{Neural Models}} &~\\         
\cline{2-23}                  & Off the Shelf                                    & \cite{volkova2017forecasting,venna2018novel,wu2020deep,ayyoubzadeh2020predicting}                                                                                                                  & \Checkmark                                                                    & \Checkmark                                                                      & \Checkmark                                                         & ~                                                      & \Checkmark                                                            & ~                                                                 & ~                                                     & \Checkmark                                                                 & ~                                                     & ~                                                                    & \Checkmark                                                       & C/R                                                     & W                                                           & \Checkmark                                                                 & ~                                                                & ~                                                                    & ~                                                                  & ~                                                          & ~                                                           & ~                                                             \\ 
\cline{2-23}
                                                                                                                          & Similarity modeling                              & \cite{adhikari2019epideep,wang2020examining,jin2021inter}                                                                                                                                          & \Checkmark                                                                    & ~                                                                      & \Checkmark                                                         & ~                                                      & ~                                                            & ~                                                                 & ~                                                     & \Checkmark                                                                 & \Checkmark                                                     & ~                                                                    & \Checkmark                                                       & C/R                                                     & W                                                           & \Checkmark                                                                 & \Checkmark                                                                & ~                                                                    & ~                                                                  & ~                                                          & ~                                                           & ~                                                             \\ 
\cline{2-23}
                                                                                                                        & Transfer Learning                                & \cite{rodriguez_steering_2021,panagopoulos2020transfer}                                                                                                                                          & \Checkmark                                                                    & ~                                                                      & \Checkmark                                                         & ~                                                      & ~                                                            & ~                                                                 & ~                                                     & \Checkmark                                                                 & ~                                                     & ~                                                                    & \Checkmark                                                       & C/R                                                     & D/W                                                       & \Checkmark                                                                 & \Checkmark                                                                & \Checkmark                                                                    & ~                                                                  & ~                                                          & \Checkmark                                                           & ~                                                             \\ 
\cline{2-23}
                                                                                                                 & Multimodal Data                                  & \cite{ibrahim2021variational,ramchandani2020deepcovidnet}                                                                                                                                          & \Checkmark                                                                    & ~                                                                      & \Checkmark                                                         & ~                                                      & ~                                                            & ~                                                                 & ~                                                     & \Checkmark                                                                 & ~                                                     & ~                                                                    & \Checkmark                                                       & C/R                                                     & D/W                                                       & \Checkmark                                                                 & \Checkmark                                                                & ~                                                                    & ~                                                                  & ~                                                          & ~                                                           & ~                                                             \\ 
\cline{2-23}
                                                                                                                        & Spatial Modeling                                 & \cite{wu2018deep,deng2020cola,kapoor2020examining,wang2020using}                                                                                                                                   & \Checkmark                                                                    & \Checkmark                                                                      & \Checkmark                                                         & ~                                                      & ~                                                            & ~                                                                 & ~                                                     & \Checkmark                                                                 & ~                                                     & ~                                                                    & \Checkmark                                                       & C/R/S                                               & W                                                           & \Checkmark                                                                 & \Checkmark                                                                & ~                                                                    & \Checkmark                                                                  & ~                                                          & ~                                                           & ~                                                             \\ 
\cline{2-23}
                                                                  & \multicolumn{21}{l}{\textbf{Density Estimation}} &~\\         
\cline{2-23}                             & Kernel density estimation                        & \cite{ray2017infectious,nelsen2007introduction,brooks2018nonmechanistic}                                                                                                                           & \Checkmark                                                                    & ~                                                                      & ~                                                         & ~                                                      & ~                                                            & ~                                                                 & ~                                                     & \Checkmark                                                                 & \Checkmark                                                     & ~                                                                    & ~                                                       & C/R                                                     & W                                                           & ~                                                                 & \Checkmark                                                                & ~                                                                    & \Checkmark                                                                  & \Checkmark                                                          & ~                                                           & ~                                                             \\ 
\cline{2-23}
                                                                                                                           & Parametric Bayesian inference                    & \cite{van2014risk,brooks2015flexible}                                                                                                                                                              & \Checkmark                                                                    & ~                                                                      & ~                                                         & ~                                                      & ~                                                            & ~                                                                 & ~                                                     & \Checkmark                                                                 & \Checkmark                                                     & ~                                                                    & ~                                                       & C/S                                                           & W                                                           & ~                                                                 & \Checkmark                                                                & ~                                                                    & ~                                                                  & \Checkmark                                                          & ~                                                           & ~                                                             \\ 
\cline{2-23}
                                                                                                                         & Non-parametric methods~                          & \cite{senanayake2016predicting,zimmer2020influenza}                                                                                                                                                & \Checkmark                                                                    & \Checkmark                                                                      & ~                                                         & ~                                                      & \Checkmark                                                            & ~                                                                 & ~                                                     & \Checkmark                                                                 & ~                                                     & ~                                                                    & ~                                                       & C                                                            & W                                                           & ~                                                                 & \Checkmark                                                                & ~                                                                    & ~                                                                  & \Checkmark                                                          & ~                                                           & ~                                                             \\ 
\cline{2-23}
                                                                                                                           & Neural uncertainty quantification                & \cite{kamarthi2021epifnp,kamarthi2021camul}                                                                                                                                                        & \Checkmark                                                                    & ~                                                                      & ~                                                         & ~                                                      & ~                                                            & ~                                                                 & ~                                                     & \Checkmark                                                                 & ~                                                     & ~                                                                    & \Checkmark                                                       & C/R                                                     & W                                                          & \Checkmark                                                                 & \Checkmark                                                                & \Checkmark                                                                    & \Checkmark                                                                  & \Checkmark                                                          & ~                                                           & ~                                                             \\ 
\hline
\multirow{10}{*}{\rotatebox{90}{\textbf{Hybrid}}}                                           & \multicolumn{21}{l}{\textbf{Mechanistic with Statistical Components}} &~\\         
\cline{2-23}                            

& Data Assimilation                                & \cite{shaman2012forecasting,kandula2019improved,pei2020aggregating,yang2021development}                                                                                                                               & \Checkmark                                                                    & \Checkmark                                                                      & ~                                                         & \Checkmark                                                      & \Checkmark                                                            & ~                                                                 & ~                                                     & \Checkmark                                                                 & \Checkmark                                                     & ~                                                                    & ~                                                       & S and C                                                  & W                                                           & ~                                                                 & \Checkmark                                                                & \Checkmark                                                                    & ~                                                                  & ~                                                          & ~                                                           & ~                                                             \\ 
\cline{2-23}
                                                                                                                          & Statistical estimation of mechanistic parameter~ & \cite{zhang2017forecasting,arik2020interpretable,bhouri2020covid,qian2020and,ghamizi2020data}                                                                                                      & \Checkmark                                                                    & \Checkmark                                                                      & \Checkmark                                                         & ~                                                      & \Checkmark                                                            & ~                                                                 & \Checkmark                                                     & \Checkmark                                                                 & ~                                                     & \Checkmark                                                                    & \Checkmark                                                       & C/S                                                      & D/W                                                       & \Checkmark                                                                 & ~                                                                & ~                                                                    & \Checkmark                                                                  & \Checkmark                                                          & ~                                                           & ~                                                             \\ 
\cline{2-23}
                                                                                                                  & Discrepancy Modeling                             & \cite{osthus2019dynamic,osthus2021multiscale,osthus2021fast,kamarthi2021back2future,wu2021deepgleam}                                                                                               & \Checkmark                                                                    & ~                                                                      & \Checkmark                                                         & ~                                                      & ~                                                            & ~                                                                 & ~                                                     & \Checkmark                                                                 & \Checkmark                                                     & ~                                                                    & \Checkmark                                                       & C/S                                                      & W                                                           & \Checkmark                                                                 & \Checkmark                                                                & \Checkmark                                                                    & \Checkmark                                                                  & ~                                                          & ~                                                           & ~                                                             \\ 
\cline{2-23}
                                                                  & \multicolumn{21}{l}{\textbf{Mechanism informs statistical model}} &~\\         
\cline{2-23}                                & Learning from synthetic and simulation data      & \cite{wang2019defsi,wang2020tdefsi,liu2020machine}                                                                                                                                                 & \Checkmark                                                                    & \Checkmark                                                                      & \Checkmark                                                         & ~                                                      & \Checkmark                                                            & ~                                                                 & ~                                                     & \Checkmark                                                                 & \Checkmark                                                     & ~                                                                    & \Checkmark                                                       & C/R and S                                           & W                                                           & \Checkmark                                                                 & ~                                                                & \Checkmark                                                                    & ~                                                                  & ~                                                          & ~                                                           & ~                                                             \\ 
\cline{2-23}
                                                                                                                        & Learning with mechanistic constraints            & \cite{kargas2021stelar,Gao2021STANSA}                                                                                                                                                              & \Checkmark                                                                    & ~                                                                      & ~                                                         & ~                                                      & ~                                                            & ~                                                                 & ~                                                     & \Checkmark                                                                 & ~                                                     & ~                                                                    & \Checkmark                                                       & S and C                                                  & D                                                          & \Checkmark                                                                 & ~                                                                & \Checkmark                                                                    & ~                                                                  & \Checkmark                                                          & ~                                                           & ~                                                             \\ 
\cline{2-23}
                                                                  & \multicolumn{21}{l}{\textbf{Wisdom of Crowds}} &~\\         
\cline{2-23}                             
& Experts and prediction markets~                  & \cite{farrow2017human,recchia2021well,nadella2020forecasting,mcandrew2020expert,mcandrew2021aggregating,polgreen2007use,tung2015using,Shea2020harnessing,hemming2018practical}                     & \Checkmark                                                                    & ~                                                                      & ~                                                         & ~                                                      & ~                                                            & \Checkmark                                                                 & ~                                                     & \Checkmark                                                                 & \Checkmark                                                     & ~                                                                    & ~                                                       & ~                                                                  & ~                                                              & ~                                                                 & ~                                                                & ~                                                                    & ~                                                                  & ~                                                          & ~                                                           & \Checkmark                                                             \\ 
\cline{2-23}
                                                                                                                      & Ensembles~                                       & \cite{yamana2017individual,kandula2018evaluation,ray2018prediction,reich2019accuracy,adiga2021all,mcandrew2019adaptively,ray2020ensemble,kim2020covid}                                             & \Checkmark                                                                    & \Checkmark                                                                      & ~                                                         & ~                                                      & ~                                                            & ~                                                                 & ~                                                     & \Checkmark                                                                 & \Checkmark                                                     & ~                                                                    & \Checkmark                                                       & Cty/S/C/R                                          & W                                                           & \Checkmark                                                                 & ~                                                                & ~                                                                    & \Checkmark                                                                  & ~                                                          & ~                                                           & \Checkmark                                                             \\
\bottomrule
\end{tabular}
\arrayrulecolor{black}}
\end{sidewaystable}

\subsection{Organization}
\begin{figure}[htb]
\includegraphics [width = 0.95\textwidth]{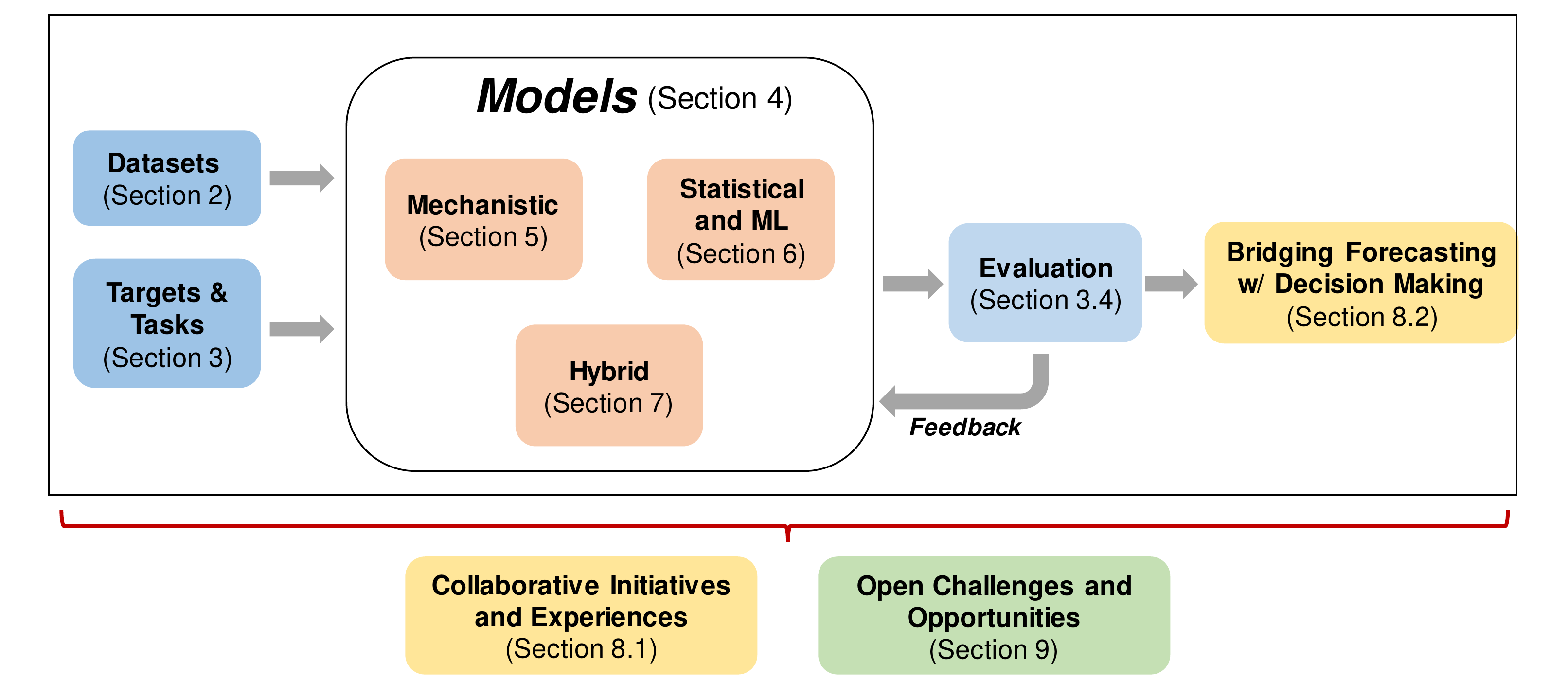}
\caption{\textbf{Survey organization.} 
Section~\ref{sec:data} provides a comprehensive view to data sources utilized to better inform epidemic forecasting.
In Section~\ref{sec:setup}, we describe 
other elements needed before modeling such as forecasting objectives and tasks, along with common evaluation metrics (Section~\ref{sec:eval}). 
Then, we move to forecasting methodologies in Sections~\ref{sec:mechanistic}, ~\ref{sec:stat}, and ~\ref{sec:hybrid}. 
In Section~\ref{subsec:decision}, we review recent work in bridging forecasts and decision-making.
In Section~\ref{sec:ontheground}, we survey initiatives and experiences that leverage these models and discuss the challenges of real-time deployment.
Finally, in Section~\ref{sec:discussion} we discuss major challenges and important open research problems across all of the topics surveyed.  
}
\label{fig:organization}
\end{figure}

This survey is organized in seven sections as depicted in Figure~\ref{fig:organization}. 
In Section~\ref{sec:data}, we discuss traditional and more recent sources of data used for epidemic forecasting. In Section~\ref{sec:setup}, we describe the forecasting setup, which includes defining the forecasting objective and specific tasks. We then address commonly used quantitative evaluation metrics.
Then, we move to forecasting modeling techniques (Section~\ref{sec:modeling_paradigms}), which we classify as mechanistic, statistical and hybrid and further classify them based on their key modeling ideas. 
For each modeling technique, we provide an in-depth review of prominent methods starting with the well studied mechanistic models (Section~\ref{sec:mechanistic}) and then shift towards statistical models (Section~\ref{sec:stat}) that are more flexible in leveraging large and varied sources of data, learn complex patterns from past data and often provide more accurate forecasts. Here, we pay close attention to deep learning innovations that have been a very active research area with multiple 
examples of state-of-art performance. 
In Section~\ref{sec:hybrid}, we describe classes of \emph{hybrid models} that have recently garnered interest and marry the interpretable, theory-grounded long-term modeling capabilities of mechanistic models with the more flexible, accurate and data-driven statistical models.
In Section~\ref{sec:ontheground}, we survey recent `in the trenches' initiatives and experiences that leverage these models for epidemic and pandemic forecasting discussing the challenges of real-world deployment including decision-making informed by forecasts. 
Finally, in Section~\ref{sec:discussion} we discuss major challenges and important open research problems relevant to the various aspects of the epidemic forecasting pipeline.

\section{Datasets}
\label{sec:data}

A diverse set of datasets has been utilized to better inform epidemic forecasting with benefits ranging from early-stage indicators of disease outbreaks to capturing complementary facets that better describe the disease spread dynamics.
Recent efforts, especially motivated by the COVID-19 pandemic, have increased the exploration and availability of novel data sources like smartphones, internet search engines, and satellite images.
In this section, we broadly classify and describe the corpora of datasets we found in prior work.

\subsection{Clinical Surveillance Data}
\label{sec:clinical}
\begin{figure}
\centering
\includegraphics[width = 0.95\textwidth]{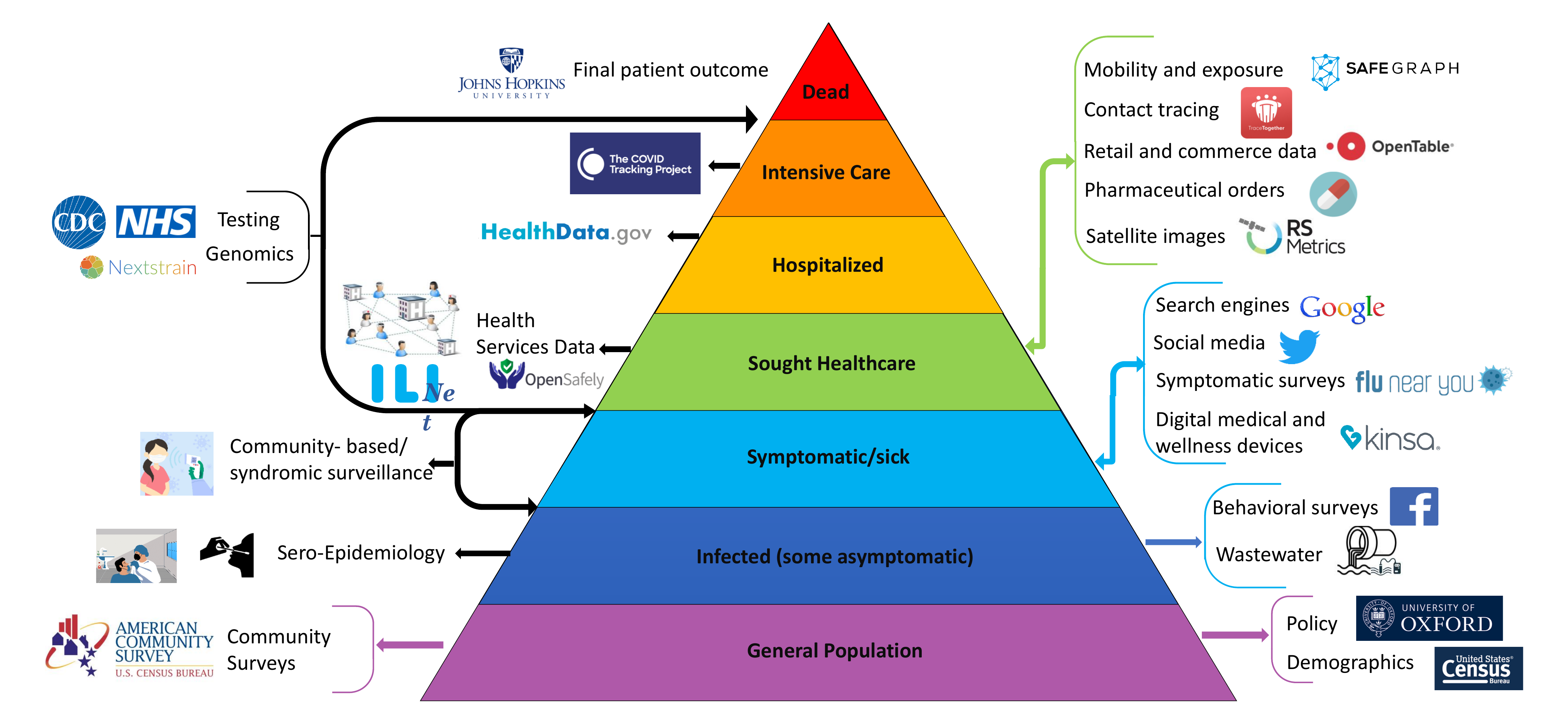}
\caption{
\textbf{Conceptualization of surveillance data sources.} 
The pyramid depicts the several stages of a person in the illness from bottom to top where the area is proportional to the number of people. We connect each of its levels with our proposed classification of datasets used in the literature to inform forecasting models and depict some of their representative examples. Left side of the pyramid: direct surveillance measures of the disease spread. Right: proxy measures of epidemiological aspects of the disease spread. 
}
\label{fig:pyramid}
\end{figure}

These datasets come from clinical information of patients (observation and treatment) by healthcare providers and public entities, thus, they provide firsthand information to perform illness surveillance (see Figure~\ref{fig:pyramid}).

\subsubsection{Line list}
These datasets have been the first ones utilized in traditional epidemiology.
Line lists are individual records that report who, when and where a person was infected, and include the number of persons infected, recovered, and deceased.  
As this is information of public interest, line lists are collected, aggregated, and promptly disseminated by public health entities around the world. For example, the Centers for Disease and Control (CDC) in the US~\cite{cdc_NREVSS}, National Health Services (NHS) in the UK~\cite{uk2020coronavirus}, and state-level public health departments in India~\cite{laxminarayan2020epidemiology}.
An important source has been the CDC's National Notifiable Disease Surveillance System~\cite{cdcnnds}, which aggregates cases from healthcare providers across the US for various diseases including Tuberculosis, Dengue, Herpes, and Botulism.
During the COVID-19 pandemic, the web dashboard from Johns Hopkins University~\cite{dong2020interactive} has been a trusted source for the CDC and the \hub for reporting the number of confirmed cases, recovered people, and deaths. 
Among the epidemiological indicators available, the number of deaths has been the primary forecasting target for the CDC due to its reliability at initial stages of the pandemic~\cite{cramer2021evaluation}. This was also a period when confirmed cases had severe biases, especially due to under-reporting~\cite{castrofino2020influenza}.

A closely related dataset is based on excess deaths estimations provided by the public health agencies like CDC~\cite{cdc_excess_deaths}, which account for the excess in deceases with respect to the expected trends. While studies \cite{cuellar2021excess} have shown that excess deaths is an important epidemiological indicator during epidemics, others found that it was not an informative signal in a data-driven model~\cite{rodriguez_deepcovid_2021}, which may be due to data collection issues such as undercounting~\cite{karlinsky2021tracking}. This suggests further investigation is needed regarding the best ways to take advantage of this information for epidemic forecasting.

\subsubsection{Testing}
It is derived from laboratory virological tests, which involve examining a sample from a substance from the patient's body.
Virologically confirmed cases are direct measures of the disease spread since the negative and total number of cases may be used to understand the social and policy considerations. %could cite
For example, an increase in reporting testing results may reflect health-care workers' and local government's efforts to mitigate disease spread~\cite{brooks2015flexible,rodriguez_deepcovid_2021}. 
While the recent COVID-19 pandemic has made virological testing widespread, this has not been the case for diseases like influenza and Ebola, where tests are conducted only for people who meet specific criteria based on risk factors and symptoms~\cite{cdc_ebola_testing,cdc_flu_testing}.
For example, ArboNET is the US' national surveillance system for arboviral diseases (e.g., the West Nile virus) that passively collects data from public health entities and laboratories. Performing diagnostic testing and even the reporting of positive cases is done at the discretion of clinicians and laboratories~\cite{lindsey2012state}. 
There are several data issues to take into consideration when working with testing data.
Testing data reported by public entities may require non-trivial curation to be ready for use in forecasting models~\cite{kraemer2021data}. Additionally, issues such as data revisions and reporting delays~\cite{dong2020interactive,altieri2021curating} can cause anomalies that may confound models. While these issues are notorious in COVID-19 testing data, this is not exclusive to this type of data, and we further elaborate this in Section~\ref{sec:discussion}.

\subsubsection{Health services records}
These datasets are collected from service records of people who seek medical care at healthcare providers.
They can be divided into outpatients (patients not hospitalized) and inpatients (hospitalized patients). 
A prominent outpatient-based indicator is the influenza-like-illness (ILI) counts, which is collected 
by the CDC from the US Outpatient Influenza-like Illness Surveillance Network (ILINet) and aggregates from healthcare providers across all US states and territories.
It measures the percentage of healthcare seekers who exhibit influenza-like-illness symptoms, defined as "fever (temperature of 100°F/37.8°C or greater) and a cough and/or a sore throat without a known cause other than influenza"~\cite{cdc_flu_surveillance}.
ILI counts are usually used as the main indicator of the influenza burden as it is the most trusted source of information.
Therefore, they were chosen as the forecasting target of CDC FluSight challenges~\cite{biggerstaff2016results,biggerstaff2018results}.
While similar networks have been put in-place worldwide~\cite{european2014data,biggerstaff2020coordinating}--for instance, the Pan American Health Organization (PAHO)-- they may differ in their ILI definitions and data collection practices~\cite{chakraborty2018know}.
Inpatient-based datasets are not as widely available. For example, both the Influenza Hospitalization Surveillance Network (FluSurv-NET)~\cite{cdc_flusurvnet} and the COVID-19-Associated Hospitalization Surveillance Network (COVID-NET)~\cite{cdc_covidnet} gather data only from 14 states, with some states' population being severely underrepresented (<20\% of its population for California). 
A recent effort led by the US Department of Health \& Human Services (HSS) aims to compile data from multiple sources to estimate COVID-19 hospitalizations~\cite{hhs_hosp}, an  indicator being used as a forecasting target in the \hub.

\subsubsection{Electronic records}
Electronic health records (EHR) are more detailed datasets which contain patient information at an individual-level.
This data has been extensively used in clinical studies, however, their utilization for public health forecasting is still in its infancy. 
Some recent works have studied the clinical factors associated with COVID-19 using EHRs, like OpenSAFELY from NHS~\cite{williamson2020opensafely}. These types of studies open a door towards further exploration of EHR datasets via a transparent and safe system built around privacy.
Other examples include \citet{zhang2017data}, who use EHR data along with contact networks to derive effective intervention for controlling epidemic spread.
Other works propose using medical claims data, which may be more readily available than line lists as syndromic indicators~\cite{claims_delphi}. Frequencies of relevant medical diagnosis codes extracted from medical claims data are usually used in such studies~\cite{Gao2021STANSA}. There are also studies which use EHR to predict mortality at individual-level such as~\citet{schwab2021real}. 

\subsection{Digital Surveillance Data}
With the widespread use of mobile devices and innovation in digital communication, electronic surveillance methods provide real-time access to useful markers for predicting disease incidence and outbreaks leveraging crowdsourcing without requiring significant human interventions. They can complement clinical data 
and provide reliable outbreak detection systems from anonymized digital data streams 
and impact real-time decision-making~\cite{tarkoma2020fighting}.
We discuss the varied sources of data and their efficacy.

\subsubsection{Social media}
Social media aggregates public opinion, commentary and news on a wide range of topics including the influence of diseases over time and across different regions.
Leveraging a large amount of multi-modal data from social media is a great source of real-time electronic surveillance at scale. Twitter posts are an important source of surveillance data. \citet{culotta2010towards} was one of the first papers to leverage Twitter posts by tracking tweets with specific flu-related keywords and using volumes of tweets as features. This was followed by multiple works that extract text features from posts \cite{lampos2010flu,lee2013real,lamb2013separating,paul2011you,chen2014flu,abouzahra2021twitter,masri2019use} for various diseases including Avian flu and Zika. \citet{kanhabua2013understanding} studied the use of statistics measuring diversity in tweets as a means to obtain a less noisy signal for syndromic surveillance. They observed a high correlation between the temporal diversity of tweets and the event of a disease outbreak.
While social media has been shown to be an effective data stream, misinformation and rumors may create significant noise, especially in early stages of disease outbreaks \cite{rosenberg2020twitter,oyeyemi2014ebola,jin2014misinformation}. As in other domains, effectively leveraging information from social media remains an open research problem.

Other sources of social media that specifically cater to health-based discussions have been explored. For instance, \citet{keller2009use} introduced HealthMap, a database of RSS feeds with health-related content, leveraging a web scraper that collected thousands of RSS feeds 
on medical articles. They parsed the HTML structure of the documents to extract information such as date, headline, summary and location. 
However, there were reported drawbacks to this process including noise in document format, the efficacy of unpublished works, discrepancies in the accuracy of systems for different diseases either due to human expertise or the system's ability to detect for a given disease.

\subsubsection{Online search and website logs}
Tracking search queries from popular search engines is another popular real-time electronic surveillance method. These methods build on top of databases of disease-related search queries from ad-hoc search engines like Yahoo \cite{polgreen2007use}, Google \cite{ginsberg2009detecting} and Baidu~\cite{liu2020machine}. 

Using search queries as well as metadata about region sources such as IP address, \citet{polgreen2007use} provide spatially distributed signals that are correlated with ILI values. Google Flu Trends (GFT) was a proprietary method that used select keywords to predict future flu incidences. However, they were misleading during the H1N1 pandemic in 2009 and in later years~\cite{ginsberg2009detecting,yang2015accurate} and soon after it was discontinued.
Similarly, \citet{milinovich2014using} developed systems to track multiple diseases including hepatitis, Dengue and tuberculosis.
Other works such as~\cite{mcdonald2021can,lampos2021tracking,rabiolo2021forecasting} similarly used online searches for COVID-19 forecasting and hotspot detection.
Recently, Google released datasets based on searches of symptoms of multiple diseases~\cite{bavadekar2020google} and vaccine-related~\cite{bavadekar2021google}. Instead of relying on keywords, they now use recent ML techniques to have a more precise understanding of the user's intent. Some insights available are trends in symptoms like fever and insomnia, vaccination intent and vaccine side effects.

A more curated source of search queries may come from specialized search engines like UpToDate and Wikipedia.
UpToDate is a database used by over 700,00 health practitioners around the world and filtered the relevant search query volumes for relevant terms and showed better performance compared to GFT~\cite{santillana2014using} .
Works such as \cite{mciver2014wikipedia,priedhorsky2019estimating,zimmer2018use} leverage Wikipedia articles relevant to diseases 
and track their search query volume. 
Such data, however, are not geographically fine-grained and need further statistical~\cite{priedhorsky2019estimating} or linguistic~\cite{mciver2014wikipedia} methods to extract spatially related features.

\subsubsection{Symptomatic online surveys}
While the practice of surveying a small sample of populations for disease symptoms is a well-known surveillance method from the past,
advances in mobile devices have made this approach more scalable and effective. Volunteers fill up a questionnaire where they are asked if they or anyone in their family is experiencing symptoms of a disease, and data from these participants are aggregated as a signal for flu incidence. Such programs are organized around the world such as 
Flu Near You (FNY) in US \cite{smolinski2015flu} (see Figure \ref{fig:data2}c) and similar in other countries
~\cite{koppeschaar2017influenzanet,moberley2019flutracking}
and Dengue na Web in Brazil~\cite{wojcik2014public}.
Public engagement is key to make these surveys a success.
An average of 250 weekly participants from each %HHS 
region was sufficient to produce a strong correlation of FNY survey results with actual ILI reports~\cite{baltrusaitis2018comparison}. 
However, the correlation varies greatly across regions and there is unequal participation across regions, with vulnerable demographics such as smokers and diabetics significantly underrepresented in the surveys \cite{koppeschaar2017influenzanet}. The number of participants also varies across time with more participation during historic peak flu incidence months. 
Therefore, all these factors should be considered during pre-processing and using symptomatic survey data.
Datasets for COVID-19 include the US COVID-19 Trends and Impact Survey~\cite{salomon2021us}, which randomly invited Facebook users to participate in a survey.
To explore the utility of such datasets, the COVID-19 Symptom Data Challenge~\cite{symptom_fb} tasked teams to present novel analytic approaches that could showcase improved situational awareness and predictions with Facebook's symptomatic surveys.
The winning solution showed improvements in real-time forecasting with deep learning models (especially in short-term predictions and in anticipating trends) and the runner-up discovered causal relations between policies and symptom data.

\subsubsection{Medical and wellness devices}
Digital health devices can be a valuable source of large scale surveillance data that minimizes the direct involvement of health care agencies. They record patients' vital readings like temperature, sleep cycle, heart rate, etc. that provide useful indicators towards future incidence. \citet{miller2018smartphone} leveraged body temperature data recorded by Kinsa smart ear thermometers that record the patient's temperatures in a mobile app. This was used to collect average temperatures, the number of times individuals experienced fever, information useful for flu and COVID-19 forecasting~\cite{rodriguez_steering_2021,rodriguez_deepcovid_2021}. \citet{radin2020harnessing} used health data collected from Fitbit devices. Specifically, they collected anonymized data of heart rates to create a model for flu prediction. Digital devices help provide individual real-time clinical data without the need for direct contact with public health workers and agencies. 
However, the data collected doesn't cover all socioeconomic categories equally. It may also increase privacy risks of sensitive health data which would require new investments in security infrastructure~\cite{alwashmi2020use}.
Another source of data are point-of-care devices.
For instance, Quidel Corporation provides influenza testing data from a network of machines called the Influenza Test System~\cite{leuba2020tracking}. 
BioFire's multiplex molecular diagnostic network is a reporting system that has been explored to be used as a novel data source for disease surveillance~\cite{meyers2018automated}.
In contrast to reports from CDC, this type of data has higher spatial granularity (e.g., at zip-code level) and is often available in real-time.

\subsubsection{Satellite images}
\begin{figure}[h]
    \centering
    \small
    \begin{tabular}{cc}
    \includegraphics[width=0.55\linewidth]{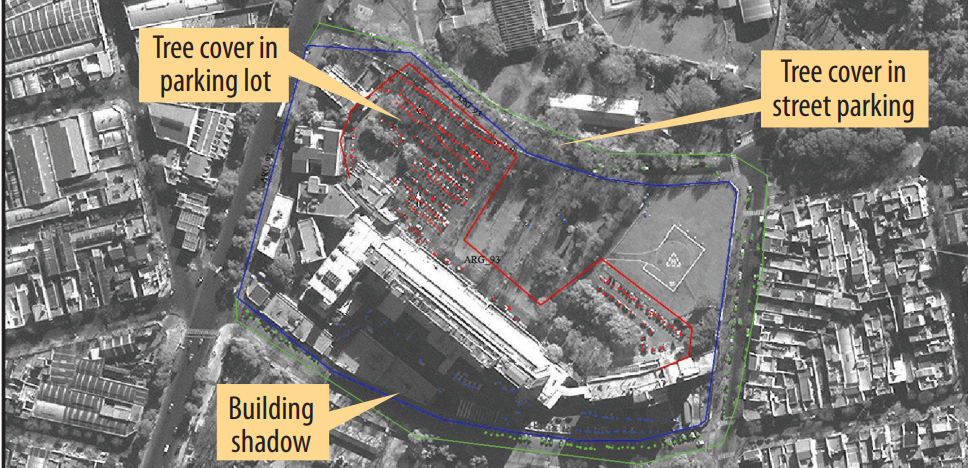} &
    \includegraphics[width=0.15\linewidth]{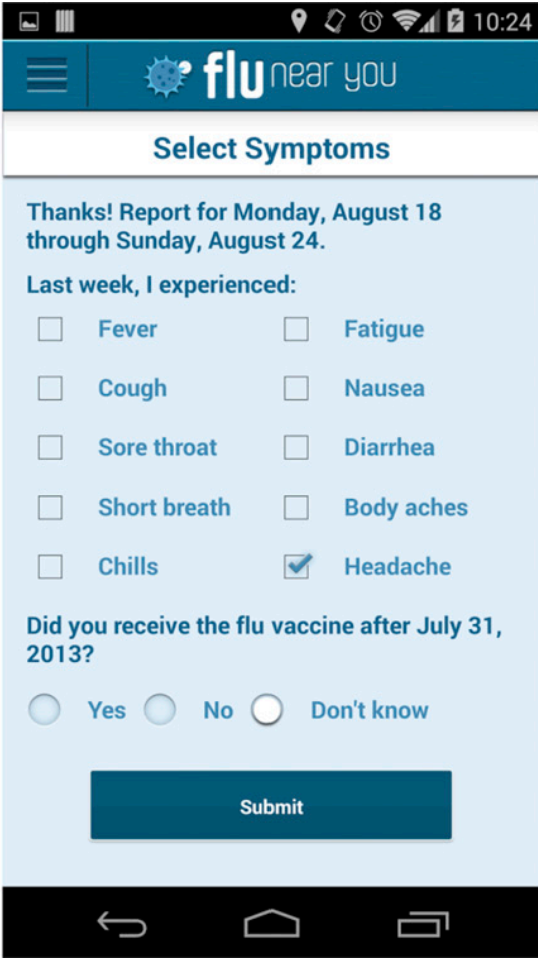} \\
    (a) Satellite image data  & (b) Online surveys \\
  \end{tabular}
    \caption{\textbf{Examples of digital surveillance data sources.} (a) Satellite image data has been utilized to quantify hospital visits in disease outbreaks \cite{butler2014satellite}.
    (b) Interface for FluNearYou (FNY) survey used from crowdsource reporting of symptoms \cite{smolinski2015flu}.}
    \label{fig:data2}
\end{figure}

Image from remote-sensing satellites can also be leveraged as a source of surveillance.  \citet{nsoesie2020analysis} and \citet{butler2014satellite} used images of parking lots of hospitals from RSmetrics, a company that leverages satellites to collect image data from across the world to track COVID-19 and influenza outbreaks. They found that satellite data can be used to extract images at specific times of the day across multiple sensitive locations like hospitals. Such images are later pre-processed for extracting a measure of hospital visits. 
Other studies such as \cite{ford2009using,patel2020mosquitoes,generous2017forecasting} leverage satellite data to monitor various environmental patterns including temperature, rainfall, and chlorophyll levels of crops to track vector-borne illnesses like cholera, hantavirus, and malaria.
While the satellite data can be extracted in real-time at small geographical granularities, it suffers from imperfections and noise due to external factors like clouds, tree covers, construction activities that make it hard to define contours to survey, and others (Figure \ref{fig:data2}a). Data collection from satellites is also constricted to different extents by privacy and security laws in different regions that may introduce irregularity in data collection.

\subsubsection{Retail and commerce data}
Data from commerce and retail that depict consumer behavior at scale can also serve as useful electronic surveillance data.
The OpenTable dataset that tracks reservations at multiple restaurants around North America where disruption and decrease in reservations could signal disease outbreaks~\cite{nsoesie2014guess}. 
While there is a strong correlation between an increase in ILI activity and an increase in the availability of reservations, data may be noisy and not fully informative since the changes in reservations could be influenced by multiple factors like seasonal changes, social unrest or changing dining preferences among studied restaurants.
\citet{sadilek2018machine} curated a dataset by combining the user search data with location data collected by Google to identify, based on the search queries related to food-borne illnesses, the source restaurant for the infection.
Another line of work uses retail data from supermarkets, where features obtained by some simple classification of the items were found useful in disease forecasting~\cite{miliou2021predicting}. 

\subsubsection{Pharmaceutical records}
Data from pharmaceutical orders (anonymized prescriptions and longitudinal data) have been previously explored in some works. \citet{jainanalysis} obtained data from Tata-1mg, an online Indian healthcare brand with 50M monthly active users who search and order medicines, and found that several search tags had a strong correlation with reported COVID-19 case numbers and suggest that these records may lead to a novel early warning indicator.
A National Retail Data Monitor was previously curated by CDC \cite{wagner2004national} by collecting sales data from multiple major health-care related retailers with the goal of detecting disease outbreaks. Another line of work tracks data from drug prescriptions and general prescription practices across regions~\cite{martins2021prescription,renny2021temporal}. These works also found that such data can be useful in modeling opioid epidemics to help inform substance abuse prevention.

\subsection{Behavioral Data}
Tracking specific behavioral markers at the individual level such as mobility patterns, adherence to interventions and recommendations, and vaccination adoption can be a useful source of surveillance data to forecast future disease incidence \cite{betsch2020behavioral,lu2021collectivism,gollwitzer2020partisan}. Here we discuss some relevant sources of behavioral data.

\subsubsection{Mobility and Exposure}
Mobility data tracks aggregated movement of individuals within a geographical region or across different regions.
They provide a useful indicator on the ratio and trends of disease spread.
A closely related metric is exposure, which measures the density of people in a location (like supermarkets, movie theaters, airports). This is usually measured as the number of overlapping devices at a location at a given time.
The collection of both mobility and exposure logging the location of personal devices like smartphones and watches \cite{wesolowski2012quantifying}. For instance, Google Mobility Reports \cite{aktay2020google} uses anonymized location history when people use Google services that leverage GPS, which capture mobility patterns at country, state and county levels. Similar initiatives are being conducted by Apple~\cite{miller2020mobility}, Facebook~\cite{fb2020mobility} and others~\cite{chevalier2021measuring,novid2021} using real-time location data.
Alternatively, SafeGraph~\cite{safegraph} %a dataset used in \cite{chang2021mobility} 
leverages GPS data to measure visitor counts, dwell times, distance traveled to locations of interest and provide anonymized data for modeling mobility.
Another line of work uses call records data from mobile devices as source of mobility data to model spread of diseases like dengue~\cite{wesolowski2015impact} and COVID-19~\cite{buckee2020aggregated}.
Mobility data can be also measured via air travel patterns between airports, from which multi-scale mobility networks can be constructed \cite{balcan2009multiscale}.
Regarding exposure data, \citet{chevalier2021measuring} used GPS data from multiple smartphones to measure overlapping exposure to other individuals in important locations to provide a Device Exposure Index (DEX) for each county and state. 

Mobility and exposure data, in general, cover a wide range of demographics and provide information about large scale movement patterns. However, they also raise privacy concerns and security risks to prevent leaking of personally identifiable location data~\cite{krumm2009survey}.
Another issue may be representativeness because the data can be skewed due to factors such as age (towards younger population), income, race and education levels~\cite{chevalier2021measuring,wesolowski2013impact}.

\subsubsection{Contact tracing}
Contact tracing involves detailed tracking of individuals who are exposed to carriers of disease and has been a standard public health practice for mitigating many disease outbreaks like HIV~\cite{rutherford1988contact}, Ebola~\cite{swanson2018contact}, tuberculosis~\cite{world2012recommendations} and more recent COVID-19~\cite{ahmed2020survey,swain2021wifi}. This can allow epidemiologists to identify the source of an outbreak and forecast its future impact~\cite{eames2015six}. While older methods leverage health care workers interviewing suspected patients and their families~\cite{swanson2018contact}, recent digital contact tracing methods provide more updated large scale contact-tracing data. 
This can be obtained from smartphones using either Bluetooth or GPS technology and can be organized via a centralized structure, like the TraceTogether app by Singapore Government \cite{cho2014learning} or decentralized peer-to-peer networks like in Vietnam \cite{li2020covid}. 

The use of digital contact tracing in epidemic forecasting is still in its infancy.
Recent works include looking into 
small spatial scales such as on a university campus. For instance, \citet{swain2021wifi} leveraged WiFi records of students who visited different locations and classrooms to build a contact network and model spread of COVID-19 and proposed selective lockdown policies that were effective while avoiding complete closure.
Possible extensions may include incorporating demographic and past clinical data to provide more accurate and effective estimates of infectiousness~\cite{bengio2020predicting}.
However, there are some challenges in the use of this novel data source~\cite{swanson2018contact,li2020covid}. First, there are privacy concerns regarding location data collected by contact tracing apps, especially using centralized systems. Involuntary tracing methods such as using security footage or recording financial transactions also raise greater security and privacy concerns. Second, most voluntary contact tracing methods don't have massive adoption. For instance, the government organized TraceTogether app was only used by less than 20\% of the population. There are also problems with location measurements due to inaccuracies of Bluetooth and GPS technologies \cite{cho2014learning}. In addition, it should be considered that in some cases 
introducing contact tracing may have undesirable effects. For instance, it may reduce the adherence to restrictions imposed to control the outbreak~\cite{kleinman2020digital}.

\subsubsection{Behavioral surveys}
Measuring attributes of social behavior concerning public health adherence can be a useful marker of forecasting~\cite{rebeiro2021impact}.
Facebook, in collaboration with CMU~\cite{delphi2021survey}, %and UMD~\cite{umd2021survey} 
conducted an online survey leveraging social media reach of their platform to survey important markers like mask-wearing, vaccine hesitancy, lockdown adherence, recent travels, attendance to indoor events, mental health, testing and financial conditions of individual Facebook users and their community. 
Phone call surveys can also be used in developing nations without access to internet~\cite{raza2013job}.
Similar to the symptomatic online surveys previously introduced, behavioral surveys can also suffer from disproportionate representation and varying participation across regions and times.

\subsubsection{Demographics} %  and co-morbidity
Demographics data of a population such as age, sex, ethnicity, and economic indicators are important aspects to understand the heterogeneity of the population and the possible contact patterns among communities. This information is especially useful in constructing mechanistic metapopulation models (in defining sub-populations) and agent-based simulation models (in further defining the agents and their contact networks)--as we further elaborate in Section~\ref{sec:mechanistic}. 
Some sources are the World Bank~\cite{world_bank} for country-level statistics and Census bureaus~\cite{us_census} for more granular data.
On the other hand, note from Table~\ref{tab:datasets} that 
there has been little exploration of the use of this type of data in statistical models.

\subsection{Genomics Data}
Genomic epidemiology, which links pathogen genomes with associated metadata to understand disease transmission, has become a key component of outbreak response~\cite{HILL20211038}. Many viral strains show seasonal patterns and constantly evolve with changing weather and medication~\cite{chakraborty2014forecasting, brooks2020pancasting}. Thus, leveraging this information should be useful for forecasting models. \citet{hadfield2018nextstrain} introduced the NextStrain dataset that collects the viral genomes of various pathogens and tracks their mutation over time as they spread to different regions. Other publicly available repositories of genome sequences include GSAID \cite{gsaid2021}, GenBank \cite{benson2018genbank}, COG-UK \cite{covid2020integrated}. There has been very recent works  that leverage genomics data of pathogens from local sources \cite{paredes2021associations,ricotracing} associated with influenza, COVID-19, dengue, Zika, etc. to design models \cite{tang2014stopping,stoddard2021using, cramer2022nextstrain}.
With advances in genome sequencing, phylodynamics datasets show great promise for use in novel outbreak detection~\cite{grad2014epidemiologic, davies2021estimated}, estimating epidemic parameters such as reproduction number~\cite{stadler2012estimating}, and predicting useful virus phenotypes that may inform the medium, the rate of spread, and the population's immune response~\cite{grad2014epidemiologic}, all of them of epidemiological significance.

\subsection{Environmental Data} 
Indicators of environmental factors can also be effective sources of surveillance data. These are typically not easily influenced by change in human behavior but can still track or impact the spread of pathogens. In this subsection, we study different sources of environmental data that have been used for disease forecasting. 

\subsubsection{Wastewater}
Analyzing wastewater for presence of biological markers of pathogens is a useful measure of community-wide affliction of disease, with the advantage that results of wastewater analysis have the potential to predict an outbreak earlier than traditional epidemiological indicators~\cite{larsen2020tracking}. This method was particularly used for diseases that travel through fecal matter like the polio virus \cite{asghar2014environmental}. However, other diseases like influenza \cite{heijnen2011surveillance} and COVID-19 \cite{peccia2020measurement} can also be surveyed by this method. Noting that many viral RNA, including the one of COVID-19, is stable to temperature fluctuations, \citet{peccia2020measurement} diluted the wastewater sludge and detected concentration of viral RNA via PCR tests. They found a strong correlation between RNA concentration and reported COVID-19 cases. \citet{larsen2020tracking} noted that wastewater based epidemiology has great potential since it is cost-effective and can be implemented even in places that lack good health care systems. It is also unbiased towards traditional surveillance methods (for example, reported cases are biased towards the number of tests conducted).

\subsubsection{Meteorological}
Climate and weather markers are influential markers for predicting the onset of seasonal diseases like flu, especially in tropical regions~\cite{soebiyanto2010modeling}. Due to their easier availability, weather data are used for disease forecasting in regions with sparse traditional surveillance data. For instance, \citet{tamerius2013environmental} studied local climate variables around the tropical regions, especially specific humidity as a useful indicator of seasonality of influenza incidence. Other works like \cite{venna2018novel,soebiyanto2010modeling} also included important climatic indicators in their statistical forecasting models. Weather and climate data, while not fully informative of all factors of an epidemic, provide year round reliable markers that are useful for epidemic modeling.

\subsubsection{Zoonotic}
Multiple infectious diseases are born in animals and/or primarily transmitted to humans by them~\cite{heesterbeek2015modeling}. % human-to-human transmissible
Therefore, identifying and obtaining tracking data from the hotspots of wildlife where zoonotic diseases are more likely to appear is very relevant for early detection \cite{allen2017global}.
In addition, data about movement of disease vectors (like mosquitoes) is informative to take action~\cite{grubaugh2019tracking}
as demonstrated for malaria, dengue and Zika~\cite{benelli2016declining}.
Recent efforts in increasing data collection include the Microsoft Premonition project~\cite{premonition}. They are developing a platform that leverages genomics robotic sensing, AI, predictive analytics and cloud systems to monitor disease-carrying animals such that we can have early warning signals of possible outbreaks.

\subsection{Policy Data}
Various policy measures implemented by government agencies with the goal of reducing outbreaks or mortality effects future forecasts. Therefore some mechanistic models~\cite{kain2021chopping} and hybrid models~\cite{qian2020and} explicitly design their models to be sensitive to such changes. For example,
as data in their hybrid model, 
\citet{qian2020and} used policy markers like school and workplace closure, stoppage of public transport, travel restrictions,
and investment in emergency health care and vaccines collected from the Oxford
COVID-19 Government Response Tracker (OxCGRT)~\cite{hale2020variation}.
However, it should be noted that the extent of available data and level of adherence to these measures is variable across different countries and regions~\cite{fischer2021mask}.

\section{Task, Targets and Evaluation Metrics}
\label{sec:setup}

In this section, we describe the most popular epidemic forecasting tasks and how model performance is evaluated in these tasks.
We start by discussing the differences between the so-called `projection' and `prediction' setups - both very popular.
Next, we formally describe the various tasks related to epidemic predictions and finally discuss the various metrics used to evaluate performance along different tasks.

\subsection{Projections vs. Predictions}
\label{sec:proj_vs_pred}
Forecasts can be divided into projections (for multiple scenarios) and predictions (for the most likely scenario)~\cite{petropoulos2020forecasting}.
In this survey, we focus on predictions because it is where most data-driven work has been proposed, but we briefly touch upon projections when relevant to decision making.
Nevertheless, several of these prediction techniques can be adapted to projections~\cite{bunn1993forecasting}.
We give a very high-level summary  here and then later discuss their use in decision making in Section~\ref{subsec:decision}. We again make some final remarks about their differences in Section~\ref{sec:discussion}.

\subsubsection{Projections}
Projections are forecasts made for specific scenarios with the intent of understanding the epidemic outcome on each of them~\cite{borchering2021modeling,probert2018real}. 
To evaluate several scenarios, models have some mechanisms or assumptions to incorporate changes based on the scenarios that they want to evaluate.
For example, a non-pharmaceutical intervention such as mask mandates is going to reduce the infection rate, but this rate  will vary with the public adoption and mask efficacy. Therefore, to provide projections for several levels of adoption rate, a model may change parameters related to infection rates to represent each mask adoption rate.

\subsubsection{Predictions}
In contrast to projections,
predictions are forecasts for the most likely scenario (a `default' scenario), i.e., the best possible guess of the evolution of the future. Works using mechanistic models make explicit assumptions on the parameters that define this default scenario. However, this may not always be reliable due to changing dynamics of an epidemic and evolving priors about parameters such as infection rates, asymptotic spread, etc, as it is commonly the case for COVID-19. In contrast, statistical models implicitly infer the default scenario from historical data. This can be directly observed in some models to directly map the current weekly patterns to historical patterns that indicate similar default scenario~\cite{adhikari2019epideep, kamarthi2021epifnp}.

In our survey, most work in epidemic prediction has been focused on short-term predictions, typically up to 4 weeks in the future. Naturally, understanding the limits on the predictability of infectious diseases has been an intriguing question.
Some works framed the question as `are long-term predictions even possible?' and pointed to some theoretical~\cite{rosenkrantz2022fundamental} and practical~\cite{scarpino2019predictability} limits of forecasting epidemics including in real-time situations~\cite{reich2021onpredictability}.
In Section~\ref{sec:discussion} we discuss how a principled incorporation of data streams might help our understanding of predictability.

\subsection{Prediction Targets}
\label{sec:targets}

To define a forecasting problem, we need to specify what is the predictive target of interest and the spatial and temporal scales for it. 
The usual predictive targets are fractions of symptomatic outpatients~\cite{cdc_flu_surveillance}, hospitalizations~\cite{rodriguez_deepcovid_2021}, mortality~\cite{kamarthi2021back2future} and other epidemiological indicators as noted in Section~\ref{subsec:epi_indicators}.
We are often interested in predicting the observed occurrence of new incidences (in a given period of time) of the forecasting target.
It is important to note that early in the COVID-19 pandemic, multiple papers used cumulative counts instead of incidence, but incidence predictions were later preferred in CDC dashboards~\cite{covid_forecasting_cdc} and CDC-authored evaluations~\cite{cramer2021evaluation}.

The targets are defined across two dimensions. 
The \emph{spatial scale}, also known as spatial granularity, is the geographical region to which the forecasts are circumscribed.
For instance, the number of cases in a country.
The common spatial scales are national, region/state/province, and county/city, being the latter one the less common due to the data sparsity at such low granularities.
The \emph{temporal scale}, also known as temporal granularity, is the period of time at which the epidemiological target is aggregated when calculating the new incidences, which is often weekly or daily.

\subsection{Prediction Tasks}
\label{sec:tasks}

Once we have picked a target, there are a variety of tasks that are deemed relevant by domain experts for decision making and preparedness \cite{johansson2019open,cdcflusight}. In this section, we describe the main tasks and their impact on limiting an outbreak or epidemic.

\subsubsection{Real-valued prediction tasks}
First, we present the tasks involving the prediction of real-valued important indicators of the unobserved or future epidemic trajectory. 

\begin{figure}[htb]
    \centering
    \small
      \includegraphics[width=.9\linewidth]{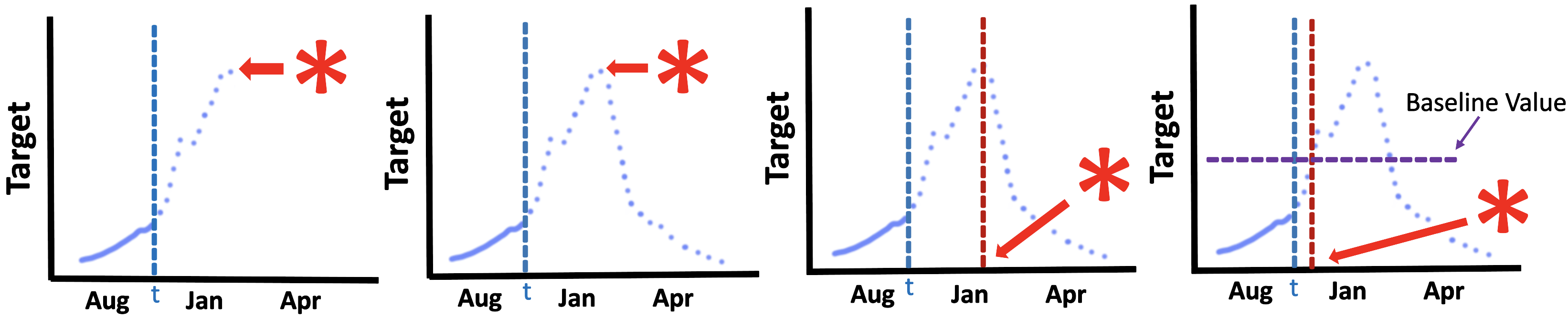}
    \begin{tabular}{cccc}
    (a) Future incidence \qquad\qquad & (b) Peak intensity \qquad\qquad & (c) Peak time \qquad\qquad & (d) Onset time \qquad\qquad
    \end{tabular}
    \caption{\textbf{Examples of epidemic prediction tasks.} The red asterisk depicts the characteristic of the epidemic curve to be predicted and the blue dotted line represents prediction week $t$. As described in Section~\ref{sec:tasks}, (a) and (b) are examples of real-valued tasks while (c) and (d) are event-value tasks. 
    }
    \label{fig:tasks}
\end{figure}

\paragraph{Task 1: Future incidence prediction}
Future incidence prediction of a target, depicted in Figure~\ref{fig:tasks}a, is a common task in epidemiological modeling since it describes the future direction of a pandemic. 
These predictions can be for short-term predictions (up to 4 or 5 weeks ahead in the future) or long-term predictions (5+ weeks ahead). As we forecast farther into the future, the accuracy of the models quickly decreases~\cite{reich2019collaborative, reich2019accuracy, cramer2021evaluation}.

\paragraph{Task 2: Nowcasting}
Nowcasting deals with prediction of targets up to the current week, sometimes also termed as `predicting the present'~\cite{choi2012predicting}.
Due to reporting errors and delays that can be caused by various factors such as testing artifacts, modelers usually have to deal with \emph{right truncation} e.g. they have access to accurate target values only up to 1-3 weeks in the past \cite{osthus2019even}. 
Therefore, nowcasting helps provide a more accurate measure of important markers. \citet{osthus2019even} also showed that using output of a fairly accurate nowcasting model as input features for forecasting models can improve their accuracy. 
Accurate forecasting models can also be directly used for nowcasting, using past weeks' features to predict for current week. 
A general strategy for other nowcasting methods involves relying on more accurate and recent digital features to refine current estimates~\cite{osthus2019even, farrow2017human, lampos2015advances, nunes2013nowcasting}. Some nowcasting methods model reporting delays and errors from past data to learn a good correction mechanism~\cite{stoner2020multivariate,gamado2014modelling,farrow2016modeling}. However, these methods are specific to nowcasting and can not be applied for forecasting future incidence.

\paragraph{Task 3: Peak intensity prediction}
Peak intensity is defined as the highest value of the forecast value across the epidemic season or wave (Figure \ref{fig:tasks}b). For instance, for the flu forecasting task, CDC defines peak intensity on wILI values. Models typically predict the peak intensity throughout the prediction season. Since a season's epidemic curve can have multiple local peaks, \citet{reich2019collaborative} observed that for most models it is very hard to predict peak intensity before the occurrence of actual peak week with some models overestimating the values \cite{kandula2018evaluation} or providing low confidence predictions~\cite{reich2019collaborative}.

\subsubsection{Event-based prediction tasks}
This set of tasks involve predicting important stages of an outbreak or forecasting season.

\paragraph{Task 4: Onset prediction}
Onset week, as is defined by the CDC, is the first week where
the target is above a defined baseline value for at least $x$ consecutive weeks (Figure \ref{fig:tasks}c), being $x=3$ for flu~\cite{biggerstaff2016results}. 
Anticipating the onset can inform vaccination efforts, preparedness tasks like staffing, inventory and treatments, and communication plans to the public to take preventive measures~\cite{kandula2018evaluation}.
Fortunately, onset week seems to be one of the easier tasks to forecast in seasonal diseases like flu: when forecasted with three varying model approaches, onset produced the highest cumulative log score among all the tasks \cite{kandula2018evaluation}. It was also found that some models have the capabilities of accurately predicting onset 6 weeks in advance at the state level \cite{pei2018forecasting}.
Similarly, some works also predict the end of prediction season~\cite{reich2019accuracy} (i.e., the first week below a threshold).

\paragraph{Task 5: Peak time prediction}
This task aims to predict the global maximum of the epidemic curve for an epidemic season or wave (Figure \ref{fig:tasks}d). By forecasting an accurate peak prediction before its occurrence, we may help hospitals better aggregate their staff and resources for a surplus of patients, and also plan for an increase of vaccinations. Peak time predictions can also be useful for informing us of non-seasonal patterns and even help with anticipating epidemics. For instance, several works accurately predicted peak week during the unprecedented H1N1 pandemic \cite{towers2009pandemic, chao2011planning}. % at the state and county level.

\subsubsection{Estimating epidemiological indicators}
\label{subsec:epi_indicators}
While previous tasks predict observable characteristics of epidemic curves, some works aim to predict widely used interpretable epidemiological indicators like  final size of the epidemic.

\paragraph{Task 6: Reproduction number}
Reproduction number, denoted by $R_0$ is the estimated number of secondary infections caused by a single infected person. Thus it is a measure of the extent of the initial spread of disease and a widely-used epidemiological indicator. Multiple works have studied estimation of $R_0$ for many diseases like influenza \cite{biggerstaff2014estimates}, tuberculosis \cite{ibrahim2013mathematical} and COVID-19 \cite{alimohamadi2020estimate} including some works that forecast time-varying reproduction number into the future \cite{ghamizi2020data}. There has been some interest in the closely related generation interval estimate as well~\cite{metaculusomicron2022}. 

\paragraph{Task 7: Final size and attack rate}
Final size measures the total fraction of the population that is infected during the entire period of an epidemic or pandemic. Similarly, attack rate measures the fraction of the population infected in a specific time interval. Most methods that estimate these values assume a mechanistic model to capture the dynamics of the entire epidemic~\cite{house2013big} including those that factor in changes in parameters due to various interventions~\cite{brauer2019final}. Final size is not often used in seasonal diseases; instead it is used to model the overall impact of short-term outbreaks and pandemics like in the case of Ebola~\cite{gaffey2018application,johansson2019open} and early phases of COVID-19~\cite{batista2020estimation}.

\subsection{Evaluation}
\label{sec:eval}
To understand how actionable and reliable forecasts are in informing decision making, several quantitative metrics have been proposed. As each evaluation metric penalizes error differently, considering the requirements of health agencies and decision-makers is important in the selection of evaluation metrics~\cite{chakraborty2018know}.
There are multiple metrics used to evaluate epidemic forecasts, which can be classified based on the point forecasts (also known as single-valued forecasts) and probabilistic forecasts (which also quantify uncertainty via prediction intervals).

\subsubsection{Real-time forecasting and evaluation}
\label{subsubsec:realtime_eval}
\emph{Real-time forecasting} of epidemics aims to inform public health officials and the public of what to expect for the future given all the currently available information. The available information at a given point in time may present some issues as noted in Section~\ref{sec:data}, e.g., prediction targets like testing data often present errors and reporting delays, which are later corrected (revised).
Therefore, this should be considerate when assessing models' performance. Indeed, previous studies found models using revised data for training and evaluation may lead to different assessments~\cite{reich2019collaborative}.

Therefore, a more controlled setting for assessment of how models would have performed in real time can be achieved via the \emph{simulated real-time forecasting} setting which is popularly utilized. This means using the version of the data available at a particular moment in time. To facilitate this, there are ready-to-use tools and repositories available such as for influenza and COVID-19~\cite{reinhart2021open,kamarthi2021back2future}. These tools enable modelers to train models with unrevised data and then evaluate them in curated data that best represents the actual burden of the disease.

\subsubsection{Evaluation of point forecasts}
Point forecasts return one single value per forecasting target, which can be aggregated in multiple ways. 
\citet{tabataba2017framework} study different point error metrics and empirically show that each of them lead to a different ranking of models. % for different epi-features -> ensemble is best forecast 
Specifically, given the prediction error $e_t$ and forecast evaluation period $\{1:T\}$, they studied mean absolute error  $\mathbf{MAE} = \frac{1}{T}\sum_{t}^{T} |e_t|$, root mean absolute error $\mathbf{RMSE} = \frac{1}{T}\sqrt{\frac{1}{T} \sum_{t}^{T} |e_t|}$, and variations like MAPE and NMSE. 
They also pointed out that the selected metrics do not make the distinction in underestimating or overestimating, which has a different impact on real decision making.
In our review of multiple methodological papers, we found that for the tasks future incidence and peak intensity, the preferred choice has been MAE and RMSE~\cite{cramer2021evaluation,adhikari2019epideep} while MAE has been preferred for event-based tasks like peak time and onset~\cite{kandula2019near}. However, official CDC evaluations~\cite{biggerstaff2016results,biggerstaff2018results,johansson2019open} usually opt for de-emphasizing point forecast evaluation in favor of probabilistic evaluation.

\subsubsection{Evaluation of probabilistic forecasts}
\label{subsubsec:eval_prob}
Probabilistic forecasts are more informative because they provide a balance between accuracy and uncertainty quantification, being the latter useful for high-stake decisions as in public health.

Recent CDC initiatives for forecasting like flu (FluSight challenge)~\cite{reich2019collaborative,biggerstaff2016results} and
Ebola~\cite{johansson2019open} adopted a modified version of the logarithmic scoring rule~\cite{gneiting2007strictly}, which is referred as \emph{log score}.
The log score assesses the probability that the forecasts are assigned to the actual value of the target on a logarithmic scale. Let $\mathbf{p}$ be the set of probabilities for a given forecast and $p_i$ is the probability assigned to the ground truth target value $i$. Then, the log score is as follows,
\begin{equation*}
     \text{Log score}(\mathbf{p},i) = \ln (pi).
\end{equation*}
For event-based tasks, the labels to predict are discrete so the calculation of log score is straightforward, but, for real-valued tasks, we have to first discretize the possible values of the target (a predefined range) in bins of equal width $B$. 
In FluSight challenges, $B$ has been set to 0.1~\cite{predict_cdc} and in other publications, $B$ has been set to 0.5~\cite{reich2019collaborative}; however, it is unclear what are the effects of the selection of $B$. 
In addition, there has been discussion on whether or not to use multi-bin or single-bin evaluation \cite{bracher2019multibin,reich2019reply} and the community has adopted single-bin.
For boundary cases, the log score will take 0 when all the probabilities are put in the right bin, and its lower bound is set to -10.
Thanks to the bounds of the log score, we can derive a more interpretable measure named \emph{forecast skill}~\cite{biggerstaff2018results}, which is defined as the exponentiated mean of log scores for a model. Therefore, a log score of 0 will lead to a forecast skill of 1, and a low log score like -10 will lead to a forecast skill close to 0.

The metrics above work well for traditional forecasting targets that are represented as percentages, where the domain of values rarely go beyond a predefined range. However, in the COVID-19 pandemic, we did not have any percentage as a forecast target. Instead, we had number of cases, deaths, and hospitalizations, which are targets where obtaining a meaningful binning is a non-trivial task.
Due to this reason, the CDC and \hub moved their evaluation to work with \emph{interval score}~\cite{gneiting2007strictly}.
Let $F$ be a given forecasting distribution, $y$ be the ground truth target value, and the lower and upper end of the central $1-\alpha$ prediction interval of $F$ is $l$ and $u$, then the interval score is
\begin{equation*}
\small
\text{IS}_{\alpha}(F,y) = (u - l) + \frac{2}{\alpha} (l - y)  \mathbbm{1}(y < l) + \frac{2}{\alpha} (y - u) \mathbbm{1}(y > u),
\end{equation*}
where $\mathbbm{1}(\cdot)$ is the indicator function.
\citet{bracher2021evaluating} built on top of this metric to incorporate multiple central prediction intervals. Specifically, for $\alpha=0.02, 0.05, 0.1, 0.2, \ldots, 0.9$ (11 intervals in total, $K=11$), we have the weighted interval score as
\begin{equation*}
\small
\text{WIS}_{\alpha_{\{ 0:K\}}}(F,y) = \frac{1}{K+1/2} \times |y - m| + \sum_{k+1}^{K} \{ w_{k} \times \text{IS}_{\alpha_k}(F,y) \} ,
\end{equation*} 
where $m$ is the predictive median.
\citet{cramer2021evaluation}, including co-authors from CDC, evaluated models in the \hub using WIS as the main evaluation metric. However, they found that not all forecasters provided predictions for the same set of locations and periods. To resolve this, they defined $\theta_{m}$ to be a score of a model $m$ which is based on pairwise comparisons of the relative WIS of model $m$ and all other models in the hub. Given a pair of models $m$ and $m'$, they first calculate $\theta_{m m'}$ to then be aggregated via geometric mean to get $\theta_{m}$ as follows:
\begin{equation*}
\small
\theta_{m m'} =  \frac{\text{mean WIS of } m}{\text{mean WIS of }m'}; \quad
\theta_{m} =  \big( \prod_{M'=1}^{M} \theta_{m m'} \big)^{1/M}.
\end{equation*}

Another popular evaluation metric used to evaluate the probabilistic prediction is the \textit{coverage score}. This score simply measures the fraction of times a the range determined by a confidence interval covers the ground truth value. The usual confidence intervals that are used are $90\%$ and $50\%$. Since confidence intervals determine the uncertainty and variation of future forecasts, \textit{coverage score} is a good measure of calibration of model predictions.

\section{Modeling Paradigms--An Overview}
\label{sec:modeling_paradigms}
As mentioned before, in this survey we broadly classify methods into three categories: mechanistic, statistical, and hybrid. Here we briefly introduce them to later develop each of them in detail.

\paragraph{Modeling paradigm 1: Mechanistic Models} We briefly visit the past literature of traditional mechanistic models that encode mechanisms of epidemic spread. We describe mass-action compartmental models that describe transmission of disease at large-scale population level and move towards more fine-grains metapopulation models that assume
of heterogeneity in population and incorporate information such
as human behavior and mobility. We finally look at agent-based
models that simulate individuals’ behaviors and their relations via
contact networks.

\paragraph{Modeling paradigm 2: Statistical, Machine Learning, and AI Methods}
This section comprises more recent state-of-art methods of data-driven statistical and deep learning models that have risen to be increasingly influential. Their influence has been strengthened due to their effectiveness and ability to leverage a wide variety of data including online search queries, text from social media as well as satellite images while requiring little assumption on transmission dynamics. We go over various regression approaches starting with linear regression models and
discuss complex and hierarchical regression methods. We then
focus on leveraging recent advances in deep learning to learn from
large datasets with high dimensional feature space and learn rich
feature representation and incorporate multimodal data and spatio-temporal structures. We finally discuss traditional and neural-based
density estimation models that learn distribution over forecast
targets and help deal with quantifying uncertainty. 

\paragraph{Modeling paradigm 3: Hybrid Models}
We finally discuss the novel class of models that incorporate expert priors of mechanistic models
with the flexible learning power of statistical and deep learning approaches at various stages of the modelling pipeline. 
We look at techniques such as data assimilation, learning
parameters of mechanistic models using statistical methods and regularizing statistical learning algorithms with mechanistic priors.
We also touch upon models based on wisdom of crowds that directly leverage expert knowledge and use ensembles of models to provide robust results in practice.

\section{Mechanistic Models}
\label{sec:mechanistic}

Mechanistic models mathematically describe the transmission mechanisms of infectious diseases in communities. 
These models have a long history dating back since the late 18th century with Daniel Bernoulli for smallpox.
Therefore, we circumscribe our work to provide a brief overview of them and highlight some of the recent data-driven advances; we refer the reader to other excellent surveys~\cite{marathe2013computational,hethcote2000mathematics,dimitrov2010mathematical} for more in-depth discussion on these models.
Following \cite{adiga2020mathematical}, these models can be broadly classified as follows: mass action compartmental models, structured metapopulation models, and agent-based models.

\subsection{Mass-Action Compartmental Models}
\label{subsec:mass_action}
\begin{figure}[h]
    % \hspace{-0.5cm}
    \centering
    \small
    \begin{tabular}{ccc}
    \includegraphics[width=0.25\linewidth]{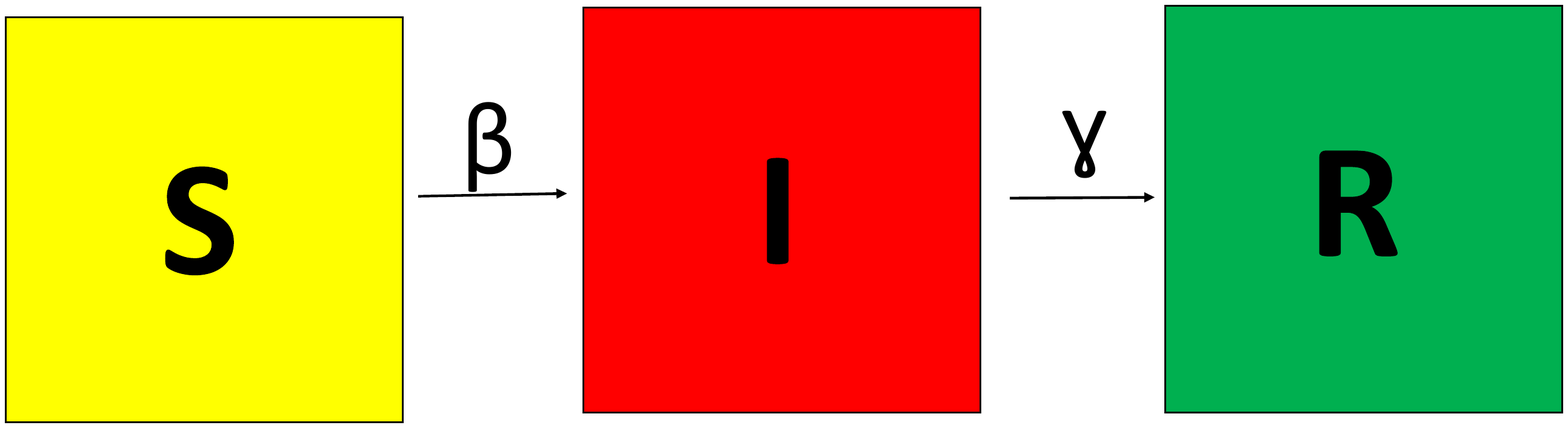} &
    \includegraphics[width=0.28\linewidth]{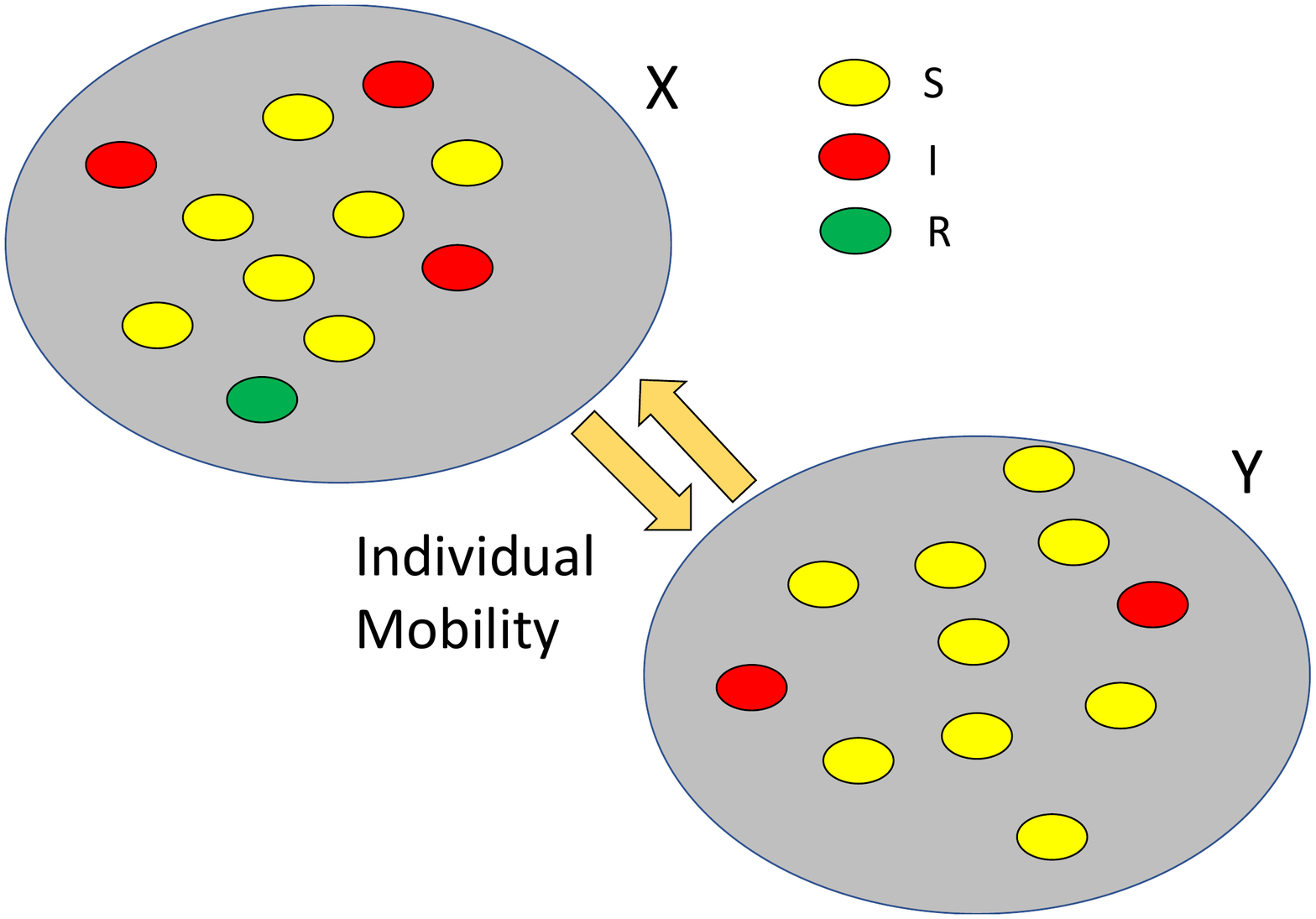} &
    \includegraphics[width=0.28\linewidth]{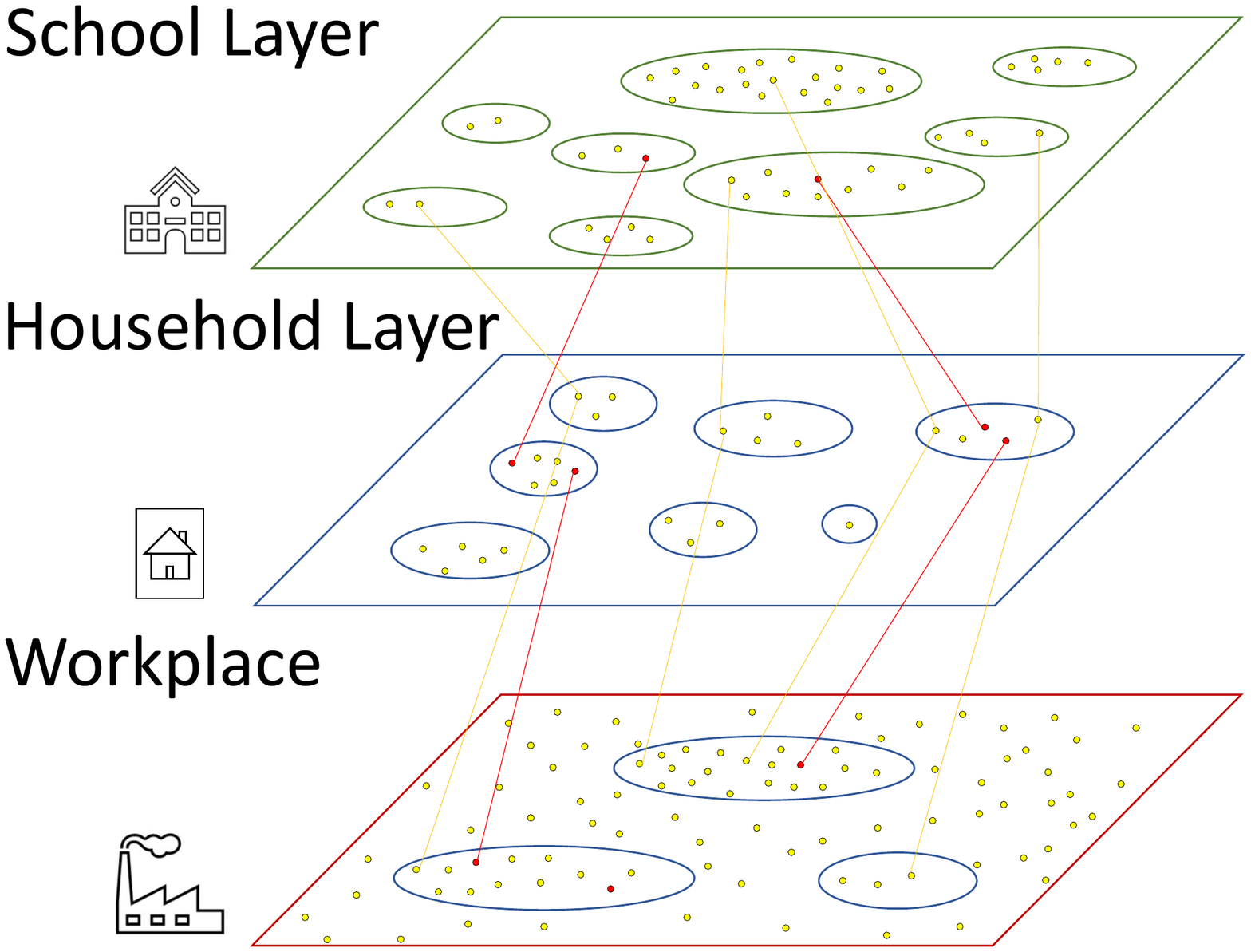}\\
    (a) Compartmental  & (b) Metapopulation & (c) Agent-based \\
  \end{tabular}
    % \vspace{-4pt}
    \caption{\textbf{Types of mechanistic models.} (a) Mass-action compartmental models like the SIR model assume each person is in contact with everyone else (perfect mixing assumption).
    $\beta$ and $\gamma$ are the parameters of the ODE equations that govern movement from one compartment to the other.  
    (b) Metapopulation models incorporate heterogeneity in the population by modeling different sub-populations separately with, and connect these models via mobility.
    (c) Agent-based network models have multiple contact networks in which people (nodes) may be involved. The spread of the disease is therefore across multiple layers. Adapted from~\cite{aleta2020modelling}.
    }
    \label{fig:compartmental}
    % \vspace{-0.05in}
\end{figure}

These models are governed by ordinary differential equations (ODEs) that describe disease transmission via the movement of individuals from one compartment to the other 
and were initially inspired by a mass-action model~\cite{marathe2013computational}.
This is when the concept of an epidemic threshold was introduced (basic reproduction number $R_0$) and multiple models were proposed taking in consideration the characteristics of a particular disease, including measles, malaria, and smallpox~\cite{hethcote2000mathematics}. 
In this type of models, transmission is considered homogeneous and population %within a compartment 
is assumed to be perfectly mixed (each person is in contact with everyone else).
These models are usually named with acronyms, often based on how individuals flow between compartments. 
For instance, in the classic SIR (susceptible, infected, removed/recovered) model, depicted in Figure~\ref{fig:compartmental}a, susceptible individuals become infected at some rate and eventually they are removed from the infected compartment through death or recovery. 
Given these compartments and associated parameters, broadly simulation optimization and Bayesian calibration have been used for calibration of the parameters~\cite{venkatramanan2018using}. 
However, a common issue in the calibration of these models is sensitivity of the parameter space to outcomes---a small error in estimating the parameters may lead to very different predictions~\cite{edeling2020model,hazelbag2020calibration}.

Recent work in forecasting using these models includes SIRS models for influenza 
that considers viral evolution across seasons and within the same season~\cite{du2017evolution}.  
For Ebola, \citet{gaffey2018application} used the EbolaResponse model, an SEIR-like model that accounts for different ways of transmission.
For the COVID-19 pandemic, \citet{wu2020nowcasting} proposed one of the first models (an SEIR model) to forecast the domestic and international spread of COVID-19 in the first weeks of the community spread at Wuhan, China.
COVID-19 CovidSim~\cite{ferguson2020impact} was an influential model that supported initial decision making in the UK. This work was followed up with criticism due to their calculation of uncertainty and the lack of study of the model's parameter sensitivity~\cite{edeling2020model}. 
The IHME model~\cite{ihme2020modeling} was CovidSim's counterpart for the US, also criticised for their uncertainty quantification~\cite{holmdahl_wrong_2020}.
\citet{morozova2021one} explored using mobile device geolocation data~\cite{crawford2021impact} as input for a parametric function that estimates the transmission rate of a SEIR-type model, which was used to forecast for the spread of COVID-19 in Connecticut, USA.
In the \hub, there have been a few good performing mass-action models~\cite{cramer2021evaluation}. 
MechBayes~\cite{gibson2020real} proposed a SEIRD model embedded into a fully Bayesian framework with informative priors to the epidemiological parameters. 
ParamSearch~\cite{gu2020covid} proposed an end-to-end differentiable SEIR model.
While these models have been useful in several applications,
the assumption that the population is uniform and homogeneously mixing may be unrealistically assumed in other cases 
as there are interactions that depend on many factors including age groups and social-economic reasons, which motivates the types of models described below~\cite{marathe2013computational}.

\subsection{Structured Metapopulation Models}
These models build on top of the mass-action compartmental models and incorporate heterogeneity in the population, but assume homogeneity at ‘right’ granularity (see Figure \ref{fig:compartmental}b. 
\citet{mossong2008social} is one of the first studies in demonstrating the benefits of modeling different sub-populations (e.g., low-granularity geographical regions and age groups, sex) in a large-scale experiment.
Spatial metapopulation models, which pay special attention to spatial transmission like commuter connections among sub-populations, have received some recent attention.
Global epidemic and mobility (GLEaM)~\cite{balcan2009multiscale} is a spatial metapopulation model based on multiscale mobility networks. Specifically, using Voronoi tassels around transportation hubs, they identify 3000+ sub-populations around airports in 200+ countries. 
Forecasts of this method for COVID-19~\cite{chinazzi2020effect} were submitted to the \hub.
Similarly, \citet{pei2018forecasting} proposed to use available commuter data to estimate human mobility, which is incorporated into a spatial metapopulation model.
This work showed improvements in using influenza metapopulation forecasts with respect to a model for each location that assumes isolation.
\citet{venkatramanan2021forecasting} 
leveraged anonymized mobility maps aggregated from millions of smartphones and show how their model can perform as good as when using commuter surveys, which are sparse and more expensive.

Another line of work focuses on modeling heterogeneity of interactions. 
\citet{srivastava2020learning} proposed a heterogeneous infection rate model which incorporates inter-regional mobility. They linearize this model for efficient training. Their model learns the parameters dynamically and automatically over time, which enables some epidemiological insights with respect to the regional response to the epidemic. They found the ensemble of their models is the best performing.
\citet{gopalakrishnan2020globally} proposed to use local mechanistic models at county granularity to obtain more accurate forecasts at the state level.
\citet{geng2021kernel} proposed to connect the spatial spread of COVID-19 with dynamic multifractal scaling. 
They looked at data at the county level over time,
revealing a dynamic multifractal scaling in the spatial correlation.
They introduce a kernel-modulated SIR model mapped onto a multifractal population framework, which allows them to exploit spatio-temporal patterns.
AutoODE~\cite{wang2021bridging} proposed a connection between mechanistic models and automatic differentiation. They introduce a spatio-temporal SuEIR model where they model the transmission among adjacent states as a learnable matrix (and solve it via low-rank matrix factorization) and use linear or piecewise linear functions for the mechanistic parameters.
Further challenges on constructing this type of models range from finding the ‘right’ granularity to modeling complex social structures~\cite{ball2015seven}.

\subsection{Agent‑Based Network Models}
\label{subsec:agent}
These models work around the idea that modeling individual behavioral adaptation and the incorporation of details in the underlying contact networks can improve our understanding of how the disease spreads~\cite{marathe2013computational}. 
These models rely on building simulations where agents host and transmit the virus within their multiple contact networks (there is a within-host disease model, e.g., SIR), as depicted in Figure~\ref{fig:compartmental}c. A discussion on the challenges on building networked models can be found in~\cite{pellis2015eight}, which include building these networks such as contact networks~\cite{eames2015six}.
\citet{dimitrov2010mathematical} surveyed some methods that work with little information but strong assumptions about the population structure. The population is assumed to resemble a random graph with a specific degree distribution, which allows them to leverage techniques such as bond percolation to model the heterogeneity of disease transmission.
Other types of methods attempt to recreate the population with great detail and realistic urban settings~\cite{eubank2004modelling}, also known as twin cities. 
These detailed simulations are computationally expensive, and there have been attempts to reduce this burden with distributed systems, e.g., EpiFast~\cite{bisset2009epifast}, which can simulate a large social contact network in minutes.
Using a simulation model like EpiFast for forecasting is a non-trivial process because the choice of the parameters is important. 
\citet{nsoesie2013simulation} presented an ML-based method to decide whether some of the predefined parameters can be used in forecasting, or if new parameters should be proposed with a combination of simulation and optimization.
\citet{tabataba2017epidemic} performed data assimilation-based forecasting for EpiFast. They used particle filters leveraging beam search \cite{mousavi2012enhanced} for more comprehensive exploration of state space.
FRED~\cite{grefenstette2013fred} is an open-source software for detailed modeling of epidemics at state and county level in the US, and a participating model in the FluSight forecasting challenge.
\citet{venkatramanan2018using} introduced an Ebola model that can incorporate specific characteristics of the disease transmission like safe funerals, although it does not model it explicitly. This work also pinpoints the challenges in calibrating these models (e.g., fixing some parameters with expertise) and the difficulties in qualitative/quantitative comparisons between agent-based methodologies.

\section{Statistical, Machine Learning, and AI Methods}
\label{sec:stat}
Statistical models (also known as phenomenological models) learn patterns from a wide variety of input signals that are assumed to influence epidemic patterns. Unlike mechanistic models, they typically require little modeling constraints on the models and provide a flexible learning approach to find the best set of parameters that can model patterns from given data.
This has enabled wide-spread use of these models on a wide range of large datasets including social media, satellite images, etc. with great effectiveness.
We also focus on recent advances in deep learning models that efficiently learn to encode useful information from large high-dimensional multi-modal datasets incorporating spatio-temporal structures.

\subsection{Regression Models}
\label{sec:regression}
Regression models learn a function $y_t = f(x_t)$ where $y_t$ is the target to predict, i.e, the future forecast, and $x_t$ is a list of signals available on week $t$. Different classes of well-known regression models have been used with a variety of signals as we describe below.

\subsubsection{Sparse linear models} 
Linear models assume that the target at week $t+k$, $y_{t+k}$ is a linear combination of fixed set of multiple indicators available at current week $t$. If the vector of indicator signals are $\mathbf{x}_t$, the models are of form $\hat{y}_{t+k} = \beta_0 + \beta^T \mathbf{x}_{t}$ and the parameters $\{\beta_0, \beta\}$ are determined by minimizing a loss functions such as the squared error
$\mathcal{L} = \left( \beta_0 + \beta^T \mathbf{x}_{t} - y_{t+k}\right)^2$. 
 In order to prevent over-fitting since the problem is sometimes under-determined,  lasso \cite{tibshirani2011regression} or ridge \cite{hoerl1970ridge} regularization is used to get \textit{sparse} parameters.

The body of work in this class of models varies based on the different sets of indicators considered and the data-preprocessing done on the indicators.
For example, \citet{polgreen2008using} and \citet{ginsberg2009detecting} were two of the earliest to use volumes of search queries as features for forecasting influenza infections using a linear model. \citet{polgreen2008using} used a carefully curated list of keywords search queries from the Yahoo search engine related to influenza. \citet{ginsberg2009detecting} introduced Google Flu Trends (GFT) in 2009, which used a proprietary set of search keywords to predict the flu. They searched over 50 million keywords to select top queries that correlated well with ILI across different regions and then built a model using the top 45 queries. 
GFT was accurate during the 2009 H1N1 outbreak, but later it became inaccurate during specific time periods and was eventually discontinued. One of the main issues was the drift in the importance of keywords possibly due to change in searching pattern as well as not considering seasonal flu patterns~\cite{olson2013reassessing,helft2008google,yang2015accurate}.
Even after such issues, this was an influential work that spanned multiple follow-up work. 

\citet{liu2020machine} used counts of media reports on COVID-19, Baidu search activity along with official incidence counts of the past 3 weeks to fit a linear model to predict future COVID-19 incidence. Due to the large number of features, they used a lasso regularizer. Similarly, \citet{mciver2014wikipedia} and \citet{santillana2014using} used a lasso regularized model with features from Wikipedia search queries and clinician's database (named UpToDate), respectively.

\subsubsection{Auto-regressive models}

Auto-regressive linear models that are classical time-series forecasting models have been widely applied to epidemic forecasting. 
\citet{soebiyanto2010modeling} use several autoregressive integrated moving average (ARIMA) and variations such as SARIMA, ARIMAX and SARIMAX; the latter two incorporate exogenous climatic signals such as rainfall and relative humidity to forecast up to 4 weeks ahead.
\citet{ray2018prediction} leveraged a standard seasonal autoregressive integrated moving average (SARIMA) model. In the FluSight challenge, they tested this model without and with seasonal differencing, being the latter the best performing among these two \cite{reich2019collaborative}. 
\citet{broniatowski2015using} used previous weeks' Google Flu Trends (GFT) and tweet volume data collected from HealthTweets as exogenous features and designed an ARIMAX model for forecasting. \citet{kandula2019near} used both GFT data and designed an ARIMA based model. They tested with and without seasonal and trend decomposition and saw a negligible difference in performance.

ARGO (Autoregression with GOogle search data) \cite{yang2015accurate} and its derivatives are a set of extensible auto-regressive models that rely on google search queries.  
Motivated from the weaknesses of the GFT system, which was susceptible to changes in search behavior~\cite{olson2013reassessing,yang2015accurate}, they instead use volumes of search query terms from Google Correlate/Trends website as features, along with previous weeks' predictions
which incorporate seasonality.
\citet{santillana2015combining} combine ILI estimates collected from multiple traditional surveillance and digital sources 
to generate forecasts. They employ a stacked linear model to select the best linear combination of multiple weak predictor models.

\subsubsection{Traditional ML models}

Next, we discuss more complex regression models which have non-linear dependencies between signals and output features.
Matrix factorization models for regression have also been explored by \citet{chakraborty2014forecasting}. They use nearest-neighbor based regression, and matrix factorization using neighbor embedding to forecasting ILI counts. 
\citet{wang2015dynamic} replaced the linear model of ARIMA with a generalized linear model with a link function corresponding to Poisson distribution for capturing the case incidence. 
To constrain the total number of parameters, they assume that models for similar time-periods and similar seasons of different years share similar structure. Therefore, they also reduce the L2 distance between parameters of nearby weeks and weeks from the same season as a form of regularization.

\subsubsection{Hierarchical models} 
\label{sec:hierarchical}
Most models studied so far are independently trained to forecast for a specific region of fixed spatial granularity (such as county-, state-, regional-, or national-level). Next, we explore the models that combine predictions from individual models for each county or state and refine these predictions using information from all regions together in a hierarchical manner.
ARGO2 and ARGOX ~\cite{ning2019accurate, yang2021use} extend ARGO to predict regional wILI through a 2-layer hierarchical model (see Figure~\ref{fig:stat1}a) where the lower level is a LASSO-regularized linear model that uses ILI and GFT data to predict region-level incidence. The second layer model uses the predictions from the lower level as well as correlations of changes in search trends across pairs of regions to incidence at higher granularity as parameters to model a multivariate Gaussian random variable of future incidence for all states to capture cross-state spatio-temporal interactions. 

\citet{zou2018multi} proposed a multi-task learning approach of jointly optimizing model parameters from multiple regions. They propose using two models for multi-task learning: elastic nets (linear regression with lasso and ridge regularization) and Gaussian Processes (GP). For Elastic nets weight and bias parameters are simultaneously optimized by a single program. For the GP-based method, they used multi-task Gaussian Process \cite{williams2007multi} where the kernel function of covariance matrix compares features from different tasks (regions in this case) thus implicitly learning cross-regional relations.
They found multi-task learning not only improves performance but also may help in scenarios where reporting may be missing temporally and geographically (sporadic reporting).
Similarly, 
\citet{matsubara2014funnel} proposed a multi-level model to capture trends over various diseases simultaneously across different locations. Their model contains learnable parameters that are disease- and location-dependent, which are optimized
using the minimum description length (MDL) principle. The model accounts for infection dynamics, periodicity, and external effects (like pandemics). They also use multi-level optimization where they first fit globally across all regions and then fine-tune region-specific parameters.

\subsection{Language and Vision Models}
Textual and image data are increasingly used as a robust source of surveillance data due to their ease of collection. In this subsection, we specifically look at methods that perform pre-processing in these data sources to extract useful signals and/or representations for forecasting.

\subsubsection{Vision models}
Not much work has been done in using computer vision algorithms to inform epidemic forecasting. 
Only some recent work started exploring the use of satellite images of locations sensitive to the effects of an outbreak like hospitals.
\citet{butler2014satellite} used satellite data from RS metrics of parking lots near hospitals to extract the number of vehicles and number of vacant spaces. 
These are used as features to determine future incidence at the national level using a lasso-regularized linear regressor. Similarly, \citet{nsoesie2020analysis} used vacancy of parking lots along with search data to model the spread of COVID-19 in China.
Using vision models to detect early outbreaks, especially in regions with a lack of other sources of traditional surveillance data is a promising research direction. However, accounting for confounding factors that cause a rise in occupancy of parking lots such as seasonal events or natural disasters is not well studied \cite{butler2014satellite}.

\subsubsection{Language based models} As described in Section \ref{sec:data}, text data from social media and search queries are useful surveillance data that provide near real-time information. Here, we specifically look at natural language processing and information retrieval techniques used to leverage useful features from complex text data.

A major challenge with social media data is the need to extract useful signals from large volumes of text and metadata of tweets. 
One way is to choose important keywords that occur in tweets or searches that are relevant to the disease of interest and derive features based on their occurrence.
\citet{lampos2010flu} and \citet{culotta2010towards} are two of the first papers working with Twitter data in tracking epidemics.
\citet{lampos2010flu} manually chose 2675 keywords related to flu and weighted each tweet that contains a keyword with the learned weight assigned to the keyword. The sum of weights for all tweets is used for ILI forecast. For this, they trained a lasso regularized linear model whose parameters are weights of keywords for each tweet.
\citet{culotta2010towards} built on top of the manual method proposed in \cite{ginsberg2009detecting} for tweets from Twitter. Specifically, they choose about 5000 candidate keywords starting from 5 manually chosen flu-related words and expanding it by looking at words most frequently occurring in tweets with hand-chosen keywords. Then, they train a logistic regression classifier for each keyword to check if a tweet with a given keyword is ILI related. The classifier filtered tweet volume for each keyword which are used as features for the final linear regression model to forecast ILI.

This was followed up by several other papers primarily proposing more sophisticated text mining models. \citet{lee2013real} analyzed the importance of different sets of keywords across different geographic areas and time periods. 
\citet{lamb2013separating} constructed different markers for each tweet based on stylography, usage of keywords (for instance whether it is used as a possessive noun), parts of speech templates that are used as features. \citet{paul2014twitter} combined these signals to train a simple logistic regression classifier on the tweet text to determine if it is related to influenza infection and combined the tweet counts with GFT features to improve on flu forecasting.
\citet{tran2019seasonal} demonstrates the benefits of incorporating a seasonal adjustment on top of using features derived from search engine logs.

An automatic unsupervised method to extract features from text is through word embeddings, a real-valued scalar mapping, based on the context in which words and phrases occur like the neighbourhood of words that occur along with given words and phrases \cite{joulin2016fasttext,mikolov2017advances}.
\citet{zou2019transfer} used search frequencies of queries related to influenza as features for a lasso and ridge regularized linear model. They chose the candidate queries based on word embedding of search queries generated by \texttt{fasttext} \cite{joulin2016fasttext} on Wikipedia corpus. They also introduced transfer learning to use the weights learned for a region in other regions that may use a different language. Specifically, they measure semantic similarity based on \texttt{fasttext} word embedding of queries trained on multi-language data. They also account for spatial similarity using Pearson correlation of query counts over multiple weeks for two different regions.
Leveraging these two similarities, they appropriately map queries from the source language to the target language and choose the best queries in the target language as features for the linear model.

\begin{figure}[h]
    \centering
    \small
    \begin{tabular}{cc}
    \includegraphics[width=0.4\linewidth]{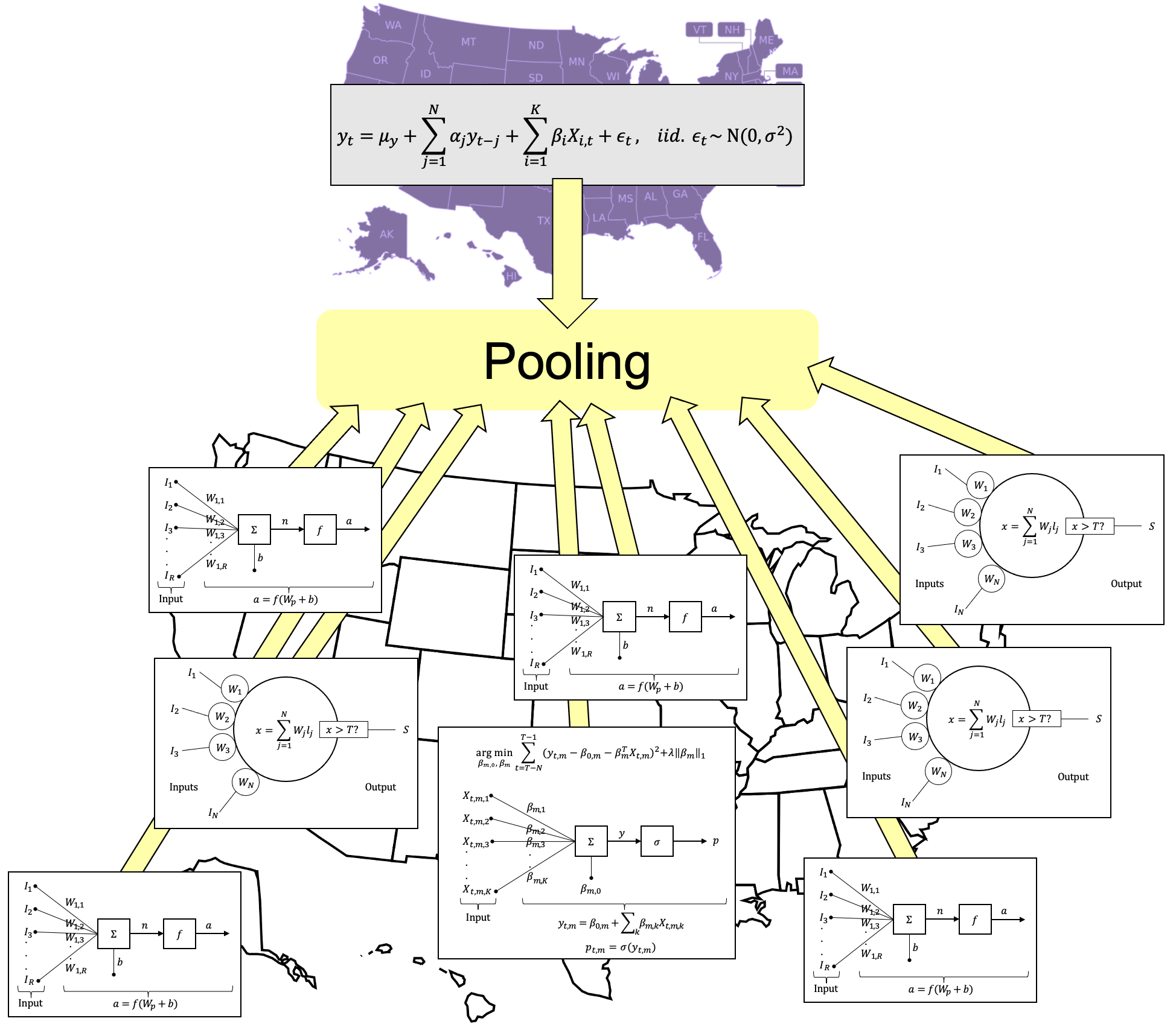} &
    \includegraphics[width=0.6\linewidth]{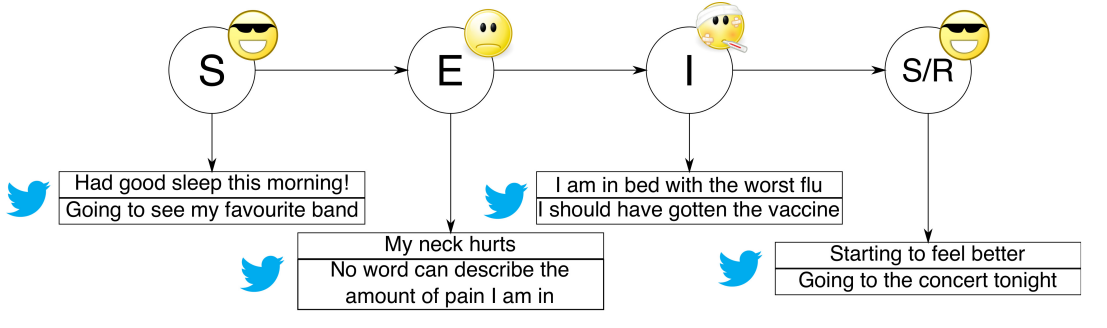} \\
    (a) Hierarchical regression  & (b) Probabilistic topic modeling \\
  \end{tabular}
    \caption{
    \textbf{Examples of regression and language models.}
    (a) ARGO2~\cite{ning2019accurate} constructs predictions at higher geographical granularities by pooling its predictions at lower granularities. 
    (b) The HFSTM model~\cite{chen2014flu} incorporates topic modeling that compartmentalizes each user to a state among SEIR compartments.}
    \label{fig:stat1}
\end{figure}

Probabilistic topic models are automated methods for extracting topics in tweets whose priors can be shaped by useful keywords.
ATAM (A Model for Ailments in Twitter)~\cite{paul2012model} is a probabilistic topic model that associates symptoms, treatments, and keywords found in Twitter with diseases (ailments) and is applied in nowcasting CDC-reported flu counts.
This method was later extended to a wide range of diseases by incorporating prior knowledge as Dirichlet priors~\cite{paul2011you}. 
HFSTM~\cite{chen2014flu} (Hidden Flu-State from Tweet Model) is a hidden Markov model that characterizes the latent state of a Twitter user into compartments (such as suspected, infected and recovered) using the sequence of tweets by a specific user (see Figure~\ref{fig:stat1}b). 
These counts are used in a linear regressor to forecast future incidence. 
Similarly, the SMS (Social Media based) model~\cite{hua2018social} used a topic model for compartmentalizing each individual as well as a network-based approach where each individual is a vertex assigned to a compartment at any given time. 
This graph is connected to all individuals with edge weights denoting the probability of infection. Then, the parameters for both components are learned simultaneously to minimize deviation in population-level numbers in each compartment. \citet{rekatsinas2015sourceseer} used a simpler topic model that uses documents from multiple locations for rare disease forecasting of Hantavirus.

\subsection{Neural Models}
\label{sec:neural}
As the scale and variety of data relevant to forecasting increases, the statistical models discussed above will fail to capture relations from high dimensional feature spaces. Deep Learning \cite{lecun2015deep} offers a potentially broader, more flexible set of models that 1) can deal with a large amount of data 2) data from different sources with different structures and peculiarities (due to the modular nature of specialized components that deal with such data) 3) richer learned representations and greater generalization power for large-scale data. 
The main ideas that have been explored in the literature can be broadly grouped as follows.

\subsubsection{Off the shelf sequence models} 
As with previously introduced models, one key aspect of modeling progression of an epidemic is its temporal evolution. Here we introduce papers whose primarily focus is on modeling temporal dynamics via off the self sequential models, like long short-term memory (LSTM) \cite{hochreiter1997long}, gated recurrent units (GRU) \cite{cho2014learning} and transformers \cite{vaswani2017attention}, with minor modifications to the neural architecture and/or training algorithm. 
\citet{volkova2017forecasting} used two LSTMs for influenza (ILI) forecasting. One LSTM ingests ILI counts and other social-media-based features (e.g., unigrams, text embeddings), whose representations are concatenated to serve as input to a fully connected layer to make predictions. They found this method outperforms other ML models such as SVM and ADABoost.
\citet{venna2018novel} proposed to leverage LSTMs in influenza forecasting in two stages: in the first stage, an LSTM makes predictions of influenza counts, and in the second stage, this prediction is improved by being multiplied by climate and spatial factors -- these factors are not part of the neural architecture and are learned separately.
\citet{wu2020deep} introduced a transformer model for influenza forecasting which outperforms both LSTM and Seq2Seq with attention models.
In addition to past influenza counts, their features include the week number and the first and second-order differences of the time series.
They also test using time-delayed embeddings
as features, but this didn't improve their performance.
During the COVID pandemic, several papers have shared success stories on using neural sequential models for forecasting. \citet{ayyoubzadeh2020predicting} used the volume of related terms obtained from Google Trends and input it into an LSTM for forecasting incidence cases early in the pandemic. They mentioned that the limited amount of training data is leading to fluctuating performance, which points to the need of models specially crafted to overcome the specific challenges of epidemic forecasting.

\subsubsection{Modeling temporal dynamics via similarity} %dynamic
To overcome data sparsity in epidemic forecasting, several papers found it useful constructing neural architectures to explicitly model similarity.
EpiDeep~\cite{adhikari2019epideep}, designed for influenza forecasting, was one of the earliest neural models leveraging \emph{dynamic similarity} between snippets of past influenza seasons and the current season snippet (see Figure~\ref{fig:neural2}a). 
To make predictions, they concatenate a similarity-based representation obtained from deep clustering with a snippet embedding resulting from an LSTM encoder.
Using deep clustering allows them to exhibit some interpretability on the predictions by communicating the most similar season to the prediction of the current week. 
For COVID-19, \citet{wang2020examining} proposed a general clustering-based deep learning method to leverage different types of similarity with the goal of augmenting training data for each region. 
They use \emph{temporal and geographical similarity} to create clusters and train on each cluster a set of stacked RNNs with attention, where each RNN is encoding the time series of one feature (e.g., cases, deaths). They defined temporal similarity as sequence similarity, and geographical similarity as geographical proximity based on state codes in the US.
ATCS~\cite{jin2021inter} proposed to leverage \emph{inter-series similarity} for forecasting multiple COVID-related indicators. They leveraged past time segments of other regions that are similar to the most recent incidence curve of the target region by using attention over these segments. To achieve this, ATCS first finds an appropriate representation of a time segment, then encodes them, and finally aggregates them via attention.

\subsubsection{Transferring knowledge representations} 
Datasets that are available for epidemic tasks and scenarios are potentially useful in others, which can be especially relevant in scenarios with scarce data. Deep learning opens a door for knowledge transfer where other statistical methods are not capable of doing so.
CALI-Net~\cite{rodriguez_steering_2021} is a framework designed for influenza forecasting when the flu counts exhibited nonseasonal behavior e.g. as they got `contaminated' with the introduction of COVID-19 in early 2020. Models trained on historical data were failing in this novel scenario, therefore, this paper proposes a heterogeneous domain transfer learning approach to adapt a historical deep flu forecasting model (trained on historical data) to the scenario where COVID and flu co-exists.
They used a multi-task spatio-temporal model to predict for all regions, which takes as input the relevant but limited COVID-related exogenous signals (that correlate better with this nonseasonal behavior) to steer the historical model. 
MPNN+TL~\cite{panagopoulos2020transfer} is a COVID-19 model that leverages Model Agnostic Meta Learning (MAML)~\cite{finn2017model} for transferring knowledge from one country where the outbreak has progressed more to another the disease spread is at early stages.
They also use mobility data for each day to build a graph between countries as nodes. These graphs are used by message passing GNNs to derive a hidden representation for each region which is then passed through an LSTM to capture temporal patterns.

\subsubsection{Leveraging heterogeneous and multimodal data}
As noted in Section~\ref{sec:data}, there are multiple sources of data and they increase in number as data collection becomes more accessible. Some of these datasets come from digital sources which are usually more sensitive to what is happening on the ground (i.e. contain complementary information about the underlying disease dynamics).
\citet{ibrahim2021variational} proposed a COVID-19 model that leverages multiple static (e.g., demographics), dynamic features (e.g., number of cases per week, indicators of government response), and spatial factors (e.g., adjacency, travel restrictions).
The static features and spatial factors are used to create a spatially weighted graph which enters to a variational graph autoencoder to derive a latent representation, and it is concatenated with an LSTM-generated representation of the previous sequence of dynamic features to forecast the future number of cases (see Figure~\ref{fig:neural2}b).
\citet{ramchandani2020deepcovidnet} included time-series signals and cross-regional mobility (dynamic) and data from regional characteristics such as demographics, vulnerability (static). This framework is largely based on DeepFM~\cite{guo2017deepfm}, which is capable of modeling interactions between multiple groups of features.

\subsubsection{Incorporating spatial structure} 
\label{sec:neurospatial}
\begin{figure}[h]
    \centering
    \small
    \begin{tabular}{ccc}
    \centering
    \includegraphics[width=0.25\linewidth]{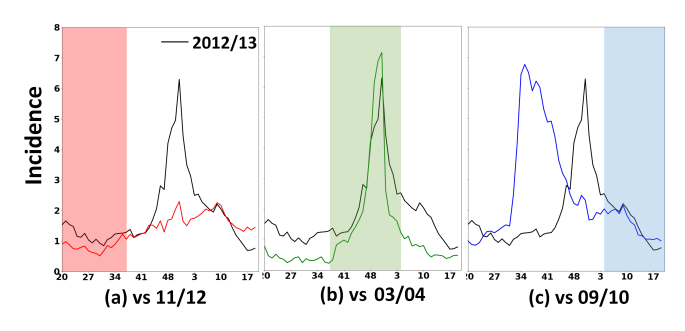} &
    \includegraphics[width=0.25\linewidth]{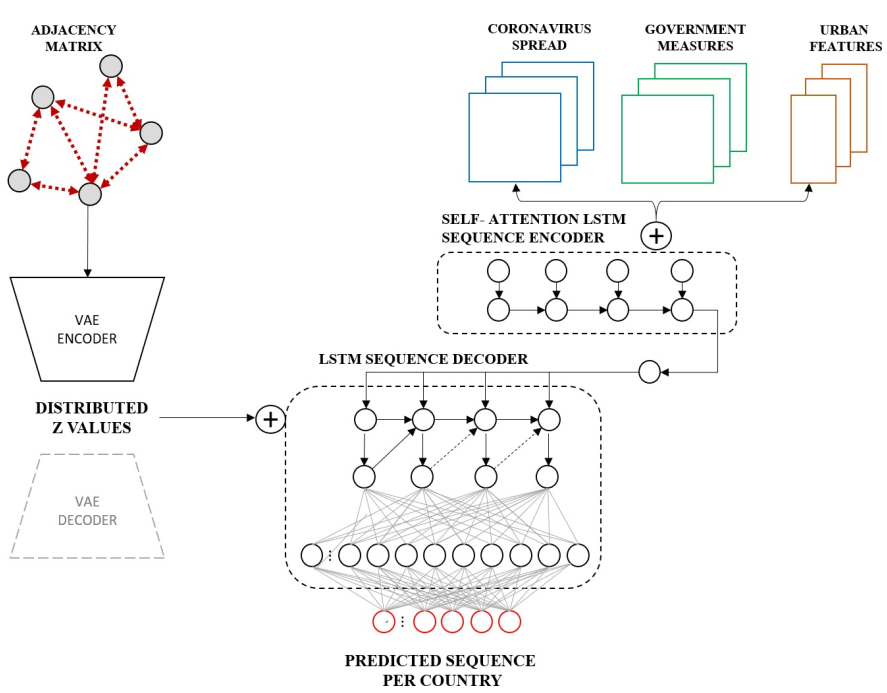} &
    \includegraphics[width=0.2\linewidth]{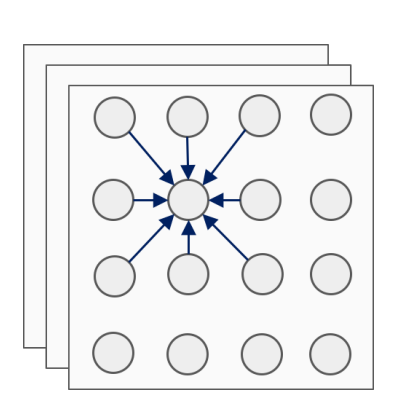} \\
    (a) Modeling dynamic & (b) Incorporating heterogeneous  & (c) Modeling spatial spread \\
     similarity  & and multimodal data & of disease via CNN \\
  \end{tabular}
    \caption{
    \textbf{Examples of neural models.}
    (a) EpiDeep~\cite{adhikari2019epideep} leverages dynamic similarity across the current and past seasons by automatically learning to infer the closest season to the current one (denoted as 2012/13). 
    (b) \citet{ibrahim2021variational} proposed a variational LSTM autoencoder model that leverage multiple heterogeneous static and dynamic signals and a spatially weighted graph. 
    (c) CNNRNN-Res~\cite{wu2018deep} incorporates the spatial propagation via convolution operations over the graph.  
    }
    \label{fig:neural2}
\end{figure}

Spatio-temporal deep learning models have the advantage of being able to extract expressive representations that capture the underlying spatial propagation of the disease.
CNNRNN-Res~\cite{wu2018deep} (Figure~\ref{fig:neural2}c) models regional proximity via a convolutional neural network (CNN) accompanied with an RNN module for the temporal dynamics. They emphasized the use of residual connections to improve generalization and hence performance. 
ColaGNN~\cite{deng2020cola} is one of the first epidemic forecasting models based on a graph neural network (GNN). This architecture uses an RNN with attention to model the co-evolving dynamics of multiple locations and their proximity, a dilated convolution layer for longer-term dependencies at different temporal granularities, and a GNN with message passing, which updates node features using propagation over the node's neighbors.
\citet{kapoor2020examining} proposed a spatio-temporal graph based on mobility and employed a spatio-temporal neural network with some skip connections for the graph convolutions to avoid dilution of the node features. 
\citet{wang2020using} proposed a graph neural network that incorporates mobility flows via recurrent message passing. \citet{roy2021deep} leveraged correlation in mobility across counties as spatial features and leveraged graph convolutional networks along with auto-regressive recurrent networks to forecast future incidence of all counties.

\subsection{Density Estimation Models}
\label{subsec:density}
Models from the above sections are largely designed to produce point estimate forecasts. As we have seen in Section~\ref{sec:eval}, probabilistic forecasts (which contain confidence bounds) are usually preferred for decision making.
Density estimation methods model the predictive distribution $p(y_{t+k}|y_{1:t})$, where $k$ is the steps ahead in prediction time. Therefore, these models can directly model uncertainty in their predictions and are comparatively more reliable on encountering anomalous data \cite{brooks2018nonmechanistic}.

\subsubsection{Parametric Bayesian inference}
For some model parameters $\theta$, these methods define the posterior $p(y_{t+k}|y_{1:t})$ 
as $ \int_{\theta} p(y_{t+k}|\theta) p(\theta|y_{1:t}) d\theta$.  The computation is based on data from previous weeks/seasons and priors for the parameters, i.e., $p(\theta|y_{1:t})\propto p(\theta) p(y_{1:t}|\theta)$, leading to a fixed functional form of the predictive distribution.
The prior $p(\theta)$ is chosen based on previous season's information or expert knowledge. The forecast distribution $p(y_{t+k}|\theta)$ is then derived via Bayesian updates.
\citet{van2014risk} introduced a semi-parametric Empirical Bayes framework to forecast dengue incidence in the 2014 World Cup game cities (in Brazil). They leveraged historical dengue data to model the new epidemic curves as transformations of the historical ones. 
\citet{brooks2015flexible} further develops this approach in influenza.
They set priors for multiple aspects of the epidemic curve such as its shape, peak height, and peak week. 
The priors are then refined based on past week's fit and the posterior (forecasts) is obtained via importance sampling.

\subsubsection{Kernel density estimation}
These non-parametric methods directly model the density of future predictions using the similarity between the characteristics of the current epidemic curve and the historical data. 
Formally, let $f(t)$ determine the distribution of a statistic of the epidemic curve at time $t$. Then, kernel density methods define the distribution as $f(t)=\frac{\sum_{t'\in N(t)}sim(t,t')f'(t')}{\sum_{t'\in N(t)}sim(t,t')}$, where $N(t)$ are neighborhood points of data point at time $t$ and $sim(t,t')$ is a function or product of kernel functions each measuring similarity of a characteristic of data point $t$ with respect to the data point $t'$. 
This similarity measure may be based, for example, on the nearest neighbors and the method of analogues~\cite{viboud2003prediction}.
\citet{ray2017infectious} introduced a set of methods based on kernel conditional density estimation (KCDE) and copulas (the latter to model dependencies among future predictions~\cite{nelsen2007introduction}). It evaluates the similarity of the wILI and GFT values of the current season with historical seasons' values using Gaussian kernels as a similarity measure to model the forecast density. 
They found that introducing a periodic kernel component (unusual in the KCDE literature) led to substantial improvements.
\citet{brooks2018nonmechanistic} proposed a delta density approach, which models relative changes across predictions of consecutive weeks. Within this approach, they present two variations: one using the Markovian assumption, and the other using an exponentially weighted sum of kernels of last week's. These models enter into an adaptively weighted ensemble for final prediction. 

\subsubsection{Other non-parametric methods}
\citet{senanayake2016predicting} used multiple kernels which captured important features across time: periodicity, long-term, short-term dependency, and space and spatio-temporal similarities across regions. They also proposed a modified Gaussian process method for capturing the time-series behavior. 
\citet{zimmer2020influenza} proposed a Gaussian process framework crafted for influenza forecasting by including the seasonal considerations of this disease such as a week of the season when the prediction is being made.  
Another very common non-parametric method of density estimation based on point-prediction models involved ensemble approaches with the combined prediction from multiple models in a probabilistic fashion. 

\subsubsection{Neural models for uncertainty quantification}
EpiFNP~\cite{kamarthi2021epifnp} and CAMul~\cite{kamarthi2021camul} leverages a class of deep generative models called Functional Neural Process \cite{louizos2019functional} to learn the forecast distribution as a probabilistic combination of functionals derived from latent embeddings of historical sources. EpiFNP also adopts the seasonal similarity idea by sampling a graph whose edge probabilities are based on the similarity between the current season and past seasons. In comparison with several methods described in this subsection (Section~\ref{subsec:density}), EpiFNP's influenza forecasts are more accurate, better calibrated, and more robust to anomalies. 
CAMul further extends the idea of leveraging similarity with training data to the case of multiple sources and modalities. It leverages the Neural Process framework to model information and uncertainty from each data source and combines their beliefs dynamically to provide accurate and calibrated forecasts.
These results show that learning probabilistic forecasts via deep generative models is a useful research direction.

\section{Hybrid Models}
\label{sec:hybrid}
Hybrid models bridge the expertise embedded in mechanistic models and priors  (Section~\ref{sec:mechanistic}) with the prediction and pattern mining power of statistical models (Section~\ref{sec:stat}). These methods can also leverage new sources of predictions coming from experts and even laypeople. 
We broadly classify them into (1) mechanistic model with statistical components, (2) mechanism informs statistical model, and (3) wisdom of crowds models.

\subsection{Mechanistic Model with Statistical Components}
\label{subsec:mec_stat}
These methods use a mechanistic model for forecasting which is aided by statistical components that cover some deficiencies like incorporating informative datasets and accounting for modeling limitations.

\subsubsection{Data assimilation} 
This is a classical technique widely used in weather modeling to incorporate informative measurements and account for their uncertainty. 
\citet{shaman2012forecasting} was one of the first to introduce the use of data assimilation in ILI forecasting by incorporating Google Flu Trends (GFT) data into their humidity-driven SIRS model. Specifically, they assimilated GFT via an ensemble adjustment Kalman filter to generate a posterior estimate of flu infection rates and found that this approach yields better predictions. 
\citet{kandula2019improved} built on top of this work and showed that their approach is portable from outpatient ILI to hospitalizations.
\citet{pei2020aggregating,yang2021development} also leverages this method to conclude that modeling individual pathogens leads to more effective influenza forecasting. This approach is based on the fact that ILI is a composite metric that contains several pathogens and viral strains.

\subsubsection{Estimate parameters of a mechanistic model from features} 
\citet{zhang2017forecasting} was one of the first papers to explore this idea. Due to the fact that the initial conditions of a mechanistic model are hard to estimate, they propose to leverage geo-localized tweets into a statistically derived formula to estimate the initial conditions. 
\citet{arik2020interpretable}~\cite{arik2021prospective} proposed using a generalized additive model to predict parameters of a modified SEIR model, which takes as input a selection of static and time-varying features.
As the SEIR model is differentiable, they train the framework in an end-to-end fashion.
To forecast, they first predict the future values of the features using a linear autoregressive model based on the most recent observations. Once they get these feature forecasts, they use them to predict the mechanistic parameters and make their forecasts. 
\citet{bhouri2020covid} proposed a similar idea using mobility and social behavior trends.
In this line of differentiable models, there are some works that proposed differentiable agent-based models~\cite{chopra2021deepabm,romero2021public} that could be potentially trained end-to-end.

In addition to forecasting, recent papers propose to leverage features in prescriptive analysis.
\citet{qian2020and} proposed a hierarchical two-layer Gaussian process (GP) (see Figure \ref{fig:hybrid}a).
The upper GP layer predicts the reproduction number $R_0$ (from which mechanistic parameters can be derived) based on country features and policy-in-place indicators, and the lower GP layer uses an SEIR model as its prior mean function to then refine it based on data. 
They perform what-if forecasting based on policy scenarios 
exploiting similarities between countries and the effect of their policies.
\citet{ghamizi2020data} incorporated statistical models to predict time-dependent reproduction numbers for a variant of the mechanistic SEIR model called SEI-HCRD. They input Google mobility data and demographics to a feed-forward neural network to output the reproduction number for the current week, which is then used in the SEI-HCRD to produce forecasts.
This work also employs genetic algorithms (GA) to choose the optimal values for current mobility features to minimize future incidence,
which is useful for planning intervention policies.

\subsubsection{Discrepancy modeling} 
These models recognize the limitations of mechanistic models in modeling the disease dynamics and address them by incorporating a statistical model that refines/corrects the mechanistic one.
Dynamic Bayesian model (DBM)~\cite{osthus2019dynamic} is a hierarchical model where a data model resolves the discrepancies with a mechanistic model. 
The data model represents discrepancies as a function of several components like trends, season-specific deviation, and state-specific deviation, modeled as stochastic processes that follow a random walk.
Dante~\cite{osthus2021multiscale} is largely based on DBM but it shares information across geographical regions, scales, and seasons. 
All the states and seasons are modeled jointly, which allows them to exploit cross-temporal and cross-regional data, and to be self-consistent at its geographical scales. 
These additions made this model outperform DBM and be the top forecasting model in the 2018/19 CDC FluSight challenge.
Inferno~\cite{osthus2021fast} is a recent Empirical Bayes analogue to Dante constructed with the goal of decreasing Dante's running time. 
This approach drops the joint modeling of all states, thus compromising self-consistency to enable parallelization.
Their results indicate that performance is slightly worse than Dante but the improvement in running time is large, from hours to minutes.

\citet{kamarthi2021back2future} proposed to refine forecasts of \emph{any} model leveraging its prediction history and temporal dynamics of data revisions.
They study the `backfill' patterns of revisions of real-time released data associated and note the negative effects of revision of features and target ground truths for forecasting and evaluation of models (either mechanistic or statistical). Their model, Back2Future, is a recurrent graph neural network based model that corrects predictions from any general forecasting models including mechanistic models to revised values (see Figure \ref{fig:hybrid}b). Back2Future showed significant improvements in top-performing \hub models, as well as the ensemble of \hub models \cite{cramer2021evaluation} and also helps with nowcasting mortality. 
DeepGLEAM~\cite{wu2021deepgleam} leverages spatio-temporal dependencies as well as prediction history of a mechanistic model to improve its  1- to 3-week ahead forecast. It
is composed of a diffusion convolutional RNN (DCRNN)~\cite{li2018diffusion} that leverages spatial structures to correct predictions from a mechanistic model.

\subsection{Priors from Mechanistic Models Inform Statistical Model}
\label{subsec:stat_mec}

\begin{figure}[h]
    \centering
    \small
    \begin{tabular}{ccc}
    \centering
    \includegraphics[width=0.3\linewidth]{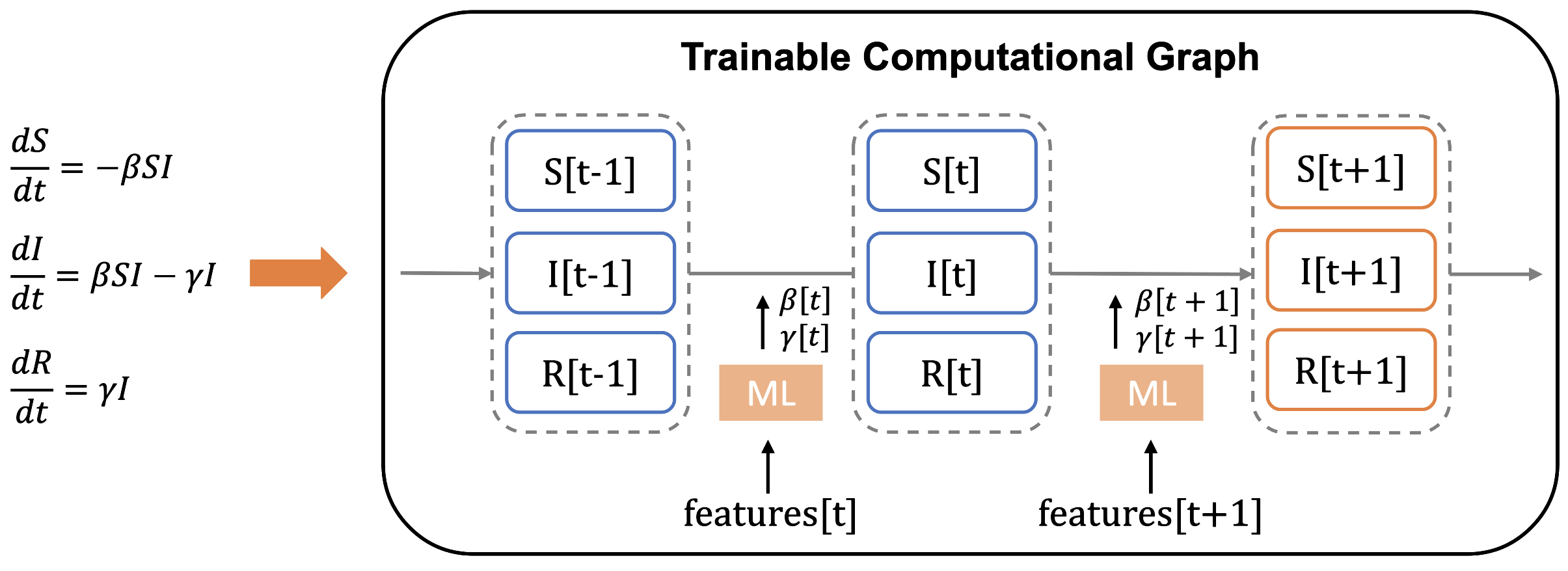} &
    \includegraphics[width=0.3\linewidth]{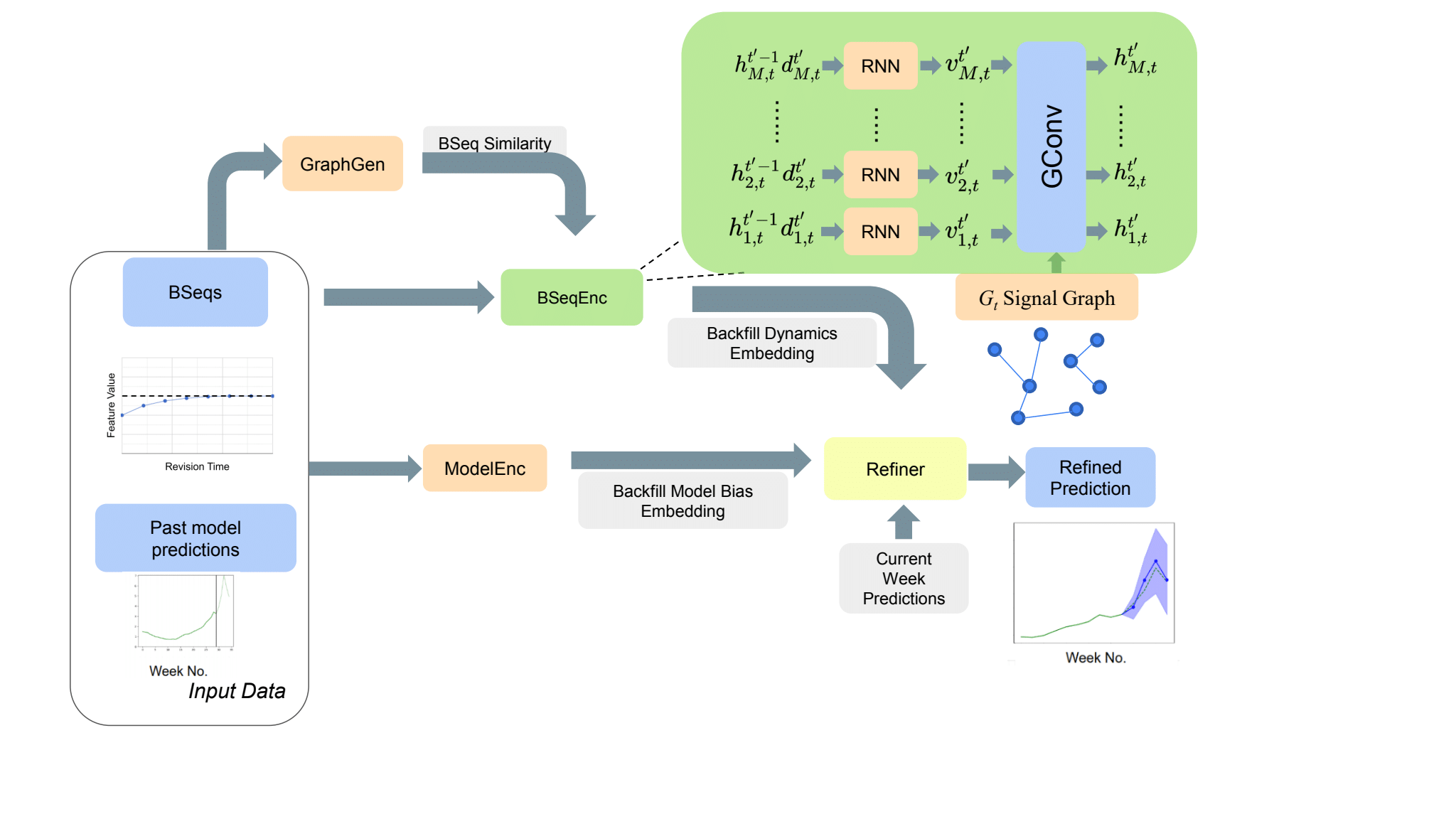} &
    \includegraphics[width=0.35\linewidth]{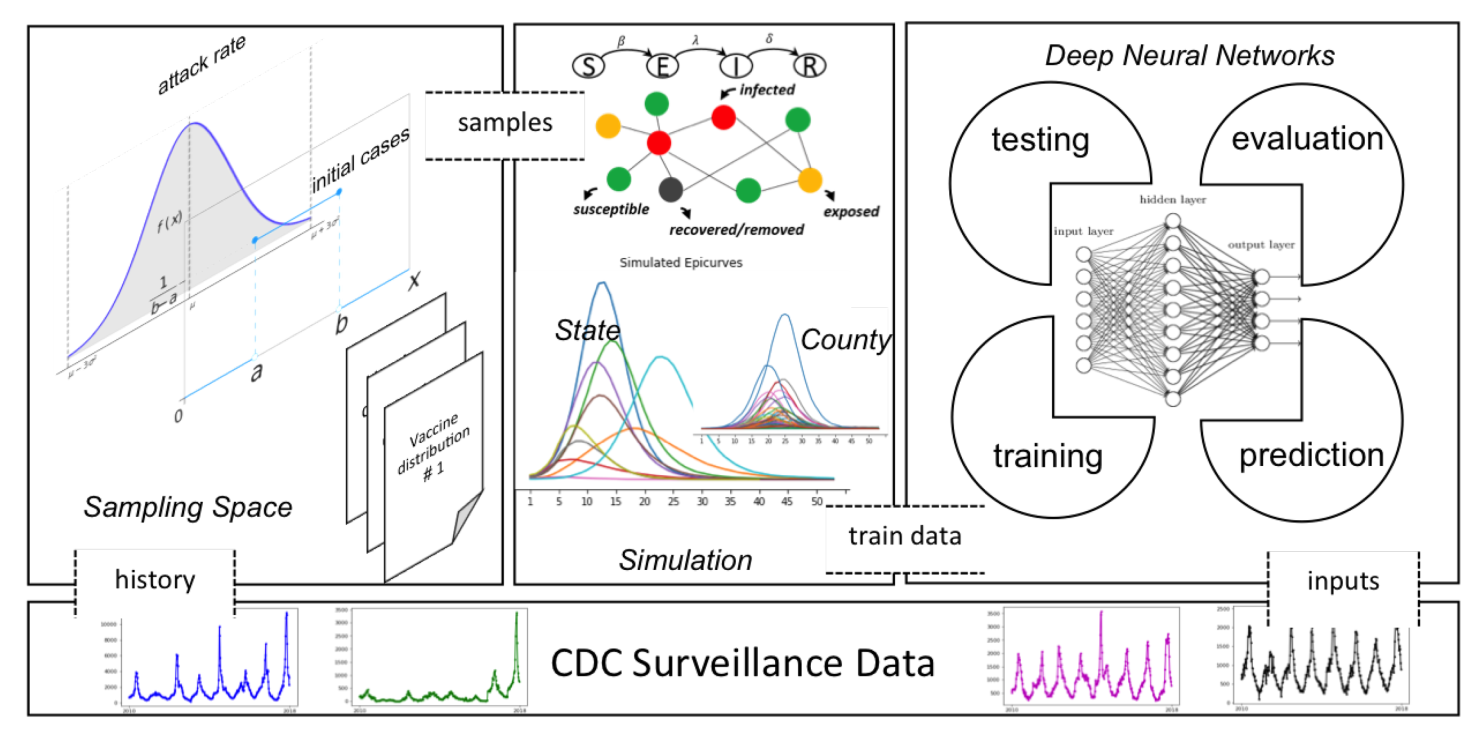} \\
    (a) Estimating param. & (b) Discrepancy modeling with & (c) Learning from mechanistic \\
    of mechanistic model & data revision dynamics & simulation data \\
  \end{tabular}
    \caption{
    \textbf{Examples of hybrid models.}
    (a) Graphical model of the compartmental Gaussian processes~\cite{qian2020and}. The upper-layer GP $f_U$ predicts $R_0$, passes to the compartmental model (red box), whose predictions sets the mean for the lower-layer GP $f_{L,i}$. 
    (b) Back2Future~\cite{kamarthi2021back2future} leverages temporal dynamics of data revisions and the history of previous predictions of a model. This approach can refine the predictions of any type model based on the revision dynamics. 
    (c) Overview of DEFSI~\cite{wang2019defsi} framework. It leverages a mechanistic simulator EpiFast to generate synthetic data to then be used as training data for a deep neural network.
    }
    \label{fig:hybrid}
    % \vspace{-0.05in}
\end{figure}

In these methods, the statistical model incorporates mechanistic priors as a source of data.
These methods directly use data from multiple simulation runs of mechanistic models or synthetic data from mechanistic models fit on real data.
DEFSI~\cite{wang2019defsi} leverages multiple runs of simulation from mechanistic models to generate high-resolution data (at finer spatial scale) because real data is sparse at this geographical granularity. 
They propose a neural architecture that learns from simulation time series data generated by a stochastic variant of the SEIR model called EpiFast (see Figure \ref{fig:hybrid}c), where their RNN captures both temporal and seasonal patterns to predict future forecasts.
TDEFSI (theory-guided DEFSI) \cite{wang2020tdefsi} extends this work 
using regularizers for encouraging hierarchically consistent predictions.
\citet{liu2020machine} proposes to use the synthetic data generated from a mechanistic model (whose parameters are fit using real data) along with complementary features (such as related to news and search engine queries) as input to a multivariate regression model. They do this motivated from the fact that there is noise and inaccuracies in real data, which a mechanistic model can address better than a statistical one.

Mechanistic priors can be also used as part of the learning frameworks of ML models. 
Recent work leverages advances in Scientific AI~\cite{karniadakis2021physics} to inform ML models of epidemic dynamics.
Epidemiologically-Informed Neural Networks (EINNs)~\cite{rodriguez2022einns} are a new class of physics-informed neural networks~\cite{raissi2019physics} that can ingest exogenous features.
Therefore, in addition to learning the latent epidemiological dynamics via interacting with a ODE-based mechanistic model, EINNs can also connect such dynamics with features to better inform forecasting.
This work aims to take advantage of the individual strengths of mechanistic models (long-term trends) and ML models (accuracy).
Other line of work leverage a recent innovations in incorporating ODEs into ML frameworks to learn unknown ODE dynamics in the latent space~\cite{rubanova2019latent,kidger2020neuralcde} and equipping ML models with known ODE equations~\cite{gaw2019integration}. The latter type of approach proposes to incorporate constraints that are imposed by mechanistic models as loss functions and regularizers~\cite{kargas2021stelar}.

\subsection{Wisdom of Crowds Models}
Wisdom of crowds (WoC) methods rely on the collective wisdom of all the previously introduced modeling approaches and/or human predictions/assumptions (either from experts or laypeople), being the latter an attempt to leverage expertise (domain heuristics which may not be included in models) and common sense.
We first start by introducing human predictions/assumptions coming from expert consensus and prediction markets, then we move to an ensemble method, capable of incorporating any kind of prediction available.

\subsubsection{Expert consensus and prediction markets}
\label{sec:expert}
Leveraging predictions and/or modeling assumptions from domain experts is useful especially when there is little data like during the early stages of a pandemic. 
Prediction markets, which are a mechanism to elicit true belief of a wide range of experts over time are used for a wide range of applications like election opinion polls \cite{antweiler19981997,forsythe1992anatomy}, forecasting product sales \cite{wolfers2004prediction,plott2000markets} and national security \cite{yeh2006using}. Prediction markets allow trading of event-outcomes backed securities with the final outcome causing the correct security to expire with 100\% of market share whereas the rest will go to 0\%. Thus, the participants who backed the correct outcome will get all the rewards based on the amount they bet on it. Prediction markets are thus also used for epidemic forecasting that enables aggregation of belief in outcomes from a large number of selected participants whose actions are motivated by the incentives of the market.

Some methods leverage simple aggregation of predictions from the general population or small sets of experts. \citet{farrow2017human} studied the efficacy of crowd-sourced predictions for influenza and chikungunya. They collected forecasts from the general public for influenza and domain experts for chikungunya and used the mean of the participant's predictions as the aggregate prediction. They observed that their aggregated predictions were better for short-term forecasting compared to top-performing statistical models.
\citet{recchia2021well} studied the efficacy of expert predictions during the early stages (April 2020) of COVID-19 pandemic over layperson predictions in UK by using responses from 140 experts. The experts were also asked to give 75\% and 25\% values to test for calibrated prediction and the consensus distribution was formed by assuming a Gaussian distribution aggregating all expert predictions.
Expert predictions were significantly more accurate than lay persons' but still overconfident. \citet{nadella2020forecasting} also performed a similar analysis on COVID-19 and the Ebola epidemic but found that experts massively overestimate in some regions (countries) and underestimate in others. 
Similarly, \citet{mcandrew2020expert} organized a survey to determine the severity of the COVID-19 pandemic from a small sample of experts from multiple disciplines and found that the predictions were well-calibrated but usually erred on being optimistic. They also extended this to predict efficacy, safety, timing, and delivery of COVID-19 vaccine \cite{mcandrew2021aggregating}.

Other works leveraged multiple experts' predictions via the prediction markets. \citet{polgreen2007use} was one of the earliest to explore how prediction markets can be used in epidemic forecasting. Each participant, who was a healthcare worker or expert. The ILI outcomes were divided into 5 bins each of which was a security the participants could bet on up to 8 weeks into the future. On average, they found that participants predicted correct levels of ILI activity 71\% of the cases with confidence in the correct outcomes increasing as they neared the target week.
\citet{tung2015using} performed a similar analysis for influenza outcomes in Taiwan. They reported that this method was better than using an average of past season values, a baseline typically used by Taiwan CDC, for up to 6 weeks ahead forecasts.

Another line of work focuses on the expert elicitation process for improving modeling assumptions.
\citet{Shea2020harnessing} discusses a multi-round structured decision making (SDM) model that leverages both the exchange of modeling assumptions and predictions from multiple forecasting groups while alleviating biases arising during group decision making like dominance effects (Figure \ref{fig:ensemble}a). The model is based on the Delphi method \cite{hemming2018practical}, and first elicits predictions from multiple groups independently. Then, they collate and anonymize the results of all groups, and make them public so that all the groups engage in a structured discussion to exchange ideas and generate new insights about the epidemic progression. This is followed by individual groups again working independently with updated information to provide revised predictions which are finally used.

\subsubsection{Ensemble methods}
\label{subsubsec:ensemble}
Ensembles of predictions from multiple models are used to obtain better predictive performance than its individual components with practical~\cite{opitz1999popular} and theoretical~\cite{bates1969combination} benefits. 
In fact, ensemble methods have consistently outperform most if not all individual methods in multiple CDC forecasting competitions, e.g., influenza~\cite{reich2019collaborative}, Ebola~\cite{viboud2018rapidd,johansson2019open}, and COVID-19~\cite{cramer2021evaluation}.

\begin{figure}[h]
    \centering
    \small
    \begin{tabular}{cc}
    \centering
    \includegraphics[width=0.6\linewidth]{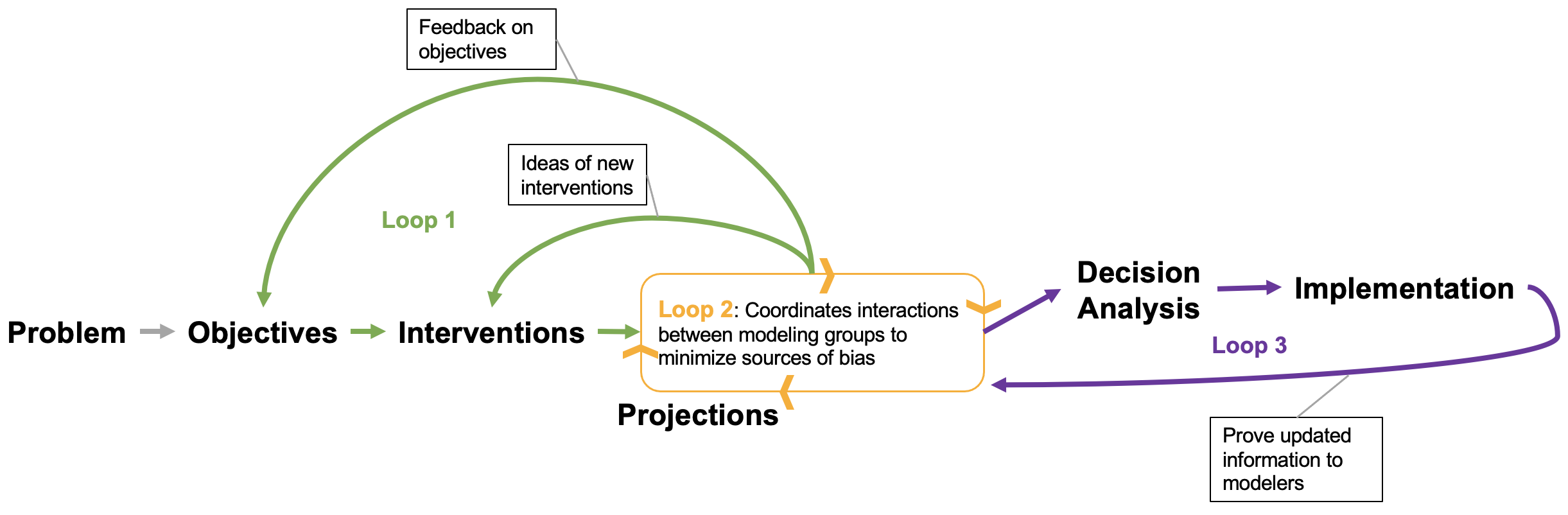} &
    \includegraphics[width=0.35\linewidth]{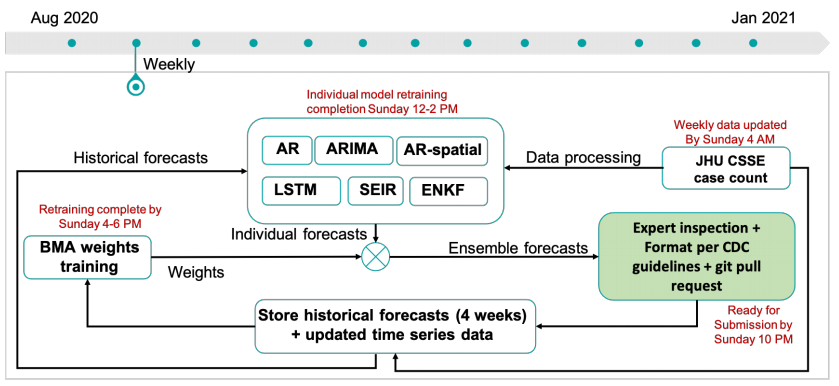} \\
    (a) Expert consensus  & (b) Ensemble pipeline \\
  \end{tabular}
    \caption{
    \textbf{Examples of wisdom of crowds models.}
    (a) Multi-round SDM model used in \cite{Shea2020harnessing} to iteratively refine multiple forecasts leveraging exchange of ideas (b) Forecasting pipeline using an ensemble of diverse models and human experts~\cite{adiga2021all}.}
    \label{fig:ensemble}
\end{figure}

\citet{yamana2017individual} demonstrated performance benefits of ensembles in influenza forecasting by testing four standard mechanistic models
with multiple filter methods (for improved calibration) and a statistical model based on ILI trajectories from prior seasons.
Their ensemble algorithm is based on Bayesian model averaging 
and takes the weighted average of individual forecast distributions for each location and prediction week.
The latter is because they found some models are more adequate at different stages of the epidemic activity. A follow-up paper~\cite{kandula2018evaluation} extends this work to include more modeling approaches.
\citet{reich2019accuracy} focused on constructing an ensemble of several influenza models participating in the CDC FluSight challenge. They use a linear combination of forecasts with weights being a function of observed features. They consider multiple weighting schemes and take into consideration factors like prediction week, a measure of model uncertainty, and ILI value at prediction time. They found their ensemble performed better (on average) than all individual component models.
More recently, \citet{adiga2021all} advocates for accounting model
complexity and diversity,
and proposes also incorporating deep learning models and including a human expert for verification of predictions (see Figure~\ref{fig:ensemble}b). 

A key limitation of the approaches above is that the models in the ensemble are required to have predictions available in all instances of historical data to learn their weights. However, in settings like the CDC FluSight challenge and the recent COVID-19 pandemic, several models join on the fly, either with each new flu season or on a weekly basis for COVID-19); thus, we may not have their complete historical predictions.
Motivated by this, \citet{mcandrew2019adaptively} presents a Bayesian framework for adaptive ensemble which can learn from the current season and more promptly update their weights on a weekly basis. 
In COVID-19, in addition to having a short amount of historical data, some models may not predict in all locations. 
\citet{ray2020ensemble} opted to use an equally-weighted average of forecasts for all models forecasting circumscribed to a specific location. This is the methodology used by the official US CDC ensemble for the COVID-19 pandemic, which leverages the probabilistic forecasts of all eligible models (i.e., submitting at least 1- to 4-week ahead) in the \hub.
\citet{kim2020covid} also worked with models in the \hub and proposed an ensemble approach using representative clustering to exploit facets where models agree. 
They map predictions (time series) to a reduced feature space to then cluster them, and later they take the median prediction which represents the ensemble prediction of the models in the cluster. This allows them to visualize (in a dashboard) a small number of representative predictions instead of the many individual predictions.

\section{Epidemic Forecasting on the Ground}
\label{sec:ontheground}

In this section we address the body of work that is focused on connecting the previously presented methodologies with practice, such as forecasting initiatives, real world experiences, and decision-making.

\subsection{Collaborative Initiatives and Experiences in Real-time Forecasting}
\label{subsec:realtime}

\subsubsection{Collaborative initiatives}
\label{subsec:initiatives}
In the introduction (Section~\ref{sec:intro}), we briefly discussed the role of forecasting initiatives (like the ones hosted by CDC) in energizing research in epidemic forecasting, fostering knowledge exchange among researchers,
and translating results into public health tools. 

The CDC FluSight forecasting challenge requests participants to submit weekly probabilistic forecasts for seasonal influenza at the US national and regional levels. Participants submit forecasts for the following targets: next 4-week future incidences, peak intensity, onset week, and peak week (note that task definitions were introduced in Section~\ref{sec:tasks}). 
Open Source Indicators (OSI)~\cite{iarpa_osi} forecasting tournament organized by U.S. IARPA was a similar challenge for multiple events including for disease forecasting (flu and rare diseases) focusing on Latin America~\cite{chakraborty2018know}.
More recently, the CDC and partners hosted a forecasting initiative \hub~\cite{reich2020reichlab}, which requested teams to provide probabilistic forecasts for one to four weeks-ahead future incidence predictions.
Given the impact of the COVID-19 pandemic, this challenge received attention from 49 teams across the US and internationally~\cite{cramer2021evaluation}, and their predictions were publicized in multiple portals, including the official CDC website~\cite{covid_forecasting_cdc}.
Among the important insights from these initiatives is that no modeling approach by any single team was effective in all instances~\cite{biggerstaff2016results,reich2019collaborative,cramer2021evaluation,mcgowan2019collaborative}. Specifically for COVID-19, the top five teams used both mechanistic and statistical approaches (including one based on deep learning)~\cite{cramer2021evaluation}, which is an indication that diversity in modeling perspectives is useful, especially when these models are used in a ensemble~\cite{ray2020ensemble}, as previously noted in Section~\ref{subsubsec:ensemble}. 
A similar collaborative initiative including an ensemble has been undertaken in Europe with the European COVID-19 Forecast Hub~\cite{eu_hub}.
The Scenario Hub~\cite{borchering2021modeling} has been created for collecting forecasts for 6-month ahead targets. The Scenario Hub focuses on projections where the targets are conditional to specific scenarios usually combining multiple factors. Examples of scenarios are various levels of expected vaccination rates and the implementation of non-pharmaceutical interventions.
It is worth noting that a pattern across these collaborative initiatives is that the ensemble is, on average, the best performing model.

\begin{figure}[h]
    \centering
    \small
    \begin{tabular}{cc}
    \centering
    \includegraphics[width=0.35\linewidth]{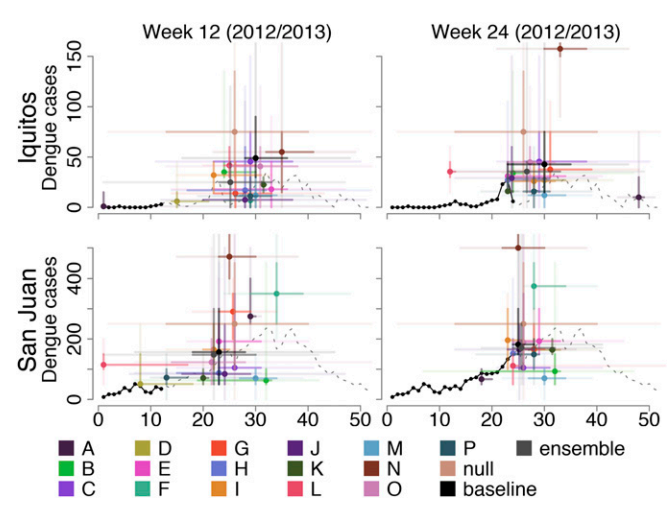} &
    \includegraphics[width=0.35\linewidth]{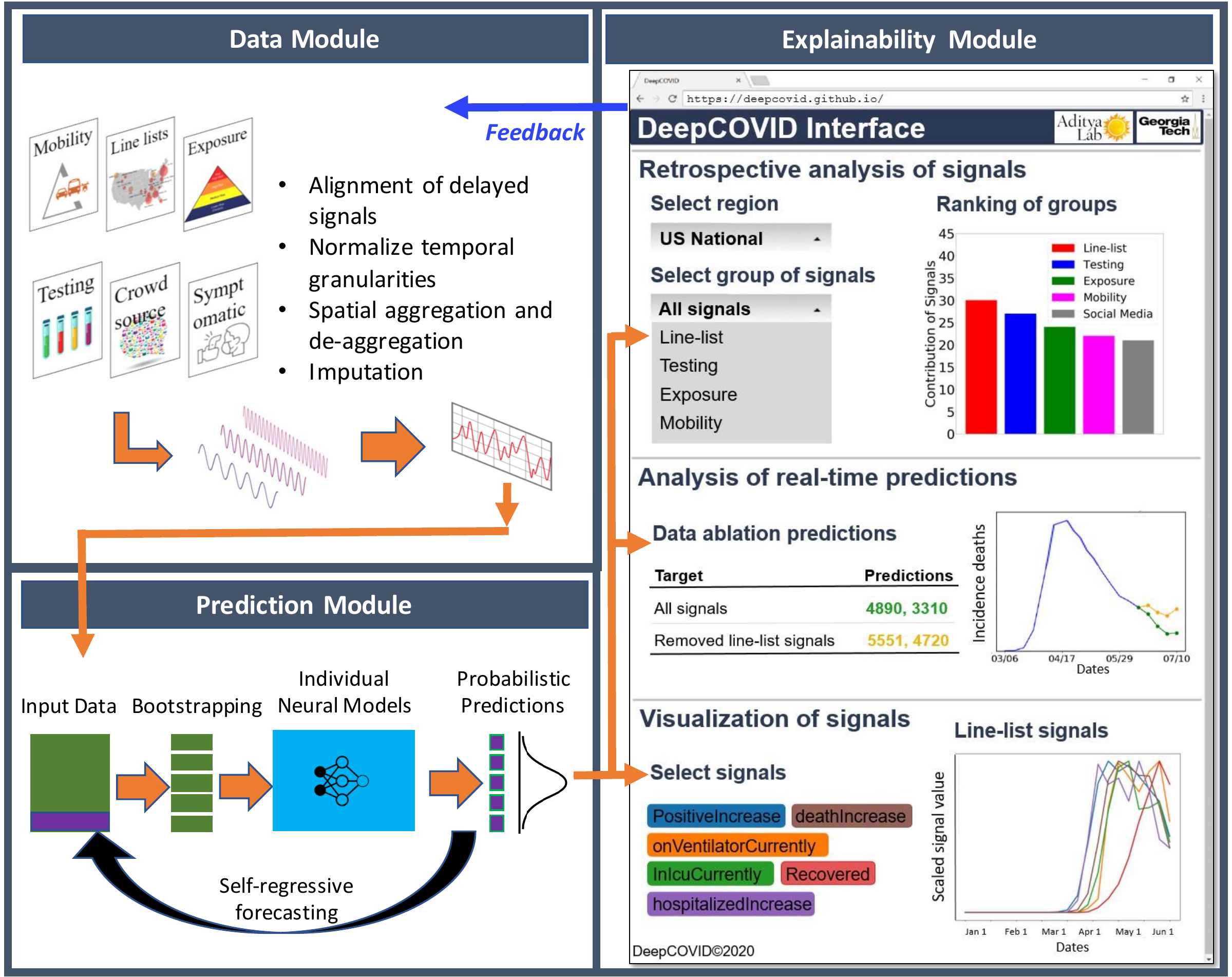} \\
    (a) Dengue collaborative initiative & (b) Real-time forecasting framework \\
  \end{tabular}
    \caption{
    \textbf{Examples of collaborative initiatives and experiences.}
    (a) Peak predictions of multiple participating models for 2012/2013 dengue season in Iquitos and San Juan~\cite{johansson2019open}. (b) Schematic of DeepCOVID~\cite{rodriguez_deepcovid_2021} operational framework for real-time COVID-19 forecasting. }
    \label{fig:real_time}
\end{figure}

Other forecasting initiatives do not request real-time forecasts but instead emulate several aspects of a real-time setting. For example, predictions were requested for  out-of-distribution samples and at different stages of the epidemic. The challenges imposed by these initiatives can provide insights for real-time deployment when needed. For instance,
the recent dengue forecasting project for San Juan, Puerto Rico and Iquitos, Peru~\cite{johansson2019open} found that forecast skill usually decreases in seasons with later and higher peaks (see Figure~\ref{fig:real_time}a).
\citet{viboud2018rapidd} presented one of the first synthetic challenges for Ebola for which they leverage a detailed agent-based simulation model to generate data and introduce the concept of `fog of war,' which is a realistic layer of noise to data. They found that their noise layer was successful as they found correlation between performance and amount of uncertainty. They also observed that model complexity was not an indicator of better performance.

There have also been other initiatives that aim to coordinate efforts to achieve common goals across multiple countries.  
For example, the Influenza Incidence
Analytics Group (IIAG) has been formed within the World Health Organization (WHO) to improve global coordination to increase the value of real-time data and analytical tools ~\cite{biggerstaff2020coordinating}. This group has identified that challenges like data quality differs across countries. Data definitions (e.g. how to define cases associated to the disease) are also inconsistent.

\subsubsection{Experiences of individual forecasters}

Here, we discuss papers that give insights about their experiences in submitting real-time forecasts dealing with issues in operationalizing different methodologies, handling changing disease patterns and data quality problems.
We further expand on open challenges of real-time forecasting and evaluation in Section \ref{sec:discussion}.

\citet{reich2019collaborative} summarize some of the challenges that individual modelers faced in the FluSight challenge for influenza forecasting. 
As we discussed briefly before, the reporting of surveillance data has delays and presents revisions which affect modeling and evaluation: large data revisions are correlated with decrease in forecasting performance. They also point out that modelers may not know how best to make use of novel data sources or may lack access to them. To facilitate this, they pinpoint the need to have data standards for collection and storage, which may increase exchange of data and expertise among forecasting groups (also related to `technical debt' we refer to in Section~\ref{sec:discussion}).
Other work from modelers noted that post-processing steps to forecasts can help to alleviate real-time challenges.
% Post-processing also seems to help in performance of real-time forecasting. 
Some refinements can be implemented to avoid being penalized harshly by the evaluation metric. For instance, for the log score metric (introduced in Section~\ref{subsubsec:eval_prob}), \citet{kandula2018evaluation} added a small probability value to forecast bins based on historical values to avoid the possibility of having zero probability in for the bin corresponding to the true value.
Additionally, there are general adjustments that can be effective for multiple modeling approaches. For example, \citet{gibson2021improving} proposes an algorithm that transforms a set of independent forecast distributions at regional level to obey the constraint that they should sum up to the national level value forecasts. Leveraging this constraint, called \textit{probabilistic coherence}, led to a 79\% increase in forecast skill. 

In COVID-19, \citet{altieri2021curating} used combined linear and exponential predictors for 2-week ahead COVID-19 forecasting, which have been displayed in an interactive visualizations publicly available along with a data repository~\cite{yu_covidseverity}.
Their experience since the initial stages of the pandemic led to a few insights.
They point to the need to uncover potential biases in the data to estimate the case fatality rate, inconsistencies across multiple geographies (e.g., how a death due to COVID-19 is attributed), and mismatch in counts from different sources.
For modeling, they use a set of linear and exponential statistical predictors whose predictions are later combined via a weighting scheme. Their cumulative and incident forecasts for cases and deaths for counties across the US were publicly accessible via an interactive dashboard.
\citet{gibson2020real} found data anomalies and reporting issues were causing unrealistic forecasts in their Bayesian mechanistic model (introduced in Section~\ref{subsec:mass_action}), ranked as top-5 model among all the models submitted to the \hub~\cite{cramer2021evaluation}. They set up a quality assurance procedure which involved looking at the recent time series of the epidemic target being forecast along with notifications of reporting issues made public by researchers from Johns Hopkins University. 
To handle these situations, they backdistributed the possible excess number of incident cases/deaths. Other data-related issues include some states not reporting data on weekends, for which they omitted those data points to avoid numerical instability or convergence failure in training their model. 

\citet{rodriguez_deepcovid_2021} describe their solutions to multiple challenges they faced deploying DeepCOVID -- one of the first deep learning based models in the \hub and   also ranked in the top-5~\cite{cramer2021evaluation}; showcasing the usefulness and challenges in operationalizing ML models for emerging epidemics (see Figure~\ref{fig:real_time}b). Their broad goals include leveraging heterogeneous data from multiple sources to provide a perspective closer to the observed data with minimal assumptions.
They describe data-related challenges in collection and preparation, like delays,  reporting interruptions and inconsistent definitions. They also identify open questions in deep learning models for pandemics, and present a framework for probabilistic forecasting with sparse and noisy data. They show their approach can obtain correlated forecasts capturing macro and micro patterns with principled uncertainty quantification.
In addition, they have a module for providing explainability to communicate to experts the important indicators that are driving their predictions.

\subsection{Bridging Forecasting with Decision Making}
\label{subsec:decision}

The primary beneficiaries of the forecasting methods discussed so far are policymakers and public health officials who aim to provide the best decision in terms of clinical, pharmaceutical and public health intervention with the aid of model predictions. \cite{lutz2019applying}.
These interventions can be strategic or tactical. 
Strategic interventions focus on large-scale decision making with the aim to minimize negative effects of an epidemic. These include vaccination decisions, lockdown and travel restrictions, economic policies, etc. In contrast, tactical interventions deal with fixed, high resolution action space to accomplish a predefined strategic goal. 
Examples of tactical intervention problems include logistical problems such as bed allocation in a hospital, ventilator allocation and vaccination distribution.
In this section, we discuss recent directions in bridging the process of forecasting with sound decision making for both kinds of interventions.

\subsubsection{Large scale simulation and scenario modeling for strategic interventions}

These methods first design a model to simulate large-scale disease dynamics to provide optimal overarching strategic decisions. 
For instance \cite{probert2018real, probert2019context} model spread of foot and mouth disease across multiple farms.

\citet{probert2018real} studies the influence of uncertainty in real-time and retrospective optimal policy planning using a simulation model. 
During the initial stages of outbreaks, they note that there is high uncertainty in predicting the future course of the outbreak. In retrospect, the optimal policy is usually dynamic, i.e., control measures change across the course of an outbreak due to changes in epidemic modeling parameters.
Thus, they  solve the optimal control problem as a sequential decision making problem using deep reinforcement learning where they model the spread of infection across multiple farms and decide to either cull or vaccinate specific farms \cite{probert2019context}. 
\citet{kwak2021deep} also used a deep reinforcement learning method leveraging demographic data and case counts to learn lockdown and travel restriction policies at the country level. 
\citet{birge2020controlling} leverages a mechanistic spatio-temporal model to simulate spread of COVID-19 infections across neighbourhoods in New York and solve an optimization problem to decide optimal subset of neighborhoods to close down for economic activity to minimize infections as well as keep the economic activity above a baseline. 
\citet{trott2021building} simulates effects of complex economic policies at state and national level such as unemployment levels, vaccination rates, cases and deaths and uses reinforcement learning to learn an optimal set of future policies that involve allocating subsidies and welfare programs.

Active management (AM) is a principled model-independent methodology of decision making in face of uncertainty that has been used for designing strategies for environmental conservation \cite{mclain1996adaptive}. 
AM involves an iterative structured approach that makes situation-dependent decisions incorporating new real-time information to resolve uncertainty in model parameters \cite{Atkins2020anticipating} (see Figure \ref{fig:decision}b).
\citet{shea2014adaptive} provided examples of AM for culling decisions based on UK foot and mouth disease outbreak data and measles vaccination campaigns. AM involves identifying an objective function, constructing alternative models, monitoring real-world outbreaks and executing intermediate decisions with a dual goal of furthering the objective and reducing parameter uncertainty in future. 
For instance, in \cite{Atkins2020anticipating} AM methodology is applied for vaccination where the initial decision to vaccinate is also influenced by resolving uncertainty about the efficacy of vaccines which is then used to make more informed decisions in future.
\citet{pei2021dynamic} model the patient pipeline during COVID-19 pandemic at a makeshift hospital and learn an optimal allocation policy for medical resources including staffing and beds.

\subsubsection{Short-term predictions for tactical decision making}
\begin{figure}[h]
    \centering
    \small
    \begin{tabular}{cc}
    \centering
    \includegraphics[width=0.4\linewidth]{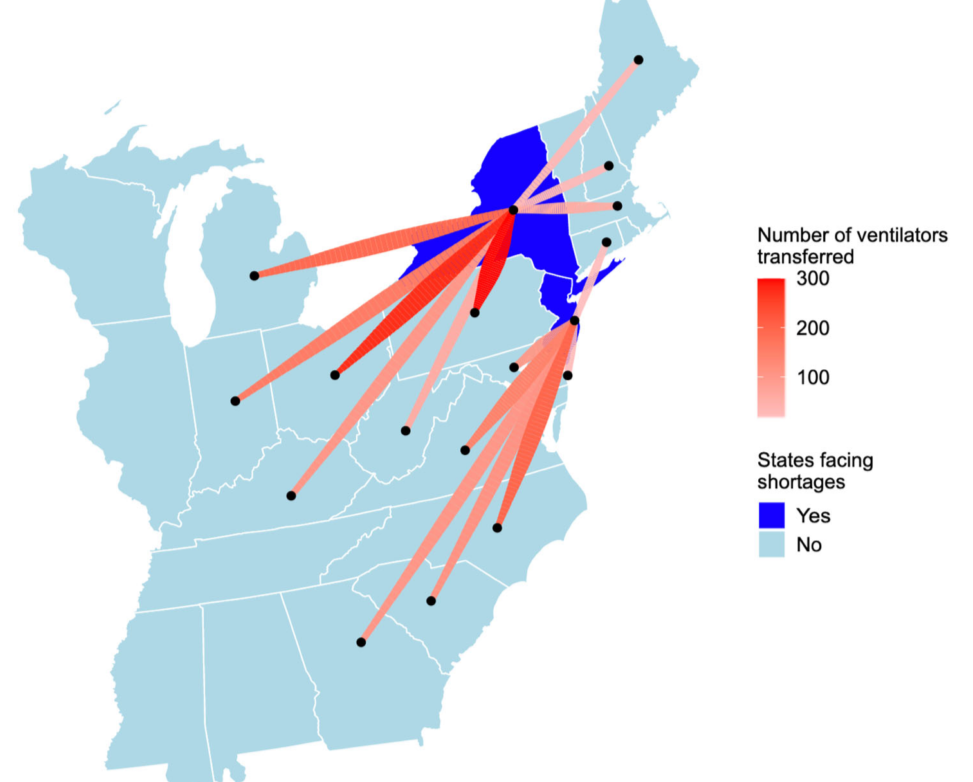} &
    \includegraphics[width=0.48\linewidth]{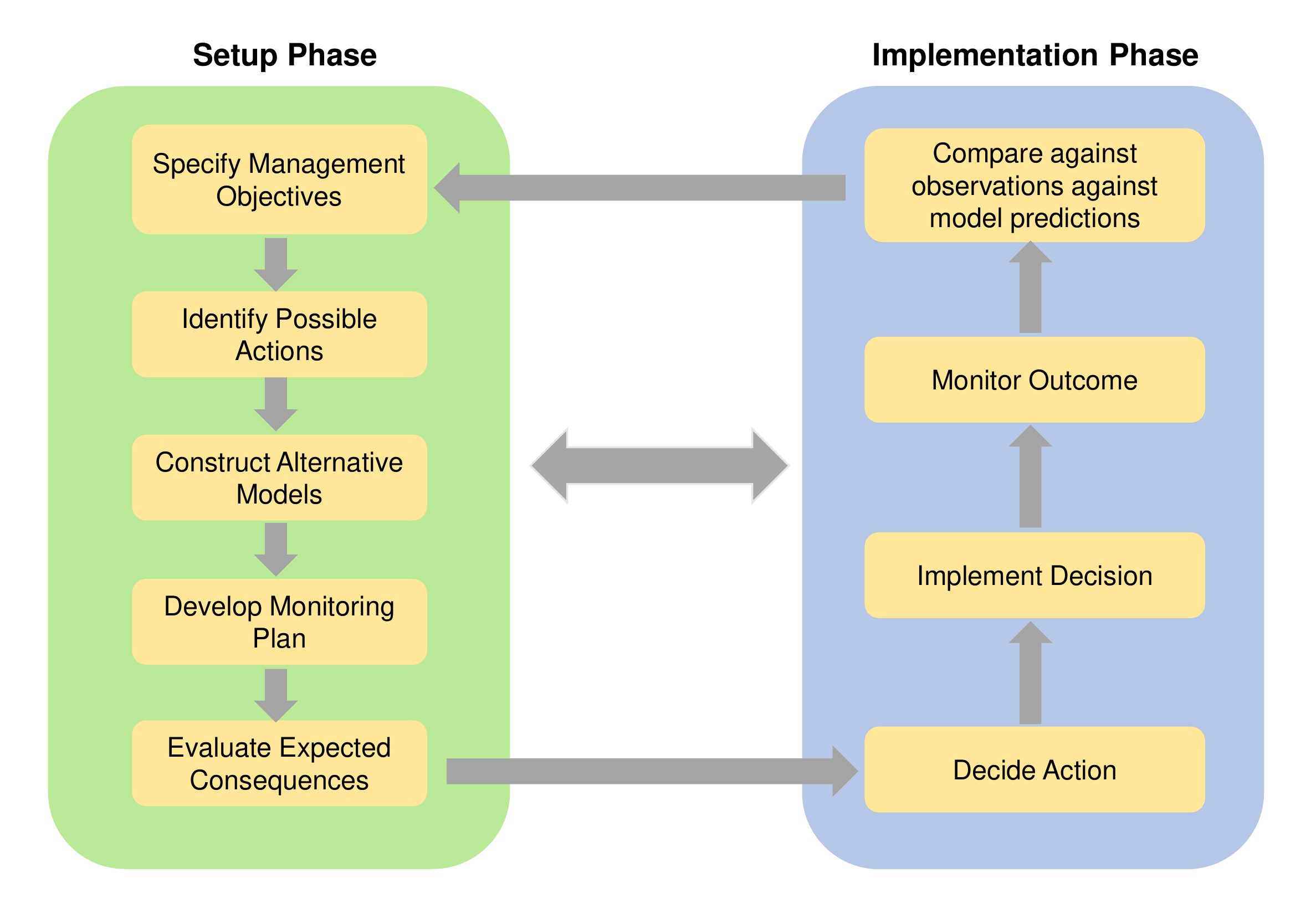} \\
    (a) Resource allocation & (b) Adaptive management \\
  \end{tabular}
    \caption{
    \textbf{Examples of forecasting in decision making.}
    (a) Optimization of ventilator allocation problem to reduce ventilator shortage and minimize transfer of ventilators across states \cite{bertsimas2021predictions} (b) Major steps of adaptive management (AM) \cite{shea2014adaptive}. AM is an iterative approach for assessing outcomes from current strategies to refine or design better interventions.}
    \label{fig:decision}
\end{figure}

Next we look at recently studied methods that leverage predictions from real-time forecasting models to aid in tactical resource-allocation problems.
\citet{bertsimas2021predictions} formulates the problem of ventilator allocation as an optimization problem that seeks to minimize ventilator shortages while reducing inter-state transportation of ventilators (Figure \ref{fig:decision}a).
They leverage data from the forecast of future cases at the state-level to prescribe an optimal strategy for allocating nationwide available ventilators.
\citet{altieri2021curating} also leveraged their prediction model for informed distribution of medical supplies directly from third party distributors (Response4Life).
\citet{nikolopoulos2021forecasting} modeled the excess demand for goods of different sectors by leveraging model predictions along with search volumes from google trends to forecast the supply chain constraints.

\section{Discussion: Open Challenges and Opportunities}
\label{sec:discussion}

We next discuss some open challenges and opportunities, based on the work we have surveyed in this article. 

\paragraph{Challenge 1: Data-related challenges}
Developing methods to address data-related issues could be a fruitful direction to improve our forecasting capabilities.
As we noted across Section~\ref{sec:data} and Section~\ref{subsec:realtime}, data is usually subjected to revisions, reporting errors, delays, anomalies, and subpar data collection standards/procedures, which vary across regions and sources of data.
We could leverage ideas from data-centric AI~\cite{strickland_2022} which advocates for improving data quality via the formulation of novel statistical learning problems, e.g., data revision dynamics~\cite{kamarthi2021back2future}, sampling bias ~\cite{sherratt2021exploring}, and inequity in data collection~\cite{rodriguez2020incorporating}.
Other problems arise in datasets containing sensitive information and involve privacy risks (e.g., EHRs).
We could exploit advances in the rapidly evolving fields of federated learning and differential privacy \cite{zhang2021survey}
as well as encouraging discussions on ethics of privacy and fairness when using this data \cite{hagendorff2020ethics}.
Finally, we want to emphasize the importance of efforts such as \cite{reinhart2021open} in building publicly accessible data archives where researchers can 
access multiple versions of the data.
Efforts on building such infrastructures that allow easy access of data archives from multiple sources 
can accelerate research on the mentioned data quality issues.

\paragraph{Challenge 2: Moving beyond short-term forecasting}
Determining and expanding our current epidemic forecasting limits remains open challenges.
Practical and theoretical discussions on the predictability of epidemics (Section~\ref{sec:proj_vs_pred}) rely on assumptions and/or current state of the art. It could be potentially useful to further formalize our current forecasting bounds and understand how they may increase given novel data streams~\cite{adiga2020bounds}. % and technical innovations.
A direction to further expand our limits could be developing methods that enable us to exploit what data can tell us about long-term patterns. 
While some works have employed mechanistic %(Section~\ref{subsec:agent}) and ML (Section~\ref{sec:neural})
(Section~\ref{sec:mechanistic}) and ML (Section~\ref{sec:stat}) approaches to address this challenge, the intersection of these two could be most fruitful.
For instance, we could leverage recent work in Scientific AI \cite{karniadakis2021physics} to bridge mechanistic and ML models~\cite{rodriguez2022einns}.
To look ahead further in the future, it is necessary to explicitly incorporate scenario-based projections (Section~\ref{sec:setup}).
These can naturally answer useful questions about the future, e.g., `what will happen if there is a new highly infectious variant?'. 
The domain of what-if scenario modeling has usually been approached using mechanistic models. However, some recent ML advances can enable new perspectives. For instance, we could leverage advances in causal inference and causal representation learning~\cite{wang2021desiderata,scholkopf2021toward} to connect multimodal data with interventions~\cite{NEURIPS2020_205e7357}.
Further, leveraging reinforcement learning to jointly model forecasting and interventions can help learn disentangle individual effects of interventions and forecasts % targets
as well as help in decision making~\cite{bastani2021efficient}.
Another interesting line of work is on developing systematic ways for assessing the quality of projections. This remains a challenge mainly because there is no ground truth for scenarios.

\paragraph{Challenge 3: Modeling multi-scale dynamics}

Another interesting direction is to develop methods to leverage multilevel relations and dynamics 
present at multiple scales. 
In the spatial scale, we can work on more principled methods which utilize spatial dynamics and hierarchical relations.
While some hierarchical regression methods (Section~\ref{sec:hierarchical}) and neural models (Section~\ref{sec:neurospatial}) incorporate spatial data, it is still an open question on how to use hierarchical and/or higher-order structures %constraints 
to produce coherent forecasts that are robust to noise and missing data.
In the spatial scale, different features and targets are typically observed at uneven temporal scales (daily vs weekly vs monthly) whereas most models assume uniform time intervals for all features. 
Therefore modeling these different temporal scales is also another potentially useful direction.
Higher order patterns and hierarchies can also be modeled along with other orthogonal aspects such as multi-scale behavioral models \cite{marathe2013computational}, evolution of phylodynamics of pathogens \cite{kraemer2021spatiotemporal} and other biological indicators.

\paragraph{Challenge 4: Improving the combination of models and wisdom of crowd predictions}
Most real-time forecasting initiatives leverage predictions from a diverse set of models (see Section \ref{subsubsec:ensemble}). %, and they might  experts
It remains a challenge on how to combine models to take advantage of the strengths of each modeling technique.
The proficiency of a single model varies across temporal  and spatial scales~\cite{cramer2021evaluation}.
Additionally, there may also be a frequent change in the set of available candidate models as new models are introduced, some teams may change their methodologies, and others may not publish predictions for some weeks~\cite{ray2020ensemble}.
Most works 
consider multiple weighting schemes on past data but there is no general consensus on the type of schemes that perform the best. 
Therefore, designing novel weighting schemes that integrate multiple models using novel techniques such as optimal ensemble weighting \cite{shahhosseini2022optimizing} and mixture of experts \cite{riquelme2021scaling,masoudnia2014mixture} is an important research area.

In this survey we also delved into another important source of forecasts: wisdom of crowds. 
Prediction markets, which are used to combine predictions from multiple sources and experts, are susceptible to similar inefficiencies that sometimes plague general markets. Information inefficiencies \cite{angelini2021informational} including successful misinformation campaigns and rampant speculation based on imperfect information can easily lead to short-term mispricing of predictive outcomes~\cite{binswanger1999stock}. Designing surveys and user interfaces that allows stakeholders and experts from multiple fields to easily input useful markers is an interesting direction.
Such inputs can be regarding uncertainty/confidence, 
beliefs on multiple conditional scenarios (to capture the underlying multimodal and conditional distributions of beliefs).
Obtaining this information from individuals with little prior training or considerable human effort is also another interesting direction of research that could build upon advances in human-computer interaction~\cite{chung2019efficient}.

\paragraph{Challenge 5: Providing well-calibrated and explainable forecasts}
Epidemic forecasts frequently enable high-stake decisions,
therefore, they should be well-calibrated (meaningful uncertainty quantification). 
Recent events suggest this remains an open challenge which may even affect which targets are communicated to the public.
In the COVID-19 pandemic, the CDC decided to temporarily remove 2- to 4-week ahead cases forecasts from their website (dashboard) because the ensemble and constituent model's probabilistic forecasts were not well-calibrated as measured by coverage~\cite{predictability_reich_2021}.
Indeed, uncertainty quantification is very challenging as there are multiple sources of uncertainty. 
Recent work with non-parametric neural models open a door to leverage richer representations of sources of uncertainty~\cite{kamarthi2021epifnp}.
Quantifying uncertainty from other multimodal data sources %(e.g., biological and genomics data) 
and distinguishing between epistemic and aleatoric uncertainty in disease spread are important directions.

Explainability of predictions to the domain experts is another important aspect to bridge forecasts with decision making.
As discussed before, simpler methods like regression and mechanistic methods can easily provide explanations, but this is harder to do with complicated neural models. Recently proposed methods used similarity with historical data points~\cite{adhikari2019epideep,kamarthi2021epifnp} and feature-level importance~\cite{rodriguez_deepcovid_2021} as means for interpretability.
There is an increasing interest in Explainable AI (XAI) which has led to useful techniques like saliency maps, importance functions, and explanations directly from the learned representations~\cite{molnar2020interpretable, ahmad2018interpretable,selvaraju2017grad}.
Exploration of such techniques in the forecasting space could be fruitful.

\paragraph{Challenge 6: Setup and evaluation for actionable forecasts}
The forecasting setup and evaluation also need to be under constant scrutiny by the community to make forecasts more actionable for public communication and decision making.
We need to work on defining new targets that can better inform public health decision making.
For instance, we could build targets that inform us of trend changes (e.g., probability of a surge).
In addition, we have to work on standard evaluation guidelines and choosing the right error metric. 
In the COVID-19 pandemic, WIS, MAE and coverage have been adopted by the CDC and the \hub~\cite{cramer2021evaluation}. However, these metrics may be susceptible to the magnitude of forecasting targets like number of hospitalizations, which largely vary across regions. This may prevent us from directly using them to quantify models’ performance across regions.
In addition, these have been used only for incidence predictions (a real-valued target). As opposed to seasonal epidemics, it remains unclear how to define and evaluate event-based tasks (e.g., peak prediction) in a pandemic. 
Reporting forecasting results is also an open question. \citet{pollett2021recommended} proposes guidelines for papers reporting methodological advances based on their experiences with influenza. 
Effective public decision-making from forecasts involves leveraging new targets and visualizations \cite{lutz2019applying} as well as training stakeholders at multiple stages for effective communication regarding both the accuracy and uncertainty of these forecasts are~\cite{predictability_reich_2021}.

\paragraph{Challenge 7: Technical debt of real-world model deployment}
Real-world deployment of forecasting systems requires a non-trivial amount of human involvement.
Human interventions require domain expertise to 
handle/remove anomalous data points, add new data streams, adapt to shifts in data distribution, or even correct/remove predictions that look incorrect to the expert eye~\cite{altieri2021curating,rodriguez_deepcovid_2021}. 
Other examples can be found in the domain inputs needed to calibrate mechanistic models, e.g., set bounds to parameter optimization~\cite{bertsimas2021predictions}.
Borrowing terminology from ML and software engineering we call these interventions as \emph{technical debt} \cite{sculley2015hidden,sculley2014machine} and they play a significant role in the success of deployed models in real-world~\cite{yu2021seven,muralidhar2021using}. 
Real-world prediction challenges as the ones described in Section~\ref{subsec:initiatives} usually assess both methodology and technical debt.
This points to the need for more work on finding better ways to disentangle the main contributors of a participating model/team's performance during real-time challenges (e.g. methodology vs expertise in technical debt activities).
On the other side, this emphasizes the importance of having simulated experiments where variables can be controlled in such a way that we can replicate results that clearly showcase the benefits of technical novelties.
Further, we should also encourage the synergy of humans and models.
Leveraging expert feedback through human-in-the-loop based reasoning systems \cite{wilder2020learning,budd2021survey} could further improve the robustness and effectiveness of our forecasting systems.

\section*{Acknowledgements}
This work was supported in part by the NSF (Expeditions CCF-1918770, CAREER IIS-2028586, RAPID IIS-2027862, Medium IIS-1955883, Medium IIS-2106961, CCF-2115126), CDC MInD program, ORNL, faculty research award from Facebook and funds/computing resources from Georgia Tech.

\newpage
\bibliographystyle{ACM-Reference-Format}

\end{document}